\documentclass[times,twocolumn,final,longtitle]{elsarticle}

\usepackage{compbime}
\usepackage{framed,multirow}
\usepackage{makecell}
\usepackage{hyperref}
\usepackage{threeparttable}

\usepackage{amssymb}
\usepackage{latexsym}

\usepackage{url}
\usepackage{xcolor}

\usepackage{hyperref}

\usepackage{makecell}
\usepackage{longtable}
\usepackage{tablefootnote}

\newcommand{\added}[1]{\textcolor{black}{#1}}
\newcommand{\changed}[1]{\textcolor{black}{#1}}

\definecolor{newcolor}{rgb}{.8,.349,.1}

\journal{/}

\begin{document}

\verso{Yihao Li \textit{et~al.}}

\begin{frontmatter}

\title{A review of deep learning-based information fusion techniques for multimodal medical image classification}%

\author[1,2]{Yihao \snm{Li}}
\author[1,2]{Mostafa \snm{El Habib Daho}\corref{cor1}}
\cortext[cor1]{Corresponding author: 
  Tel.: +33-7-52-07-30-52}
\ead{mostafa.elhabibdaho@univ-brest.fr}
\author[1,3]{Pierre-Henri \snm{Conze}}
\author[1,2]{Rachid \snm{Zeghlache}}
\author[4,5]{Hugo \snm{Le Boité}}
\author[6,5]{Ramin \snm{Tadayoni}}
\author[1,2,7]{Béatrice \snm{Cochener}}
\author[1,2]{Mathieu \snm{Lamard}}
\author[1]{Gwenolé \snm{Quellec}}

\address[1]{LaTIM UMR 1101, Inserm, Brest, France}
\address[2]{University of Western Brittany, Brest, France}
\address[3]{IMT Atlantique, Brest, France}
\address[4]{Sorbonne University, Paris, France}
\address[5]{Ophthalmology Department, Lariboisière Hospital, AP-HP, Paris, France}
\address[6]{Paris Cit\'e University, Paris, France}
\address[7]{Ophthalmology Department, CHRU Brest, Brest, France}

\received{xxx October 2023}

\begin{abstract}
Multimodal medical imaging plays a pivotal role in clinical diagnosis and research, as it combines information from various imaging modalities to provide a more comprehensive understanding of the underlying pathology. Recently, deep learning-based multimodal fusion techniques have emerged as powerful tools for improving medical image classification. This review offers a thorough analysis of the developments in deep learning-based multimodal fusion for medical classification tasks. We explore the complementary relationships among prevalent clinical modalities and outline three main fusion schemes for multimodal classification networks: input fusion, intermediate fusion (encompassing single-level fusion, hierarchical fusion, and attention-based fusion), and output fusion. By evaluating the performance of these fusion techniques, we provide insight into the suitability of different network architectures for various multimodal fusion scenarios and application domains. Furthermore, we delve into challenges related to network architecture selection, handling incomplete multimodal data management, and the potential limitations of multimodal fusion. Finally, we spotlight the promising future of Transformer-based multimodal fusion techniques and give recommendations for future research in this rapidly evolving field.
\end{abstract}


\end{frontmatter}


\section{Introduction}
\subsection{Context}

In recent years, the field of medical image analysis has seen a surge in efforts to apply deep learning-based methods to the classification of various diseases, notably related to the brain \citep{kong2022multi,zhang2022bpgan,liu2018multi}, breasts \citep{qian2020combined,dalmis2019artificial,qian2021prospective}, prostate \citep{le2017automated,yang2017co,mehrtash2017classification} and eyes \citep{li2022multimodal,yoo2022deeppdt,huang2022detecting}. The ability to accurately classify and diagnose diseases from medical images has the potential to revolutionize healthcare by improving diagnostic accuracy, reducing human error, and enabling more personalized treatment planning. This trend has highlighted the need for robust and efficient methods for analyzing medical images across multiple imaging modalities.

With advances in medical image acquisition systems, many new imaging modalities have been developed to diagnose patients \citep{muhammad2021comprehensive,azam2022review,hermessi2021multimodal}, resulting in larger and more diverse datasets. An imaging modality alone does not often provide all the information needed to ensure accurate clinical diagnosis. Therefore, clinicians increasingly base their diagnosis on images obtained from a variety of sources: a combination of abundant information can be used in clinical practice with more confidence. Following this trend and to improve diagnosis results, artificial intelligence-based classification models are increasingly being developed by combining data from multiple sources to take advantage of both redundancies and complementarities across modalities. 


Several surveys have been conducted in recent years to analyze the trends in the application of multimodality in various fields \citep{ramachandram2017deep,baltruvsaitis2018multimodal}, among which the field of medicine is gaining a great deal of attention. In medicine, several such survey papers focus on specific image analysis tasks: image fusion \citep{azam2022review}, image synthesis \citep{https://doi.org/10.48550/arxiv.2202.06997}, image segmentation \citep{zhou2019review}, or image registration \citep{el2016current}. However, medical image classification was never addressed in a comprehensive manner. A few surveys target specific fields such as neurology \citep{shoeibi2022diagnosis} or oncology \citep{lipkova2022artificial}, but they do not provide a comprehensive discussion of how multimodal fusion may be applied to other fields. To fill this gap, we propose to review deep learning-based information fusion techniques for multimodal medical classification across all medical fields. We restrict this analysis to classification, given the sufficient coverage of other analysis tasks \citep{azam2022review,https://doi.org/10.48550/arxiv.2202.06997,zhou2019review,el2016current}. We include in the scope of classification methods any method assigning class labels, possibly with probabilities, to a patient or a region of interest in the patient, regardless of the application (diagnosis, prognosis, risk estimation, etc.). Throughout our review, we summarize and discuss the advantages and disadvantages of various information fusion methods that can be applied to various organs and imaging modalities. As the first review to examine the use of deep learning in multimodal medical classification, this paper aims to guide future investigations into medical diagnosis using multiple imaging modalities.

\subsection{Traditional methods}
\label{sec:traditional-methods}

Information fusion not based on deep learning strategies, relying on traditional image processing and machine learning, has been reviewed in a previous survey \citep{kline2022multimodal}. We summarize hereafter the main developments and highlight the benefits of non-deep learning-based information fusion.

Input fusion is the most commonly used strategy among traditional methods. It involves the fusion of images from various modalities into structured data and fuses them into different categories depending on the fusion domain: spatial fusion \citep{el2016current,stokking2001integrated,bhatnagar2013directive,he2010multimodal,bashir2019swt}, frequency fusion \citep{princess2014comprehensive,parmar2012comparative,sadjadi2005comparative,das2013neuro,liu2010pet,xi2017multimodal} and sparse representation \citep{zhang2018sparse,zhu2019phase,liu2015general}. In spatial fusion, multimodal images are combined at the pixel level, but this approach often leads to spectral degradation \citep{mishra2015image} and color distortion \citep{bhat2021multi}. Frequency fusion, which involves transforming the input image into the frequency domain, is more complex and results in limited spatial resolution \citep{sharma2020pyramids}. Sparse representation, on the other hand, can be sensitive to registration errors and lacks attention to details \citep{bhat2021multi}.

Other strategies include intermediate and output fusion, which do not require registration of the input images. Intermediate fusion involves extracting features from different imaging modalities, concatenating them, and feeding them into a classifier, generally a support vector machine (SVM), for diagnosis \citep{lee2019machine,tang2020elaboration,quellec2010case}. This approach requires extensive testing and rich domain knowledge for feature extraction and selection. On the other hand, output fusion involves stacking the data results from unimodal models and combining them \citep{lalousis2021heterogeneity}. While this approach circumvents the need for early integration, it presents its own set of challenges. Individual models in output fusion may be heavily influenced by their respective modality-specific idiosyncrasies, potentially introducing biases into the final combined output. Moreover, if these unimodal models yield correlated or redundant information, the utility of stacking them diminishes, as it might not deliver significant additional value.

Traditional methods typically involve complex pre-processing steps paired with relatively simple model structures. Such a combination frequently leads to information loss during feature extraction, thereby complicating efforts to fully leverage the synergies between various imaging modalities.

Besides requiring domain knowledge, these traditional multimodal fusion approaches do not fully utilize the complementarity between multimodal features. These limitations highlight the need for more advanced techniques, such as deep learning-based multimodal fusion methods, able to overcome the challenges faced by traditional methods. Deep learning network architectures offer complex models that can explore more possibilities for multimodal fusion. Furthermore, various end-to-end models significantly reduce the amount of domain knowledge required for diagnosis purposes, albeit at the cost of interpretability \citep{salahuddin2022transparency}. 

\subsection{Development trends}

\begin{figure}[!t]
\centering
\includegraphics[width=0.4\textwidth]{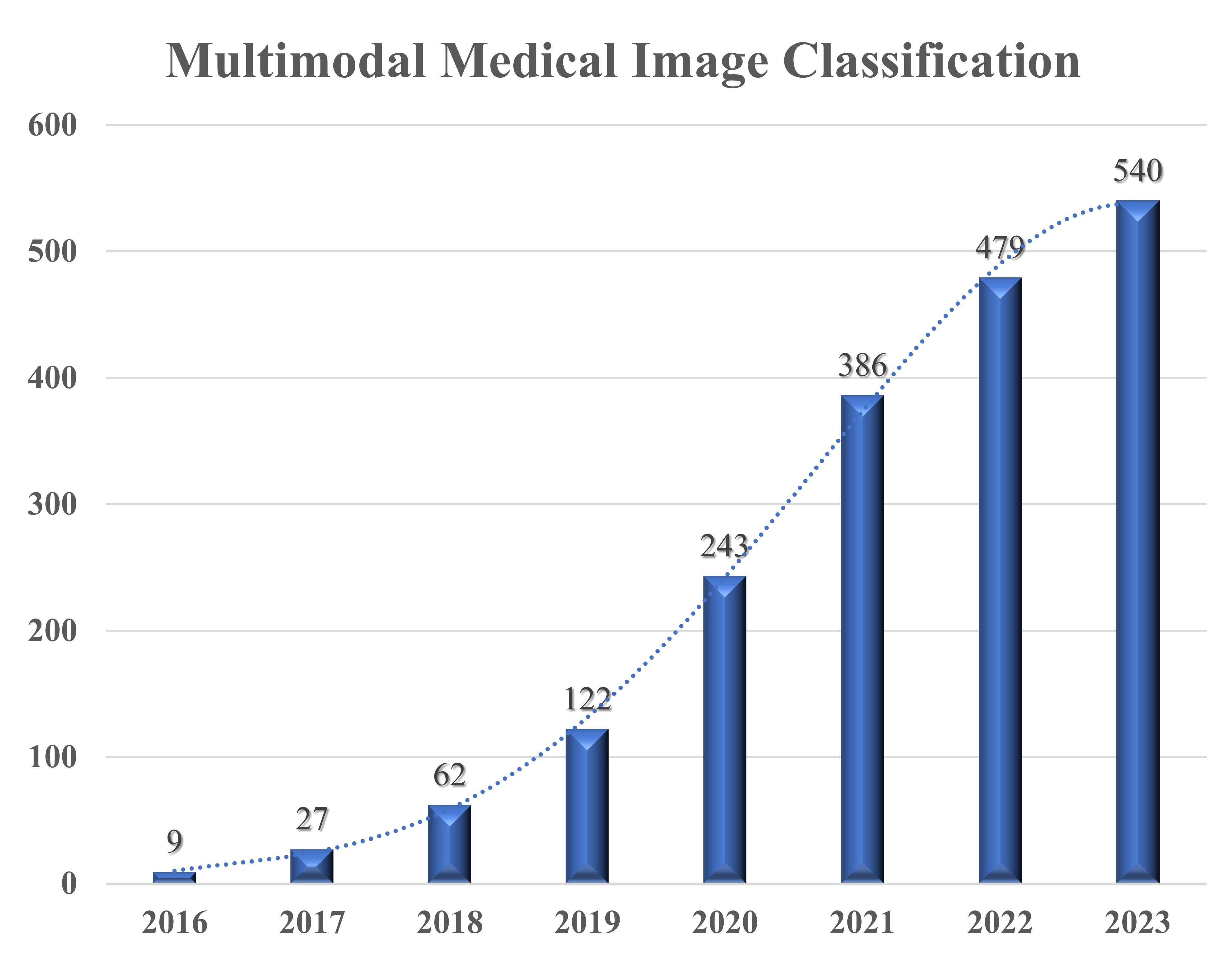}
\caption{Number of publications on medical multimodal image classification. Per-year statistics obtained using PubMed from 2016 to \changed{2023}.}\label{fig_number}
\end{figure}

\begin{figure}[!t]
\centering
\includegraphics[width=0.4\textwidth]{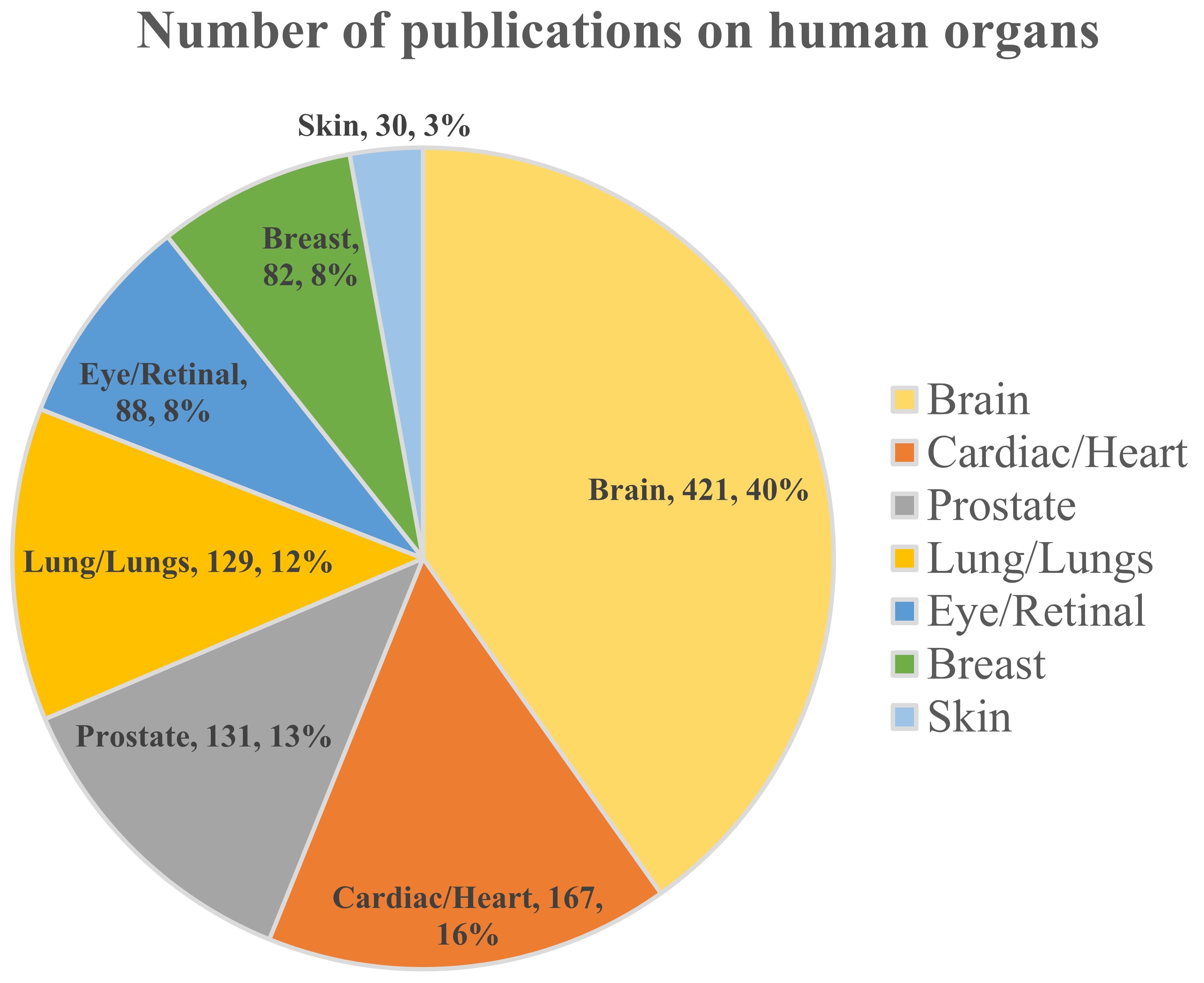}
\caption{Number of publications dealing with medical multimodal image classification on human organs, from 2016 to \changed{2023}. Tags: organ, number of publications, percentage.}\label{fig_organ}
\end{figure}

Recognizing the potential of deep learning-based methods for multimodal medical image classification, researchers have increasingly focused on this area. In order to obtain more accurate diagnoses, multimodal medical image analysis have also become a growing trend. Fig.~\ref{fig_number} shows the number of publications about multimodal medical classification each year, which was queried on \changed{February 27, 2024}, on PubMed. As illustrated by the figure, the number of papers has increased yearly from 2016 to \changed{2023}, indicating that multimodal medical classification tasks based on deep learning have gained greater attention in recent years. Furthermore, we report the number of publications on different organs in multimodal diagnosis tasks in Fig.~\ref{fig_organ}. We found that brain-related publications currently account for a substantial portion of multimodal studies. This is due to the disclosure of many large multimodal image datasets on the brain. On the other hand, not many studies were conducted on other organs, except whenever a public dataset was released. This finding motivated us to focus the review on studies performed on public datasets from various organs. One advantage is to allow direct quantitative comparisons between methods. 

\begin{figure*}[!p]
\centering
\includegraphics[width=2.0\columnwidth]{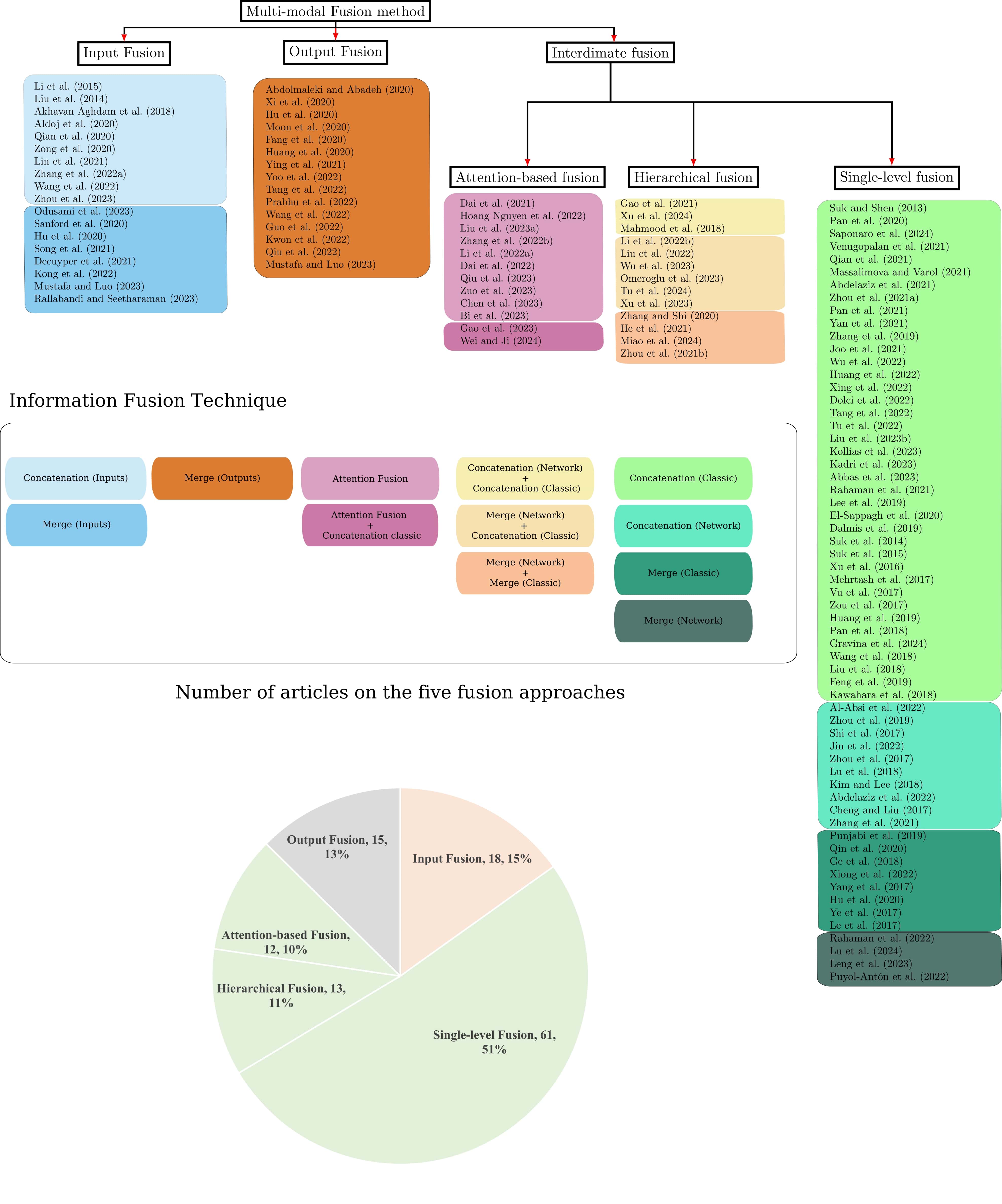}
\caption{\added{Overview and proportion of deep learning-based information fusion techniques for multimodal medical image classification presented in this paper.}}
\label{fig_overview_lit}
\end{figure*}

\subsection{Paper selection}

\begin{table*}[!t] 
\caption{A list of multimodal image datasets. The list is sorted by the number of publications on PubMed (Keywords: dataset name AND 'multimodal'). Details about imaging modalities are given in Tab.\ref{tab_medical}}
\label{tab_dataset}
\centering
\begin{threeparttable}
\begin{tabular}{|l|c|l|l|l|l|}
\hline
Dataset & \makecell[c]{Year} & \makecell[c]{Modalities} & Body Organ(s) & Medical Diagnosis & EHR  \\
\hline
ADNI\tnote{1} & \makecell[c]{(2004-2009)\\(2010-2016)\\(2016-2022)\\(2023-2027)} & sMRI, fMRI, PET & Brain & Alzheimer’s Disease & Available \\
\hline
BraTS\tnote{2} & \makecell[c]{(2012-2023) \\yearly} & MRI (TI, T2, T1c, FLAIR) & Brain & Brain Tumor & N/A \\
\hline
TCIA\tnote{3} & 2014 & \makecell[l]{CT, MRI, PET, US, etc.} & \makecell[l]{Brain, Breast, Lung, \\Kidney, Head-Neck, \\Liver, Pancreas, etc.} & Common Cancer Disease & Available \\
\hline
OASIS\tnote{4} & 2007 & MRI, PET & Brain & Alzheimer’s Disease & Available \\
\hline
SPC\tnote{5} & 2018 & \makecell[l]{Dsc, Clinical Image, \\Metadata} & Skin & Skin Lesion & Available \\
\hline
TCGA\tnote{6} & 2006 & \makecell[l]{Pathological data, \\ Genomic data} & Brain, Lung, etc. & Common Cancer Disease & Available \\
\hline
ABIDE\tnote{7} & 2012 & sMRI, fMRI & Brain & \makecell[l]{Autism Spectrum \\ Disorder (ASD)} & Available \\
\hline
ADHD-200\tnote{8} & 2011 & sMRI, fMRI & Brain & \makecell[l]{Attention Deficit \\Hyperactivity Disorder \\(ADHD)} & Available \\
\hline
COBRE\tnote{9} & 2012 & sMRI, fMRI & Brain & \makecell[l]{Schizophrenia} & Available \\
\hline
GAMMA\tnote{10} & 2021 & OCT, Fundus Image & Eye & \makecell[l]{Glaucoma} & N/A \\
\hline
CPM-RadPath\tnote{11} & \makecell[c]{2019, \\2020} & MRI (TI, T2, T1c, FLAIR) & Brain & \makecell[l]{Brain Tumor} & N/A \\
\hline
ISIT-UMR\tnote{12} & 2019 & \makecell[l]{White Light RGB, \\Narrow Band Imaging\\(NBI)} & Digestive Tract & \makecell[l]{Gastrointestinal Lesions} & N/A \\
\hline
MRNet\tnote{13} & 2018 & MRI (TI, T2) & Knee & Knee Injuries & N/A \\
\hline
CTU-UHB\tnote{14} & 2014 & CT (FHR, UC) & Uterus & Fetal Distress Diagnosis & Available \\
\hline
\end{tabular}
\begin{tablenotes}
\footnotesize
    \item[1] \url{https://adni.loni.usc.edu/}
    \item[2] \url{http://braintumorsegmentation.org/}
    \item[3] \url{https://www.cancerimagingarchive.net/}
    \item[4] \url{https://www.oasis-brains.org/}
    \item[5] \url{https://derm.cs.sfu.ca/Welcome.html}
    \item[6] \url{https://www.cancer.gov/ccg/research/genome-sequencing/tcga}
    \item[7] \url{http://fcon_1000.projects.nitrc.org/indi/abide/}
    \item[8] \url{http://fcon_1000.projects.nitrc.org/indi/adhd200/}
    \item[9] \url{https://fcon_1000.projects.nitrc.org/indi/retro/cobre.html}
    \item[10] \url{https://aistudio.baidu.com/aistudio/competition/detail/90/0/introduction}
    \item[11] \url{https://zenodo.org/records/3718894}
    \item[12] \url{http://www.depeca.uah.es/colonoscopy_dataset/}
    \item[13] \url{https://journals.plos.org/plosmedicine/article?id=10.1371/journal.pmed.1002699}
    \item[14] \url{https://physionet.org/content/ctu-uhb-ctgdb/1.0.0/}
\end{tablenotes}
\end{threeparttable}
\end{table*}

In our initial literature search, we identified a total of 14 public multimodal image datasets. These datasets are detailed in Sect.~ \ref{sec:multimodal image datasets}, with a summary presented in Tab.~\ref{tab_dataset}. The methodology for finalizing the list of papers for this review was as follows: 

\begin{enumerate}
    \item For each of the 14 datasets, we conducted a search on PubMed for publications that mentioned the dataset name, coupled with any of the following terms: (multimodality), (multimodal), (multi-modal), (multiparametric), or (multi-parametric).
    \item We then concatenated the 14 resulting lists.
    \item Based on the abstracts, we handpicked articles that addressed multimodal information fusion through deep learning methods.
\end{enumerate}

Notably, there are gaps in the availability of public multimodal datasets focused on classification tasks for certain organs—namely, the breast, lung, prostate, kidneys, larynx, heart, and liver, even though they are frequently discussed in the multimodal medical image analysis literature. To ensure a comprehensive review, we expanded our scope to include 19 pertinent articles that target these organs but utilize private datasets. This brought our final tally to \changed{114} publications.

\subsection{Highlights}

\begin{figure}[!t]
\centering
\includegraphics[width=0.5\textwidth]{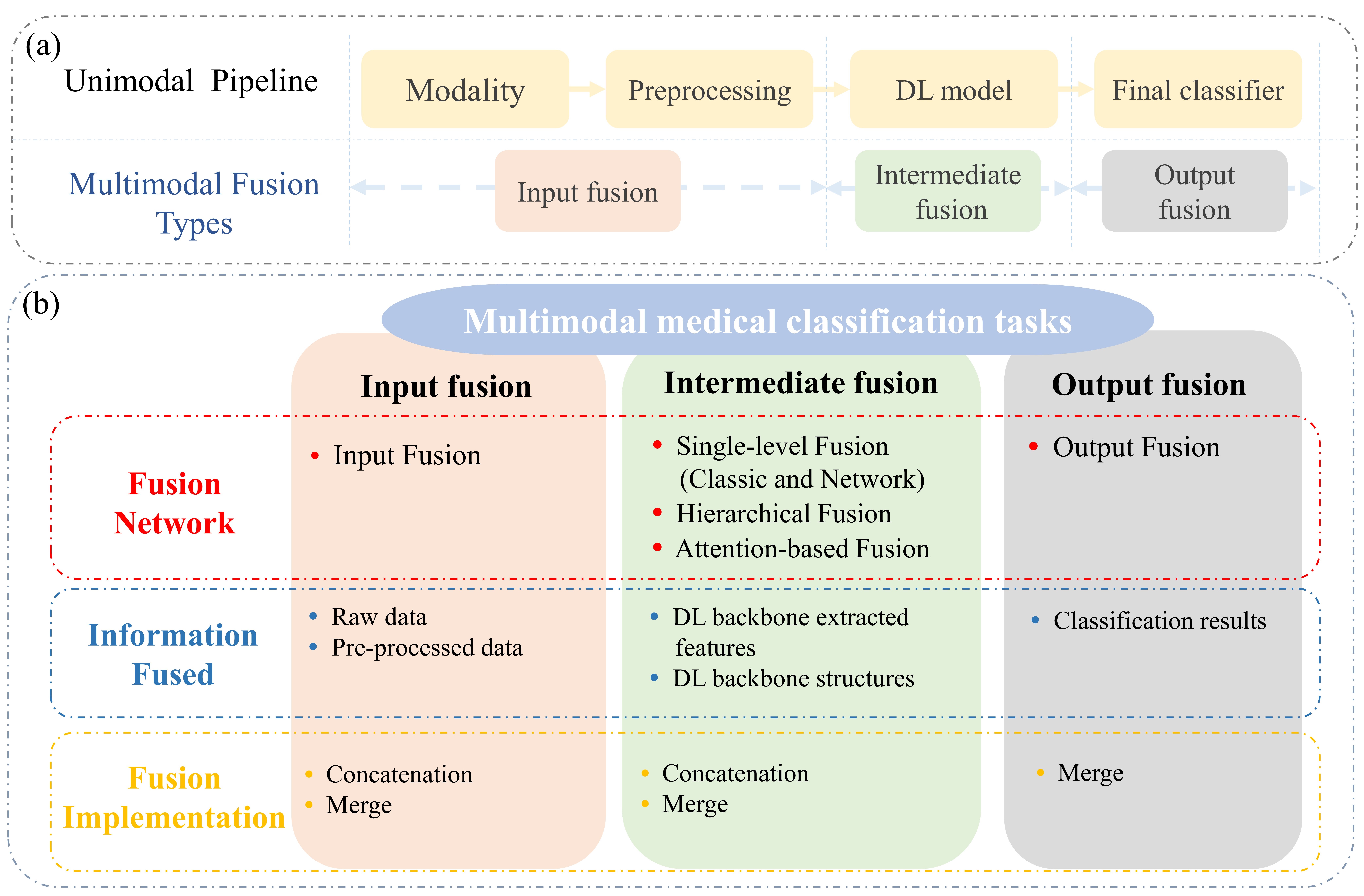}
\caption{(a) Unimodal classification task flow and different types of multimodal fusion based on the level in which they perform information fusion. (b) Information fusion networks for the three types of multimodal fusion, inputs to information fusion, and the implementation of information fusion.} \label{fig_taxnomy}
\end{figure}

Through our examination of the deep learning-based multimodal image classification literature \added{(overview presented in Fig.~\ref{fig_overview_lit})}, we propose in this paper an updated taxonomy for multimodal information fusion. As discussed in Sect.~\ref{sec:traditional-methods} and other surveys \citep{ramachandram2017deep,muhammad2021comprehensive,boulahia2021early}, multimodal fusion methods are traditionally classified as \textit{input fusion}, \textit{intermediate fusion} or \textit{output fusion}, based on the stage of information fusion in the classification pipeline, as in Fig.~\ref{fig_taxnomy}(a). Note that some publications refer to input fusion as early fusion, while intermediate fusion may be considered as feature-level fusion, and output fusion is equivalent to decision-level fusion or late fusion \citep{boulahia2021early,ramachandram2017deep,li2022multimodal}. Our analysis points to intermediate fusion as the prevailing category at present. To grant readers a more in-depth understanding of multimodal deep learning networks, we further segment intermediate fusion into \textit{single-level fusion}, \textit{hierarchical fusion}, and \textit{attention-based fusion}, as illustrated in Fig.~\ref{fig_taxnomy}(b). The proposed taxonomy is detailed and discussed in Sect.~\ref{sec:Information fusion taxonomy for multimodal image classification}: it covers the majority of the current multimodal classification network architectures, providing insight into their stages and styles of information fusion.

In this paper, we present the following contributions:

\begin{itemize}
  \item [(1)] \textit{Identify the process of medical multimodal classification.}\\
  The methodological approach of deep learning-based multimodal classification can divided into four steps: data processing, deep learning network, multimodal information fusion, and the final classification algorithm. Crucially, the classification of multimodal information fusion methods hinges on the sequential positioning of these stages.   
  Propose network architectures apt for generic multimodal fusion classification endeavors.
  \bigskip
  \bigskip
  \item [(2)] \textit{Propose network architectures for generic multimodal fusion classification tasks.}\\
  In order to address medical classification tasks involving different organs and imaging modalities, we summarized five \changed{strategies of multimodal fusion}: input fusion, single-level fusion, hierarchical fusion, attention-based fusion, and output fusion. These fusion methods can be applied to any multimodal classification problem in medicine, allowing for greater flexibility and potential for improved results.
  \bigskip
  \bigskip
  \item [(3)] \textit{Present the prevailing challenges and predict future trends.}\\
  While it's apparent that multimodal fusion is still in its early stages, our paper analyzes specific challenges tied to this domain and predicts future trends in the field.
\end{itemize}

The remainder of the paper is organized as follows: Sect.~\ref{sec: multimodal medical images} describes commonly used multimodal data for medical multimodal classification tasks and their publicly available datasets. Sect.~\ref{sec: multimodal classification pipeline} describes the multimodal medical image classification task process mentioned in contribution 1. A review of papers implementing each of the five fusion strategies of contribution 2 is presented in Sect.~\ref{sec: multimodal classification networks}. The purpose of Sect.~\ref{sec: Discussion} is to discuss the existing problems and to make predictions regarding future fusion methods in contribution 3. Finally, Sect.~\ref{sec: Conclusion} contains our concluding comments. A list of frequently used abbreviations throughout the paper is shown in Tab.~\ref{tab_terms}.

\section{Multimodal medical images}
\label{sec: multimodal medical images}

\subsection{Imaging modalities}

For medical diagnosis purposes, each imaging modality has its own characteristics and information. Different medical imaging modalities use different frequency bands of the electromagnetic spectrum in order to screen and diagnose different medical conditions in the human body \citep{azam2022review}. There are different wavelengths and frequencies associated with each imaging modality, as well as different characteristics (structure, function, etc.) \citep{singh2012application}. Furthermore, medical imaging modalities can be classified as invasive or non-invasive. Invasive methods involve inserting an object into the body through an incision or needle injection in order to examine an organ, while non-invasive methods utilize some form of radiation or sound \citep{azam2022review}. Table~\ref{tab_medical} shows some modalities that appear in multimodal medical image datasets.

\begin{table*}[!t]
\centering
\caption{Some examples of imaging modalities and organs found in the multimodal medical image analysis literature.}\label{tab_medical}
\begin{tabular}{|c|c|c|c|}
\hline
Modalities & \makecell[c]{Body organs \\examined} & \makecell[c]{Invasive/\\Non-invasive} & Description  \\
\hline
\makecell[c]{Magnetic \\Resonance Image\\ (MRI)} & \makecell[c]{Brain, Prostate,\\Breast, etc.} & Non-Invasive&  \makecell[c]{In addition to high spatial resolution \\and exquisite soft tissue contrast, \\MRI can also display dynamic physiologic \\changes in three dimensions \citep{plewes2012physics}.}\\
\hline
\makecell[c]{Positron Emission \\Tomography\\ (PET)} & \makecell[c]{Brain, Prostate,\\Breast, etc.} & Invasive &  \makecell[c]{The PET provides information \\about the organs' activity, as well as \\its sugar use as energy\cite{bailey2005positron}.}\\
\hline
\makecell[c]{Computed \\ Tomography\\ (CT)} & \makecell[c]{Lung, Bone,\\ Oral, etc.} & \makecell[c]{Non-Invasive\\(harmful)}&  \makecell[c]{CT is an excellent tool for detecting bone,\\ joint, and soft tissue lesions that may\\ affect bone, joints, or soft tissues \citep{buzug2011computed}.}\\
\hline
\makecell[c]{Ultrasound\\ (US)} & \makecell[c]{Abdomen, \\Breast, etc.} & \makecell[c]{Non-Invasive}&  \makecell[c]{In addition to showing the activity and \\function of certain organs in the body, US can \\also identify whether a tissue or organ contains \\fluid or gas \citep{leighton2007ultrasound}.}\\
\hline
\makecell[c]{Optical Coherence \\Tomography\\ (OCT)} & \makecell[c]{Eye, Heart} & \makecell[c]{Non-Invasive}&  \makecell[c]{Biological tissues can be visualized \\in high-resolution with OCT scanning in \\two-dimensional or three-dimensional modes \citep{huang1991optical}.}\\
\hline
\makecell[c]{Dermatoscope\\ (Dsc)} & \makecell[c]{Skin} & \makecell[c]{Non-Invasive}&  \makecell[c]{Dsc allows better visualization of \\subsurface structures and improved \\identification of skin diseases \citep{mackie2002use}.}\\
\hline
\makecell[c]{Color Fundus \\Photographs (CFP)} & \makecell[c]{Eye} & \makecell[c]{Non-Invasive}&  \makecell[c]{CFP monitors the progression of\\ eye disorders using color photographs \\taken with a fundus camera \citep{besenczi2016review}.}\\
\hline
\end{tabular}
\end{table*}

Due to their complementary nature, there has been a significant focus on the following combinations of modalities targetting various diseases: (1) multi-parametric MRI (TI, T2, T1C, and FLAIR) \changed{\citep{decuyper2021automated,ye2017glioma,kollias2023btdnet,xu2024cross,wu2023aggn}}, (2) MRI and PET \changed{\citep{liu2014multimodal,fang2020ensemble,gao2023multimodal,gravina2024multi}}, (3) PET and CT \citep{qin2020fine}, (4) multi-view ultrasound (US B-mode, US color Doppler) \citep{qian2020combined,qian2021prospective}, (5) Color Fundus Photographs (CFP) and Optical Coherence Tomography (OCT) \citep{wu2022gamma,li2022multimodal,elhabibdaho2023OMIA}, (6) Dermatoscope (Dsc) and Clinical Images \changed{\citep{tang2022fusionm4net,kawahara2018seven,wei2024multi}}, and (7) combined diagnosis of Image Data and Clinical Data \changed{\citep{yap2018multimodal,prabhu2022multi,venugopalan2021multimodal,liu2023improving}}. The complementary relationships between these modalities will be briefly discussed. 

Neurology and neurosurgery frequently use MRI.  Different MRI images can be obtained by changing the factors affecting the magnetic resonance (MR) signal, and these different images are referred to as sequences. Depending on the sequence used, the behavior of tumors may vary, and it is essential to use multiple sequences to accurately determine tumor location and size \citep{pai2020brahma}. T1-weighted (T1) and T2-weighted (T2) MRIs are the most common MRI sequences. Tomographic anatomical maps can be observed with the T1 sequence, and the T2 sequence clearly shows the location and size of the lesion \citep{lindig2018evaluation}. The Fluid Attenuated Inversion Recovery (Flair) sequence provides better visualization of the area around the tumor site, making it easier to detect the tumor's boundaries \citep{hecht2001mri}. Furthermore, contrast-enhanced T1-weighted (T1c) sequences can be used to detect intra-tumor conditions and distinguish tumors from non-tumorigenic lesions \citep{kuban2003long}.  T2 and Flair are suitable for detecting tumors with peritumoral edema, while T1 and T1c are suitable for detecting tumors without peritumoral edema \citep{zhou2019review}.

Diffusion-weighted imaging (DWI) is another useful sequence designed to detect the random movements of water protons. Therefore, DWI sequence is a highly sensitive method for detecting acute strokes \citep{preston2006magnetic}. An increased apparent diffusion coefficient (ADC) value with lower signals of DWI images could reveal the fast diffusion of water molecules \citep{shen2011use}. 
In addition to using multiple sequences, co-diagnosis using structural MRI (sMRI) and functional MRI (fMRI) is becoming increasingly popular \citep{akhavan2018combination,liu2022attention}. fMRI measures the small changes in blood flow that occur with brain activity. This test can be used to determine which parts of the brain are performing critical functions and to determine the effects of strokes and other diseases on the brain \citep{bandettini2012twenty}.

\begin{figure}[!t]
\includegraphics[width=0.5\textwidth]{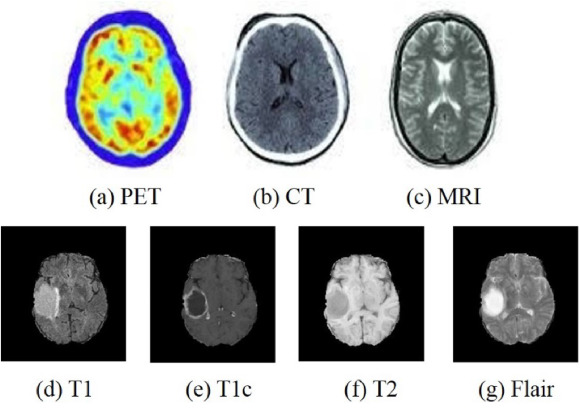}
\caption{(a)-(c) are the images of PET, CT, and MRI. (d)-(g) are the different sequences of MRI. Images from \citet{zhou2019review}.} \label{fig_mri}
\end{figure}

The combination of PET and MRI, PET and CT has been recognized as a valuable method for screening and diagnosing various diseases \citep{calhoun2016multimodal,liu2017multi,huang2019diagnosis,xu2022joint,andrearczyk2022overview}. The PET scan is preceded by the administration of a radioactive agent to the patient. This allows doctors to determine the metabolic processes in which the brain tissue is involved \citep{bailey2005positron}. Compared to other imaging methods such as CT and MRI, PET has a high sensitivity and can detect lesions even if MRI/CT does not yet show abnormalities. PET also has high specificity, making it possible to determine whether a tumor is malignant based on its metabolism at the time of MRI/CT detection \citep{muehllehner2006positron}. However, because PET scan lacks information about organ anatomy, they should be conducted in conjunction with CT/MRI scans \citep{akhavan2018combination}. Indeed, the combination of PET and MRI/CT scans provides structural and functional information related to various diseases, improving the effectiveness of diagnosis. Fig.~\ref{fig_mri} shows the images of PET, CT, and MRI, as well as several sequences of MRI. 

\begin{figure}[!t]
\includegraphics[width=0.5\textwidth]{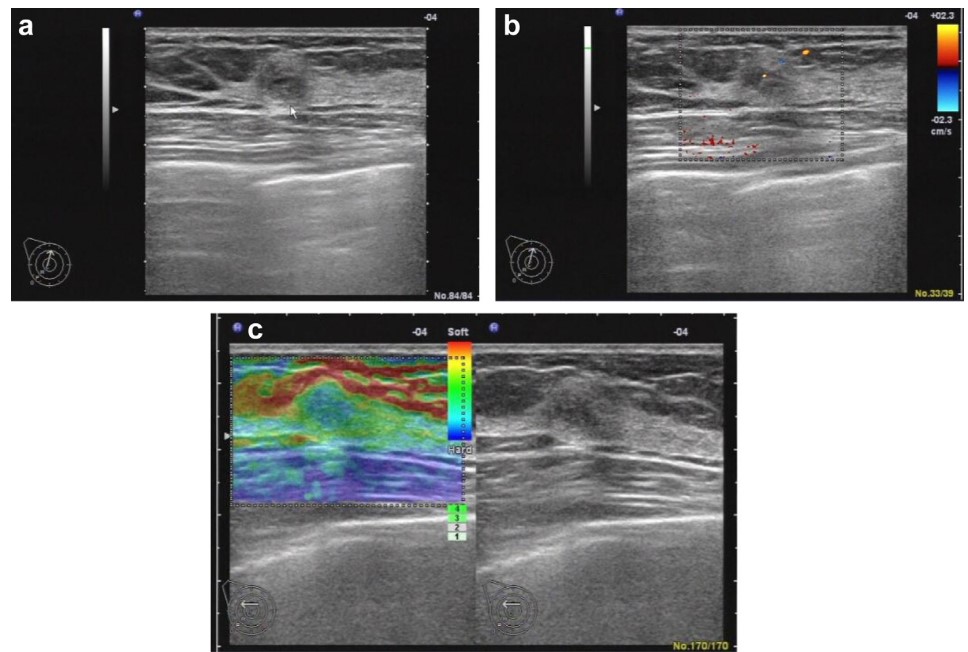}
\caption{Breast ultrasound images of a 37-y-old woman with fibroadenoma. (a) Conventional B-mode, (b) Color Doppler, and (c) Strain elastography (SE) image \citep{li2017b}. } \label{fig_US}
\end{figure}

Availability, low cost, and safety make ultrasonography the most widely used clinical diagnostic tool, with applications ranging from breast cancer diagnosis to cervical lymph node detection. Conventional B-mode imaging is used to examine abnormal masses in tissues, Color Doppler imaging shows the distribution of blood vessels within tissues \citep{zwiebel2005introduction}, while Strain Elastography (SE) is a qualitative technique and provides information on the relative stiffness between one tissue and another. For example, the combined use of Conventional B-mode imaging and Color Doppler is common in identifying cervical lymph nodes \citep{abdelgawad2020b}, diagnosing breast cancer \citep{qian2020combined,qian2021prospective}, and so forth \citep{lu2010comparative,schelling2000combined,schelling1997optimized}. Fig.~\ref{fig_US} shows the US images of Conventional B-mode, Color Doppler, and Strain Elastography.

\begin{figure}[!t]
\centering
\includegraphics[width=0.4\textwidth]{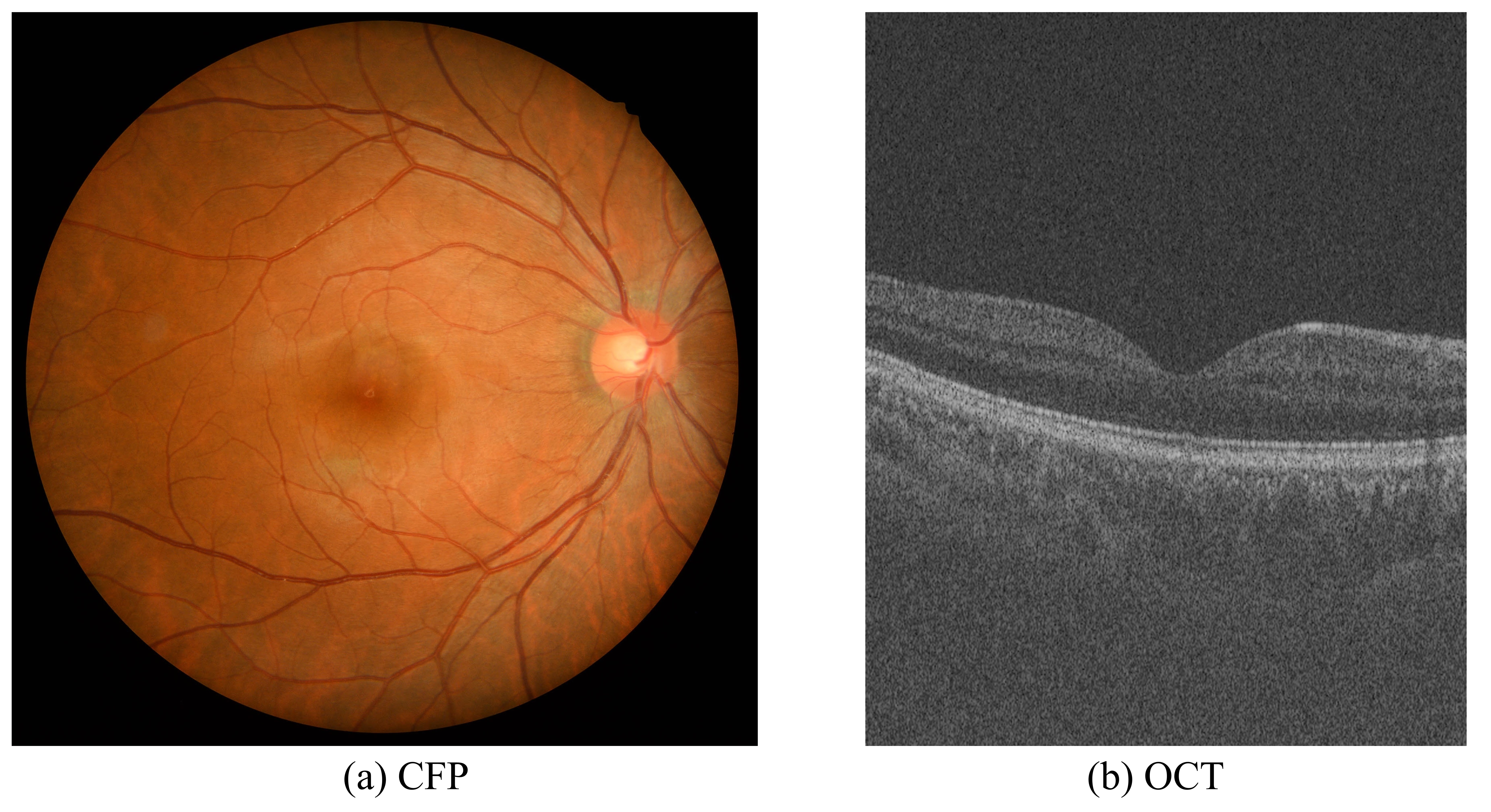}
\caption{Images of CFP and OCT from GAMMA challenge \citep{wu2022gamma}.} \label{fig_oct}
\end{figure}

In the diagnosis of ophthalmic diseases, CFP and OCT are the two most cost-effective methods \citep{li2022multimodal}. These imaging modalities provide prominent biomarkers that can be used to identify glaucoma suspects, such as the vertical cup-to-disc ratio (vCDR) on fundus images and the retinal nerve fiber layer thickness (RNFL) on an OCT image. A more accurate and reliable diagnosis, compared to a single modality, is often achieved by taking both screenings in clinical practice \citep{wu2022gamma}. Fig.~\ref{fig_oct} shows the images of CFP and OCT. 

\begin{figure}[!t]
\centering
\includegraphics[width=0.4\textwidth]{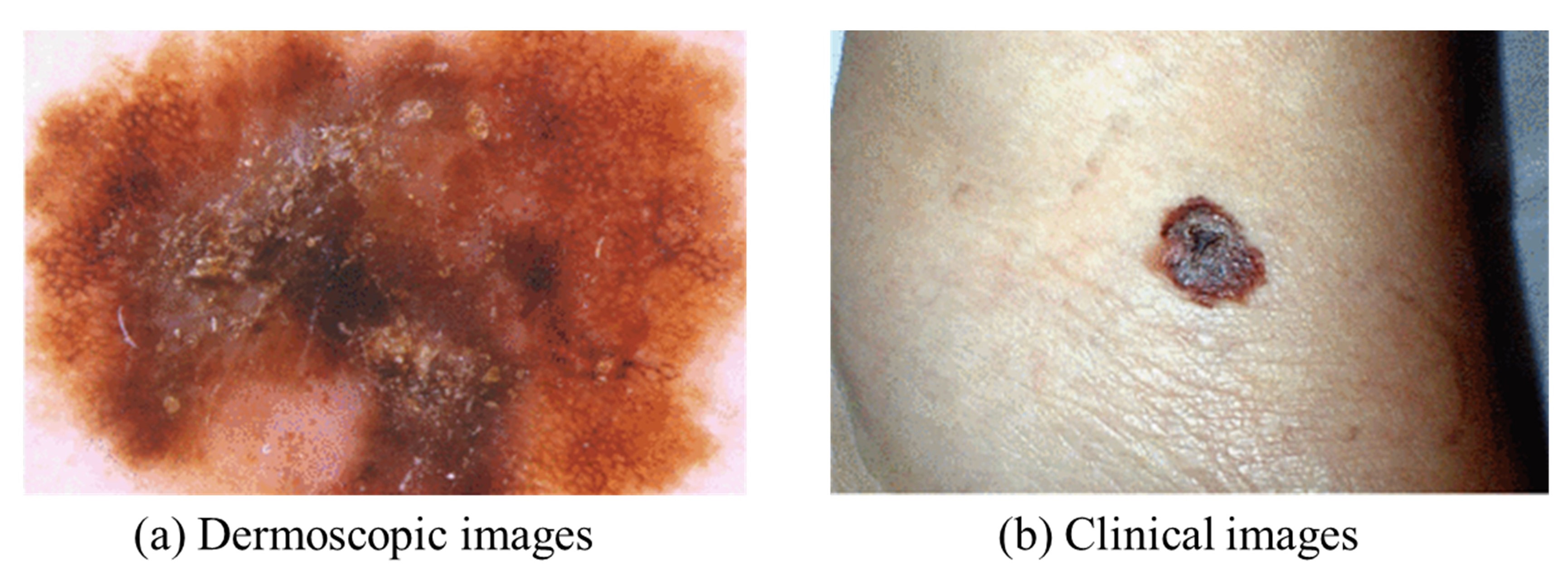}
\caption{Dermoscopic and clinical image. Image from public datasets SPC \citep{kawahara2018seven}.} \label{fig_dsc}
\end{figure}

In the diagnosis of skin cancer, a combination of dermoscopic and clinical images is often used \citep{tang2022fusionm4net}. The clinical image is captured using a digital camera and shows the visualized feature in different views and lighting conditions. On the other hand, dermoscopic images provide a clear view of the skin's subsurface structures and are obtained using a specific skin imaging technique in contact with the skin \citep{ge2017skin}. Fig.~\ref{fig_dsc} shows examples of the dermoscopic and clinical images.

In addition to multimodal image combinations, clinical information regarding the patient's medical history and symptoms can significantly contribute to the diagnosis of the disease. These data may contain implicit features that may improve the model's classification performance. Electronic Health Records (EHR) are commonly used to detect brain diseases by integrating image analysis features \citep{prabhu2022multi,venugopalan2021multimodal}. Similarly, skin cancer detection also relies heavily on metadata \citep{tang2022fusionm4net,yap2018multimodal}.

\subsection{Multimodal image datasets}
\label{sec:multimodal image datasets}

In multimodal medical diagnosis, multimodal datasets are particularly valuable for testing various networks and developing fusion methods. However, the privacy and cost of medical images often make obtaining more comprehensive multimodal datasets challenging for researchers. Fortunately, there are several freely available multimodal datasets. These datasets provide information regarding the diagnosis of diseases at various locations in the body, as well as the analysis of various multimodal combinations. These datasets are expected to contribute to the analysis of fusion methods and serve as a foundation for the future development of multimodal fusion methods.

Alzheimer's Disease Neuroimaging Initiative (ADNI) is a multi-center longitudinal study to discover clinical, imaging, genetic, and biochemical biomarkers for Alzheimer's disease (AD). ADNI has three stages: ADNI 1 included 400 subjects diagnosed with mild cognitive impairment (MCI), 200 subjects with early AD, and 200 elderly control subjects \citep{petersen2010alzheimer}; ADNI 2 added new participant groups: 150 elderly controls, 100 EMCI subjects, 150 late mild cognitive impairment (LMCI) subjects, and 150 mild AD patients \citep{beckett2015alzheimer}; ADNI 3 added hundreds of new MCI subjects, mild AD subjects, and elderly controls \citep{weiner2017alzheimer}. \\
The MRI Brain Tumor Segmentation (BraTS) challenge has been held since 2012 and currently includes classification tasks in addition to tumor segmentation \citep{menze2014multimodal}. Each subject has four MRI modalities (T1, T1C, T2, and T2 FLAIR), human annotation of tumor segmentation, and tumor grade. \\
The Cancer Imaging Archive (TCIA) is a large-scale public database containing medical images of common tumors (lung cancer, prostate cancer, etc.) and corresponding clinical information (treatment protocol, genetics, pathology, etc.) \citep{clark2013cancer}. \\
Open Access Series of Imaging Studies (OASIS) seeks to make neuroimaging datasets freely accessible to the scientific community \citep{marcus2010open}. OASIS-3 contains 755 cognitively normal adults and 622 individuals at various stages of cognitive decline ranging in age from 42-95 years \citep{lamontagne2019oasis}. \\
Seven-point Criteria Evaluation Database (SPC) provides a database for evaluating computerized image-based prediction of the 7-point malignancy checklist for skin lesions. The dataset contains more than 2000 clinical and dermoscopy color images and structured metadata for training and evaluating computer-aided diagnosis systems \citep{kawahara2018seven}. \\
As part of the Cancer Genome Atlas (TCGA), an internationally recognized cancer genomics project, more than 20,000 primary cancer samples and matched normal samples were molecularly characterized \citep{weinstein2013cancer,tomczak2015review}. \\
The Autism Brain Imaging Data Exchange (ABIDE) initiative now includes two large-scale collections, ABIDE I and ABIDE II, whose ultimate goal is to facilitate discovery science and comparative analysis across samples. ABIDE I contains 1112 datasets, including 539 from individuals with ASD and 573 from typical controls (ages 7-64 years, median 14.7 years across groups) \citep{di2014autism}. ABIDE II contains 1114 datasets from 521 individuals with ASD and 593 controls (age range: 5-64 years) \citep{di2017enhancing}. \\
ADHD-200 Sample is a grassroots initiative that aims to improve scientific understanding of the neural basis of ADHD through the implementation of open data sharing and discovery-based research methods \citep{adhd2012adhd}. \\
The Center for Biomedical Research Excellence (COBRE) is providing raw anatomical and functional magnetic resonance imaging data from 72 patients with schizophrenia and 75 healthy controls (ages ranging from 18 to 65 in each group) \citep{calhoun2012exploring}. \\
The Glaucoma Grading from Multimodality Images (GAMMA) Challenge is intended to facilitate the development of fundus and OCT-based glaucoma grading \citep{wu2023gamma}. GAMMA contains 2D fundus images and 3D OCT images of 300 patients. \\
Computational Precision Medicine: Radiology-Pathology Challenge on Brain Tumor Classification 2020 (CPM-RadPath) is a brain tumor classification challenge. There are 221 cases in the training dataset, each with a paired radiology and digital pathology image. Within the 221 cases, there are 54, 34, and 133 cases for lower grade astrocytoma, IDH-mutant, oligodendroblioma, 1p/19q codeltion, and glioblastoma and diffuse astrocytic glioma with molecular features of glioblastoma, IDH-wildtype, respectively \citep{hsu2022weakly,kurc2020segmentation}. The CPM-RadPath 2020 challenge also contains 35 and 73 validation and testing sets, respectively. Each patient contains multiple MRI sequences: T1, post-contrast T1-weighted (T1Gd), T2, and FLAIR. \\
ISIT-UMR is a dataset for the classification of gastrointestinal lesions in regular colonoscopy. The dataset consists of 76 polyps with white light and NBI videos from the same polyp \citep{mesejo2016computer}. \\
The MRNet dataset consists of 1,370 knee MRI exams performed at Stanford University Medical Center between January 1, 2001, and December 31, 2012. There were 1,104 (80.6\%) abnormal exams in the dataset, with 319 anterior cruciate ligament (ACL) tears and 508 meniscal tears \citep{bien2018deep}. \\
CTU-CHB Intrapartum Cardiotocography is a database containing 552 cardiac tomography recordings from the Czech Technical University (CTU) in Prague and the University Hospital in Brno (UHB). As part of each CT, a fetal heart rate time series (FHR), as well as a uterine contraction (UC) signal, are recorded \citep{chudavcek2014open}.

The previously mentioned datasets provide valuable resources for developing and testing multimodal fusion methods. They contain images of different medical modalities of the same patient, as well as images of different patients. Access to these datasets is available upon request and at no cost. In this review, we summarize the fusion methods presented in \changed{53} articles that use ADNI, 11 articles that use TCIA, 7 articles that use BraTS (2015, 2017, \added{2019 and 2021} editions), \added{7 article that uses OASIS}, \changed{4} articles that use COBRE, \added{4 articles that use SPC, 4 articles that use ABIDE,} 3 articles that use ADHD-200, 2 articles that use CPM-RadPath (2020 edition), 2 articles that use GAMMA, 2 articles that use MRNet, 1 article that uses TCGA, 1 article that uses CTU-UHB and 1 article that uses ISIT-UMR. As mentioned earlier, 19 papers discussed in this review are not based on public datasets.

\section{Multimodal classification pipeline}
\label{sec: multimodal classification pipeline}

Multimodal fusion of biomedical data using deep learning remains an evolving field. The terminology used to describe fusion methods often varies between publications, leading to ambiguity. For instance, terms like input, intermediate, and output fusion are commonplace, but their interpretations may differ. To bring clarity and standardization to the multimodal classification area, we adopt the five-stage pipeline proposed in \citet{sleeman2022multimodal}, referenced in Tab.~\ref{tab_pipeline}. This pipeline offers a structured approach to encapsulate all medical multimodal classification tasks. Within this section, we elucidate each of these stages, detailing their definitions and the methodologies for their implementation. Subsequently, based on the sequence and structure of the information fusion stage paired with the deep learning (DL) backbone stage, we categorize multimodal fusion techniques into five distinct \changed{strategies} in Section \ref{sec: multimodal classification networks}.

\begin{table*}[!t]
\centering
\caption{Multimodal classification pipeline.}\label{tab_pipeline}
\begin{tabular}{|l|l|l|l|l|}
\hline
Stage & Description\\
\hline
\makecell[l]{Data preprocessing}& \makecell[l]{The initial step of the classification task is to perform operations such as\\ registration, denoising, and data augmentation on the raw data.}\\
\hline
\makecell[l]{DL backbone}& \makecell[l]{Extraction of high-dimensional features of data by the deep learning network structure.}\\
\hline
\makecell[l]{Information fusion}& \makecell[l]{Fusion of multimodal data/features by different methods.}\\
\hline
\makecell[l]{Final classifier}& \makecell[l]{The final stage of generating classification results from multimodal data.}\\
\hline
\makecell[l]{Model evaluation}& \makecell[l]{Different metrics are used to evaluate the performance of multimodal models.}\\
\hline
\end{tabular}
\end{table*}

\subsection{Pre-processing}

Image pre-processing is crucial for multimodal medical classification tasks, as it enhances DL network efficiency and effectiveness in extracting features. Pre-processing techniques, such as image registration, cropping, denoising, resampling, intensity normalization, regions-of-interest (ROI) extraction \citep{cuingnet2011automatic,davatzikos2011prediction,kohannim2010boosting}, and feature selection \citep{liu2012ensemble,liu2014hierarchical,akhavan2018combination,qian2020combined}, prepare the data for more accurate and efficient analysis by DL models.

To further improve the performance of these models, data augmentation techniques play an essential role in the pre-processing pipeline. For example, data augmentation helps prevent overfitting \citep{wang2018automated} using methods like random cropping, flipping, and rotation during training. In addition, increasing the training dataset's diversity improves the model's generalization capabilities.

Considering the large volumes of data generated by multimodal medical images, it is noteworthy that only a small fraction is relevant to diagnosing diseases. Therefore, feature selection emerges as a crucial pre-processing step, aiming to reduce data dimensionality while retaining pertinent information. Common feature selection methods include manual selection \citep{zou20173d,shi2017multimodal,kim2018identification} and Principal Component Analysis (PCA) \citep{li2015robust,el2020multimodal,zhou2021use}.

Another critical aspect of pre-processing multimodal medical images is image registration. It involves aligning images from different modalities (e.g., MRI, CT, and PET) into a common coordinate system,  enabling the accurate matching of corresponding anatomical structures across image types \citep{azam2021multimodal,el2016current}. Such alignment facilitates comprehensive data analysis and becomes particularly critical for input-level fusion, where combining complementary information from different modalities requires proper alignment \citep{li2022multimodal}.\\
Image registration in this context presents several challenges. A significant one is the lack of ample training datasets for supervised deep learning. Another is defining accurate similarity measures, especially with the varied appearance of different modalities. The registration process can be further complicated when trying to align images from different patients or even the same patient over time due to factors like changes in anatomy and metabolic processes.\\
In tackling these challenges, deep similarity metrics have shown promise, especially in traditional frameworks. While multimodal registration has seen advancements, direct transformation prediction lags behind, especially compared to single-modality methods. One innovative solution is using Generative Adversarial Networks (GANs) to make multimodal images more consistent.

\subsection{Information fusion}

\begin{figure}[!t]
\includegraphics[width=0.5\textwidth]{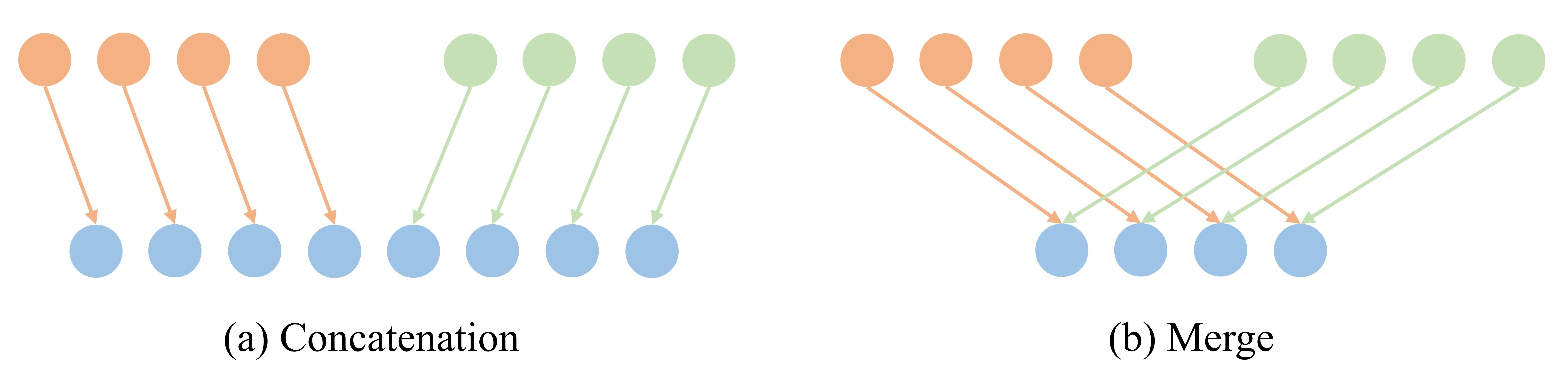}
\caption{Two types of fusion. Orange and green: data of different modalities. Blue: the output fused data.} \label{fig_fusion_ways}
\end{figure}

A key component of multimodal image classification is information fusion. Based on the level at which information is fused, information fusion can be divided into input fusion, intermediate fusion, and output fusion. And there are two ways to achieve fusion \citep{sleeman2022multimodal}, namely concatenation and merge. Concatenation involves the concatenation of data from different modalities into a single tensor for the next step. Merge involves complex calculations such as adding data from different modalities, and the final result is a smaller amount of data. Fig.~\ref{fig_fusion_ways} illustrates the two types of fusion. Our study focuses on the fusion of different medical imaging modalities, and in Sect.~\ref{sec: multimodal classification networks}, we will examine the different fusion methods in greater detail.

\subsection{Deep learning backbone}

DL backbones are used to extract high-dimensional features of modalities during the classification process. Over recent years, several high-performing network architectures have emerged, including AlexNet \citep{krizhevsky2017imagenet}, VGG \citep{simonyan2014very}, GoogLeNet \citep{szegedy2015going}, ResNet \citep{https://doi.org/10.48550/arxiv.1512.03385}, DenseNet \citep{https://doi.org/10.48550/arxiv.1608.06993}, AE \citep{suk2013deep,suk2015latent,yan2021richer}, ViT \citep{xing2022advit}, and others, providing state-of-the-art performance in classification. A summary of the common architectures for DL is presented in Tab.~\ref{tab_dlnet}. DL has developed rapidly due to several factors, including the development of hardware devices like graphics processing units (GPUs) and tensor processing units (TPUs), which have greatly improved the training speed of DL networks. Additionally, publicly available datasets such as ImageNet \citep{deng2009imagenet} have facilitated the training and testing of various models. Furthermore, DL is capable of learning advanced features directly from data without requiring extensive expertise or prior experience, making it easily adaptable across various domains.

\begin{table*}[!t]
\centering
\caption{Some common architectures of deep neural networks. Different architectures are more suitable for different types of data.}\label{tab_dlnet}
\begin{tabular}{|c|c|}
\hline
Architecture &  Description \\
\hline
\makecell[c]{Fully Connected Neural Network\\ (FCNN)} &  \makecell[c]{FCNN are the most traditional deep neural networks. \\Every neuron in a layer is connected to \\every neuron in the layer below it \citep{goodfellow2016deep}.}\\
\hline
\makecell[c]{Convolutional Neural Network\\(CNN)} & \makecell[c]{CNN can model spatial structures, \\such as images or volumes. \\Convolutional kernels model local information by \\sliding over input data \citep{goodfellow2016deep}.}\\
\hline
Autoencoders (AE) & \makecell[c]{By compressing and reconstructing the input data, \\AE learns low-dimensional encoding.\\ There are different types of layers, \\such as convolutional and fully connected \citep{ballard1987modular}.} \\
\hline
Transformer & \makecell[c]{Transformer is a model that uses \\a multi-headed attention mechanism.\\ Feature extraction is solely based on attention \citep{https://doi.org/10.48550/arxiv.1706.03762}.} \\
\hline
\end{tabular}
\end{table*}

In input fusion, a single backbone can extract features from fused modalities. However, in other fusion schemes such as intermediate or output fusion, multiple DL backbones may be used to extract features from different modalities. In current multimodal fusion research, Convolutional Neural Networks (CNN) are the preferred choice of the majority of researchers due to their effectiveness in feature extraction from medical images. Many pre-trained models have already been tested on large datasets, making them suitable for use in medical imaging research. In the articles analyzed, CNNs were used in 65 articles, Fully Connected Neural Networks (FCNN) in 10 articles, Auto-Encoders (AE) in 8 articles, and Transformers in 6 articles. 

\subsection{Final classifier}

Multimodal classification employs a final classifier to generate the classification results based on multimodal features or multiple independent classification results, depending on the employed fusion scheme. In DL networks, the Fully Connected (FC) layer \citep{akhavan2018combination,qin2020fine,liu2022attention,he2021hierarchical,li2022multimodal,elhabibdaho2023OMIA} is often used as the final classifier. Other methods, such as SVM \citep{li2015robust,suk2014hierarchical}, Random Forest \citep{dalmis2019artificial}, and Score Merge \citep{wang2022combining,hu2020deep} can also be used as final classifiers.  

\subsection{Evaluation metrics}

Evaluation metrics for multimodal fusion tasks are similar to those used in unimodal classification tasks. Commonly used indicators for assessing the performance of multimodal fusion methods and DL networks in the context of medical classification tasks include True Positive (TP), True Negative (TN), False Positive (FP), and False Negative (FN). These indicators can be used to calculate several performance metrics, such as sensitivity, specificity, accuracy, precision, and F1 score, among others. Additionally, AUC and Kappa are commonly used metrics to evaluate medical classification tasks.\\

\begin{itemize}
\item[-] Accuracy (ACC) = $\frac{TP + TN}{TP + TN + FP + FN} $
\item[-] Sensitivity (SEN) = $\frac{TP}{TP + FN} $
\item[-] Specificity (SPEC) = $\frac{TN}{TN + FP} $
\item[-] F1 Score = $\frac{2 \times TP}{2 \times TP + FP + FN} $
\item[-] Positive Predictive Value (PPV) = $\frac{TP}{TP + FP} $
\item[-] Negative Predictive Value (NPV) = $\frac{TN}{TN + FN} $
\item[-] Area Under the receiver operating characteristic Curve (AUC)
\item[-] Cohen's Kappa (Kappa) = $\frac{p_{0}-p_{e}}{1-p_{e}}$
\end{itemize}
\noindent where $p_{0}$ is the accuracy and $p_{e}$ the sum of the products of the actual and predicted numbers corresponding to each category, divided by the square of the total number of samples.

\section{Multimodal classification networks}
\label{sec: multimodal classification networks}

\subsection{Information fusion taxonomy for multimodal image classification}
\label{sec:Information fusion taxonomy for multimodal image classification}

\begin{figure}[!t]
\includegraphics[width=0.5\textwidth]{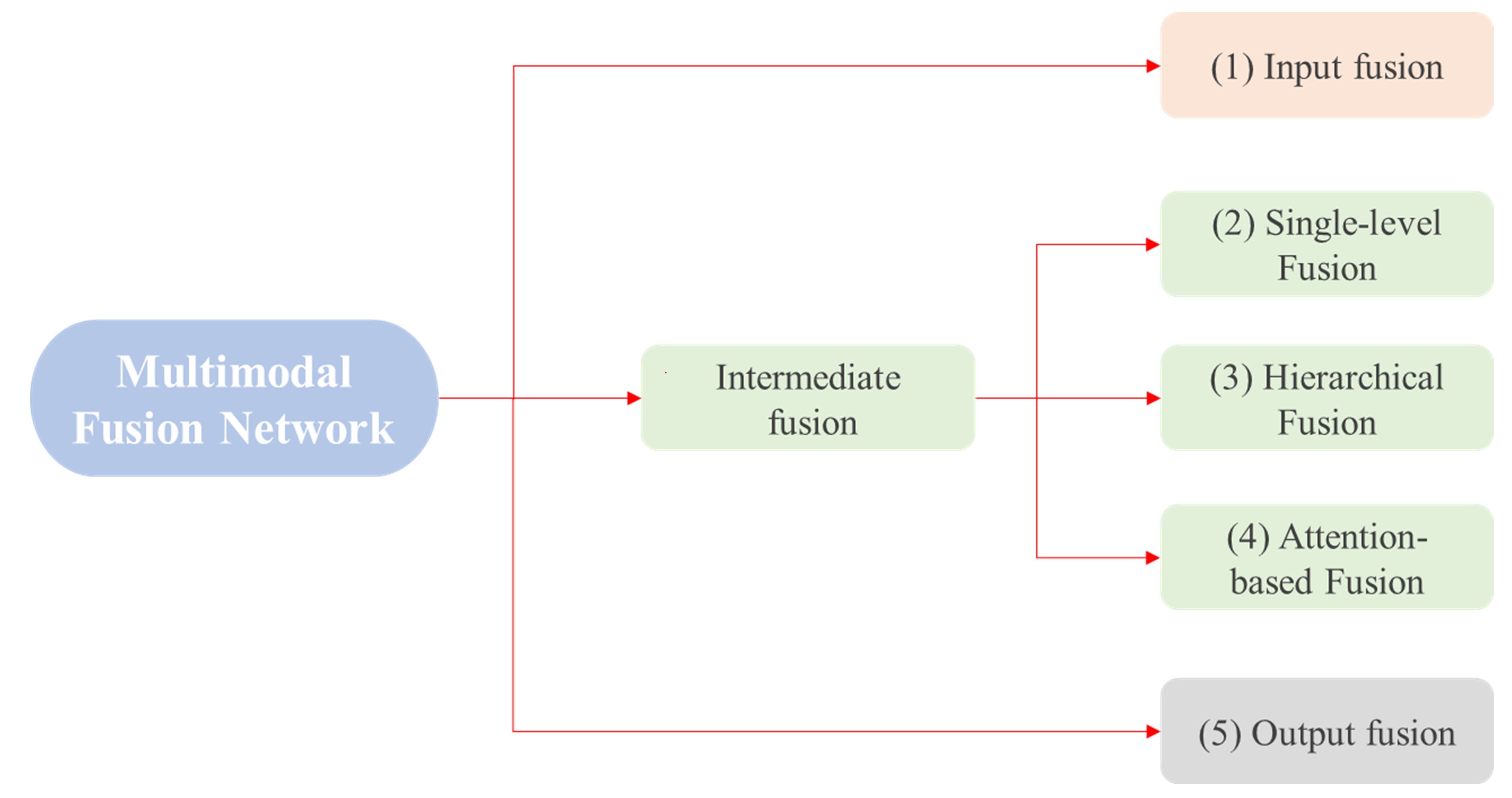}
\caption{Five types of multimodal fusion networks.} \label{fig_net}
\end{figure}

The positions of pre-processing and the final classifier are fixed during the process of multimodal classification. Based on the number and sequence of DL backbones and information fusion step, multimodal DL network architectures can be categorized into five types: input fusion, single-level fusion, hierarchical fusion, attention-based fusion, and output fusion, as shown in Fig.~\ref{fig_net}. As explained hereafter, single-level, hierarchical, and attention-based fusion are sub-categories of intermediate fusion. These categories describe how the network processes and combines the input modalities to produce classification results.

\begin{figure}[!t]
\includegraphics[width=0.5\textwidth]{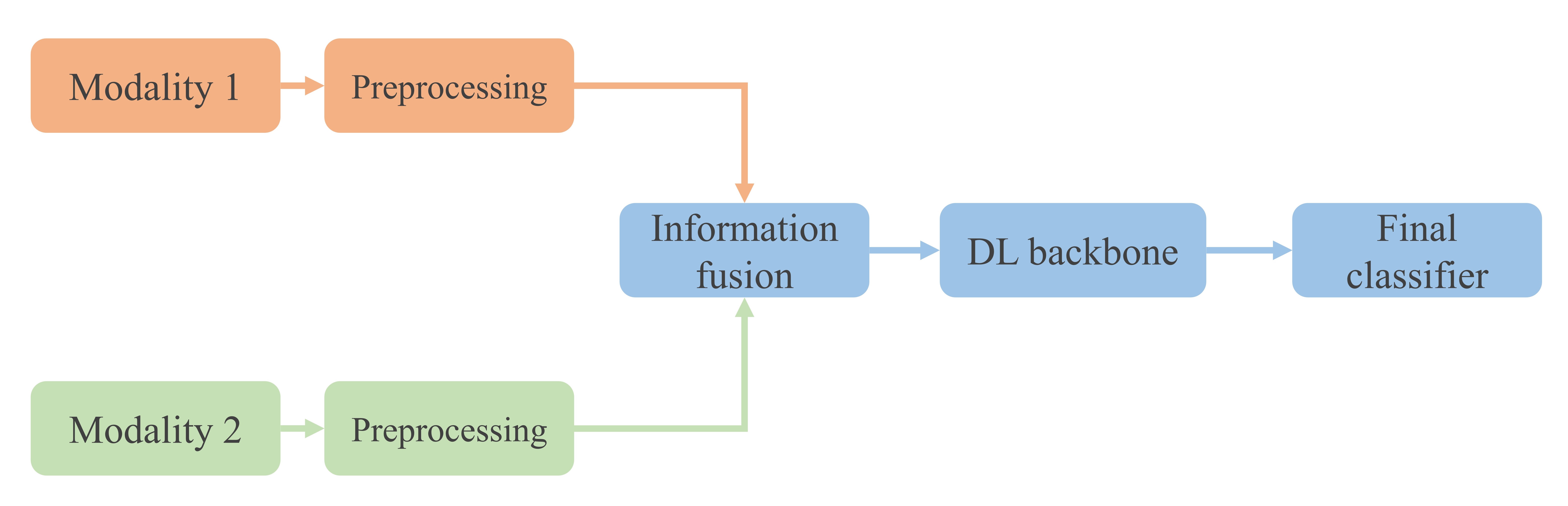}
\caption{Input fusion process diagram. Information fusion: Concatenation/Merge (Inputs).} \label{fig_early}
\end{figure}

(1) \textbf{Input fusion} can also be referred to as input-level fusion, where the information fusion phase precedes the DL backbone. Concatenation and Merge are two methods of information fusion. For the concatenation method, data of different modalities are used as different channels of the input. In the merge approach, data is fused at the pixel or voxel level, and the merged images are used as inputs for the DL classifier. The process diagram for input fusion is shown in Fig.~\ref{fig_early}.

\begin{figure}[!t]
\includegraphics[width=0.5\textwidth]{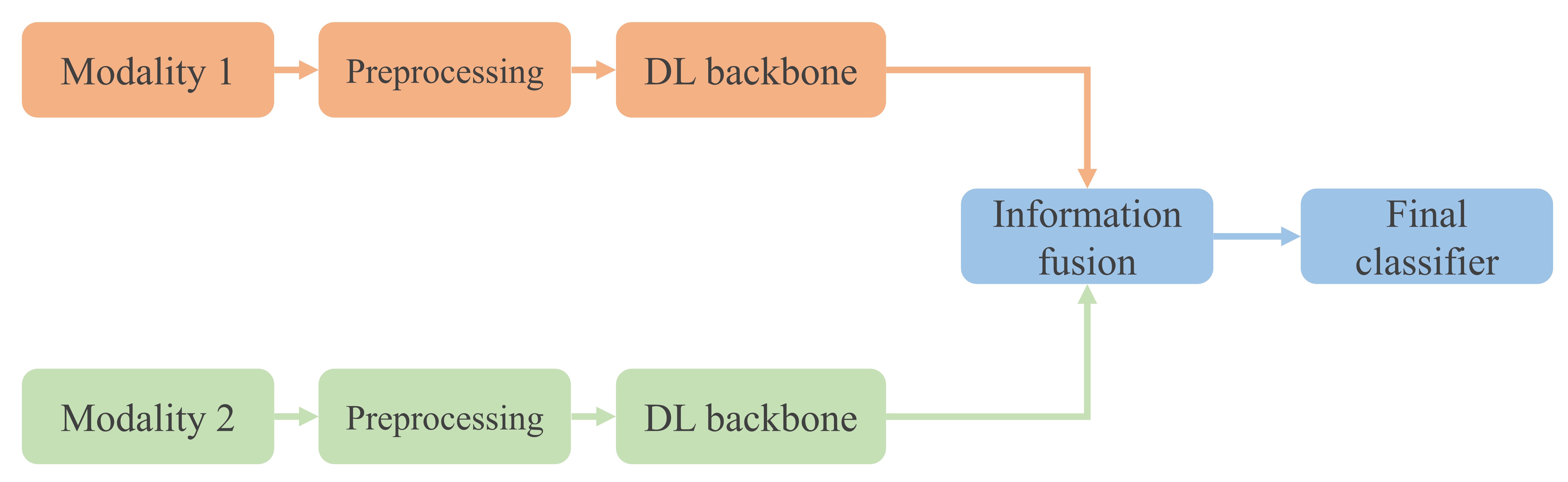}
\caption{Single-level fusion process diagram. Information fusion: Concatenation/Merge (Classic).} \label{fig_inter1}
\end{figure}

\begin{figure}[!t]
\includegraphics[width=0.5\textwidth]{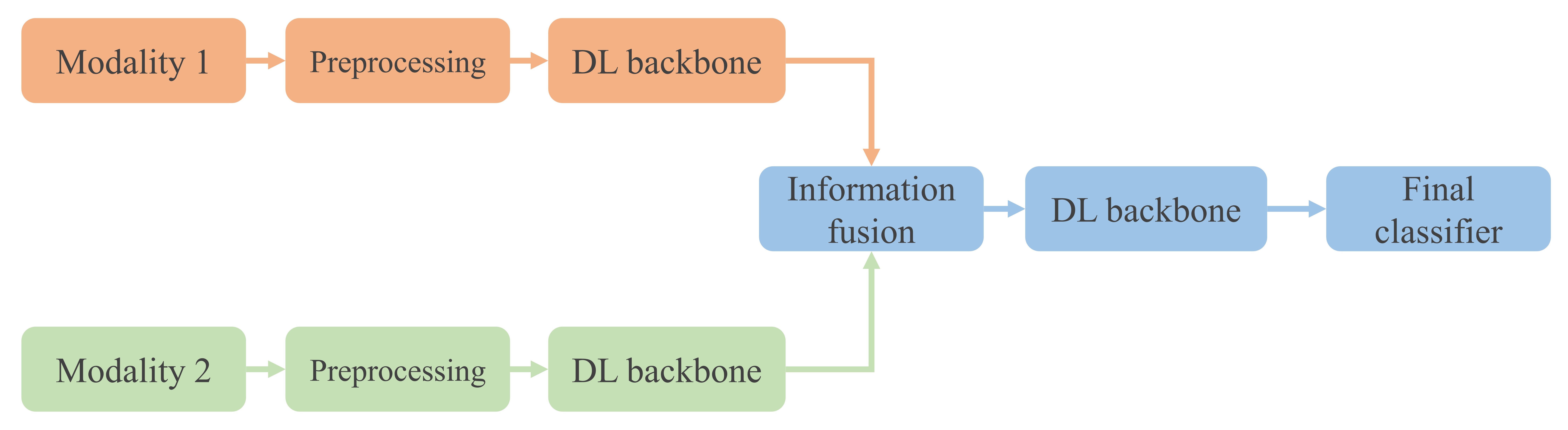}
\caption{Single-level fusion process diagram. Information fusion: Concatenation/Merge (Network).} \label{fig_inter2}
\end{figure}

(2) \textbf{Single-level fusion} involves information fusion after the DL backbone but before the final classifier. As part of a single-level fusion, the features extracted by the DL backbone are fused only once at some point before the classifier is applied. It can be divided into two types: Classic Fusion and Network Fusion, depending on the network structure. In Classic Fusion, high-dimensional features are extracted from different modalities using different DL classifiers and then merged or concatenated. This is the most common network structure in intermediate fusion, so we call it \textit{Classic}. Fig.\ref{fig_inter1} illustrates the process diagram of classic fusion. In network fusion, the intermediate features of different modalities are first extracted using DL classifiers, followed by the extraction of high-level features of the fused modalities using additional DL backbones. Fig.\ref{fig_inter2} shows the process diagram of network single-level fusion.

\begin{figure}[!t]
\includegraphics[width=0.5\textwidth]{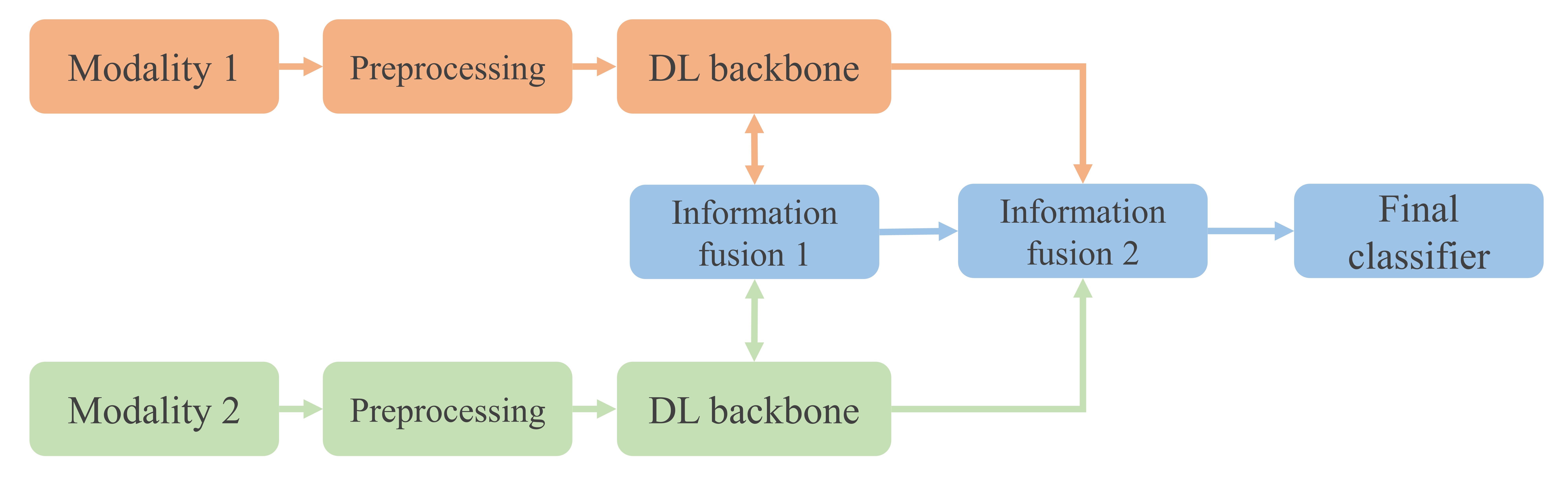}
\caption{Hierarchical fusion process diagram. Information fusion 1: Concatenation/Merge (Network). Information fusion 2: Concatenation/Merge (Classic).} \label{fig_hirar}
\end{figure}

(3) \textbf{Hierarchical fusion} is an improvement over single-level fusion. In this approach, DL backbone extracts features from the data of different modalities, while features from each level are then fused at the network level by concatenation or merging. Additionally, further feature fusion is performed following the DL backbone. This allows for more complex feature combinations to be learned, improving classification accuracy. The process diagram for output fusion is shown in Fig.~\ref{fig_hirar}.

\begin{figure}[!t]
\includegraphics[width=0.45\textwidth]{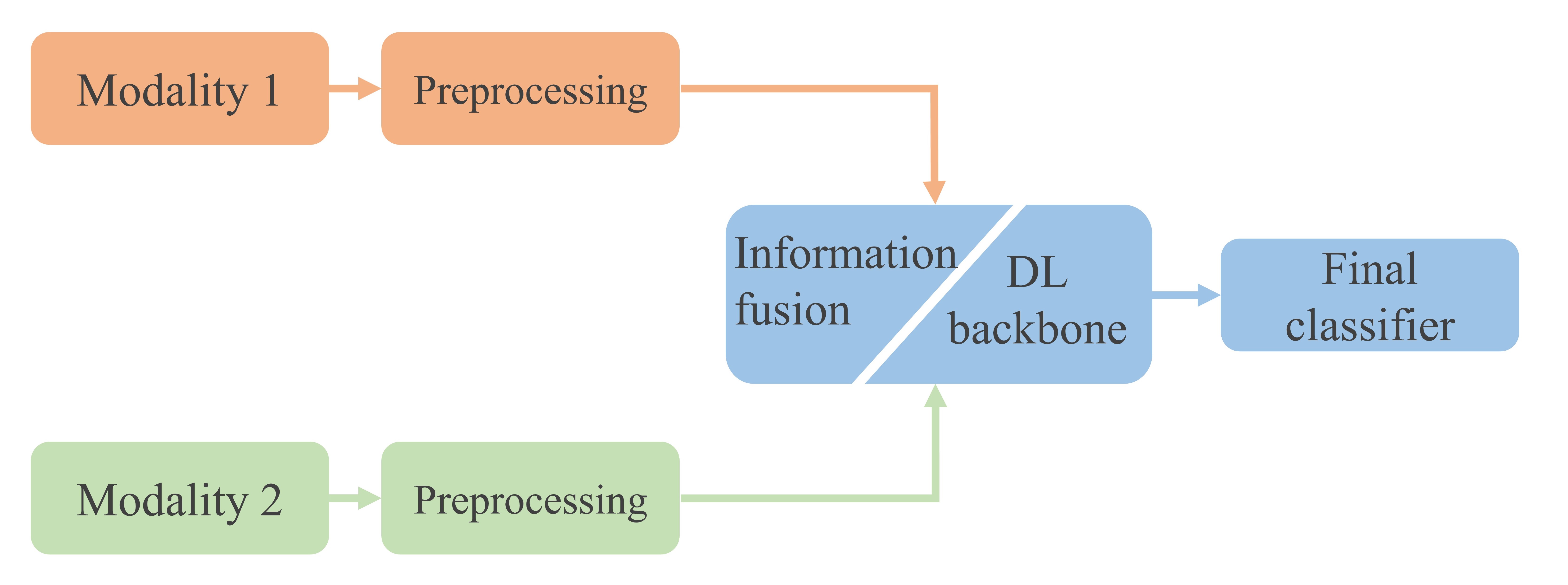}
\caption{Attention-based fusion process diagram.} \label{fig_att}
\end{figure}

(4) The emergence of Transformers has led to the development of \textbf{Attention-based fusion} as a new network architecture. Through its unique DL backbone, this architecture is able to extract features and implement feature fusion based on the attention relationship between different modalities. Fig.~\ref{fig_att} illustrates the process of attention-based fusion. A more detailed analysis of the network architecture is presented in Sect.~\ref{sec:Attention-based fusion networks}. 

\begin{figure}[!t]
\includegraphics[width=0.5\textwidth]{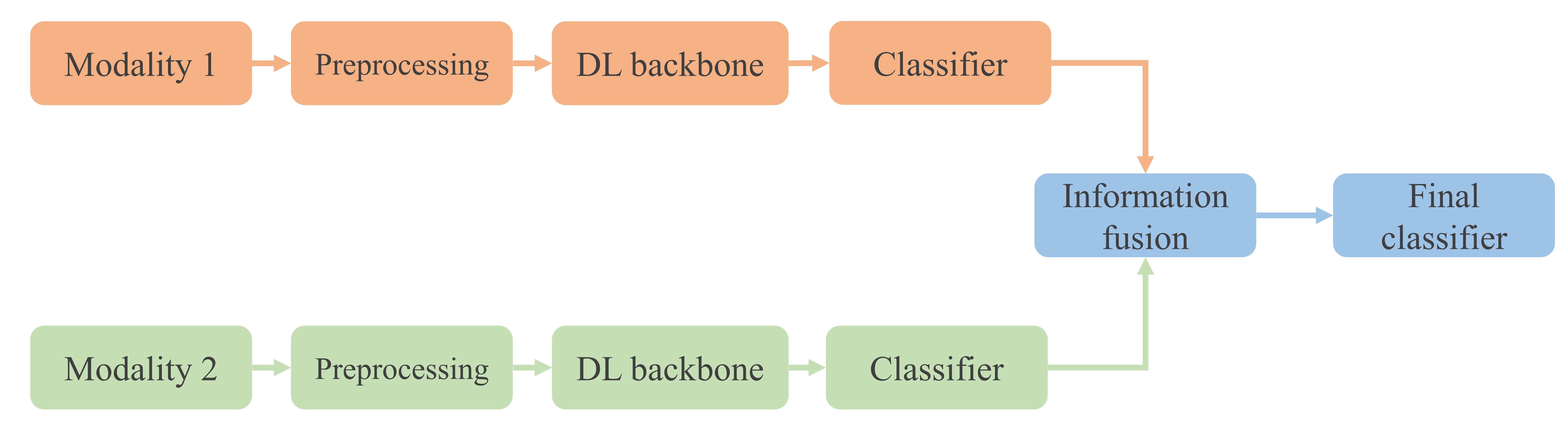}
\caption{Output fusion process diagram. Information fusion: Merge (Outputs).} \label{fig_late}
\end{figure}

(5) \textbf{Output fusion}, also known as decision-level fusion or late fusion, involves the use of DL backbones to extract high-dimensional features from different modalities of data. The extracted features are then used to generate separate classification results for each modality. These results are then combined using a fusion technique, such as majority voting or averaging, to produce a final classification result. The process diagram for output fusion is depicted in Fig.~\ref{fig_late}.

Recent years have seen a growing trend toward the use of deep learning networks in multimodal fusion research. Fig.~\ref{fig_overview_lit} illustrates the distribution of five fusion \changed{strategies} in the scope of the study. In contrast to traditional methods, single-level fusion is the most commonly used method in DL multimodal fusion, followed by input and output fusion. Hierarchical fusion and attention-based fusion are also gaining attention and present great potential for research. These more recent fusion methods offer more complex ways of combining modalities, enabling deep learning networks to learn more powerful representations of multimodal data.

\subsection{Input fusion networks}

Input fusion combines data from multiple modalities into a single feature tensor fed into the deep neural network as an input. Input fusion typically involves the fusion of modalities with similar structures, making implementation relatively straightforward. Some modalities can be acquired together at the time of clinical photography (e.g., CT and PET). In many cases, these modalities have the same voxels and spacing after data processing, making obtaining registered multimodal data easy. Furthermore, the majority of input fusion tasks do not require re-modeling, only modifying the input part of the unimodal model to achieve multimodality. Fusion can be accomplished in three ways: concatenating or merging multimodal medical images, extracting high-dimensional features from multimodal images, and then fusing them.

\begin{figure}[!t]
\includegraphics[width=0.5\textwidth]{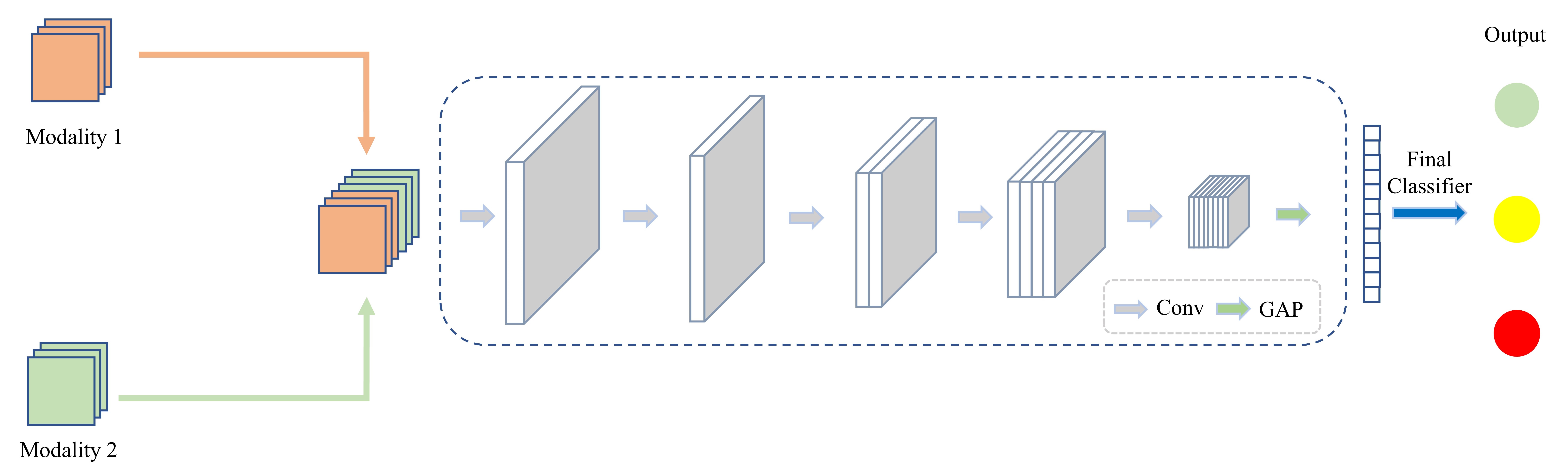}
\caption{Schematic diagram of the network architecture for input fusion. Information fusion method: Concatenation (Inputs).} \label{fig_early_ex}
\end{figure}

(1) The registered multimodal data are fed into the DL classifier as input for different channels to obtain classification results, which is the most common input fusion approach. Fig.~\ref{fig_early_ex} illustrates this typical input fusion network architecture used in the research of \citet{aldoj2020semi,lin2021bidirectional,zong2020deep,zhou2023prediction}. \citet{aldoj2020semi} proposed a semi-automatic method for the classification of prostate cancer without feature selection. Several combinations of 3D volumes (e.g., ADC, DWI, and T2) are utilized as inputs of the CNN network. Each sequence is considered an input channel; the output is the classification of significant versus nonsignificant lesions. \citet{lin2021bidirectional} employed MRI and PET to diagnose Alzheimer's disease. PET and MRI are used as two channels for the input of the CNN classification network, based on an ROI crop model to learn a classifier and fuse different features from MRI and PET. \citet{zong2020deep} concatenated T2, ADC and DWI for tumor foci classification using an end-to-end CNN network. In order to diagnose triple-negative breast cancer, \citet{zhou2023prediction} concatenated manually segmented multiparametric MRI images (PEI, DWI) into a CNN network. Despite the ease of implementing this fusion architecture, it has some limitations with regard to the modal data requirements. For instance, the registration performance of different modal data can influence the classification results. Moreover, this approach is not suitable for fusing heterogeneous data, such as 3D medical images and 1D clinical records, which have different characteristics and dimensions.

(2) The merging of images is another input fusion method in addition to concatenation. Various image modalities are fused at the pixel or voxel level in order to create a new fused image that is used for classification \changed{\citep{song2021effective,kong2022multi,rallabandi2023deep}}. \citet{song2021effective} proposed an effective multimodal image fusion method for Alzheimer's disease diagnosis using MRI and PET. Through registration and mask coding, they were able to fuse gray matter (GM) and 18-fluorodeoxyglucose positron emission tomography (FDG-PET) images to create a new imaging modality called "GM-PET".  In the resultant composite image, the GM area is clearly highlighted, allowing AD diagnosis to be made while maintaining both the contour and metabolic characteristics of the subject's brain tissue. They then fed the fused images to the CNN for classification. The GM region cropped from the MRI image is mapped onto the PET image, resulting in the fusion of PET and MRI data in \citet{kong2022multi} research. In addition to providing anatomical and metabolic information about the brain, the fusion modality also allows the viewer to focus on the main features of the brain by reducing the visual noise. \added{\citet{rallabandi2023deep} employed a fusion approach integrating images from MRI and PET for the diagnosis of Alzheimer's disease. The fusion process involved applying two-dimensional Fourier and discrete wavelet transform (DWT) to combine MRI and PET images. Subsequently, the MR-PET fused image was reconstructed using inverse Fourier and DWT methods.} The benefit of fused images is that they contain a wealth of medical information, but the process of generating them often requires an extensive amount of prior medical knowledge.

(3) Some studies have performed input fusion after extracting features from multimodal images instead of performing a direct fusion of medical images \citep{li2015robust,liu2014multimodal}. \citet{li2015robust} used PCA to extract features from MRI, PET, and cerebrospinal fluid (CSF) and then concatenated these features into the Restricted Boltzmann Machine (RBM) network for the diagnosis of Alzheimer's disease. \citet{liu2014multimodal} manually extracted features from MRI and PET and then used stacked auto-encoder (SAE) to classify the concatenated multimodal features in order to diagnose Alzheimer's disease. The architecture of extracting features and combining them can solve the problem of multimodal heterogeneity. However, PCA-based or manual feature extraction requires prior knowledge and does not fully utilize image information.

In input fusion, fused data is used in single-branch feature extraction, and the network architecture design significantly reduces network parameters and deployment difficulties. However, due to the fusion of the data at the input level, the complementary information from the different modalities is not utilized to the fullest extent possible.

\subsection{Single-level fusion networks}

The single-level fusion process uses different DL backbones to extract features from different modalities separately, followed by an information fusion process before making the final decision. Based on the position of information fusion within the network architecture, it can be divided into classic fusion structures and network fusion structures.

\begin{figure}[!t]
\includegraphics[width=0.45\textwidth]{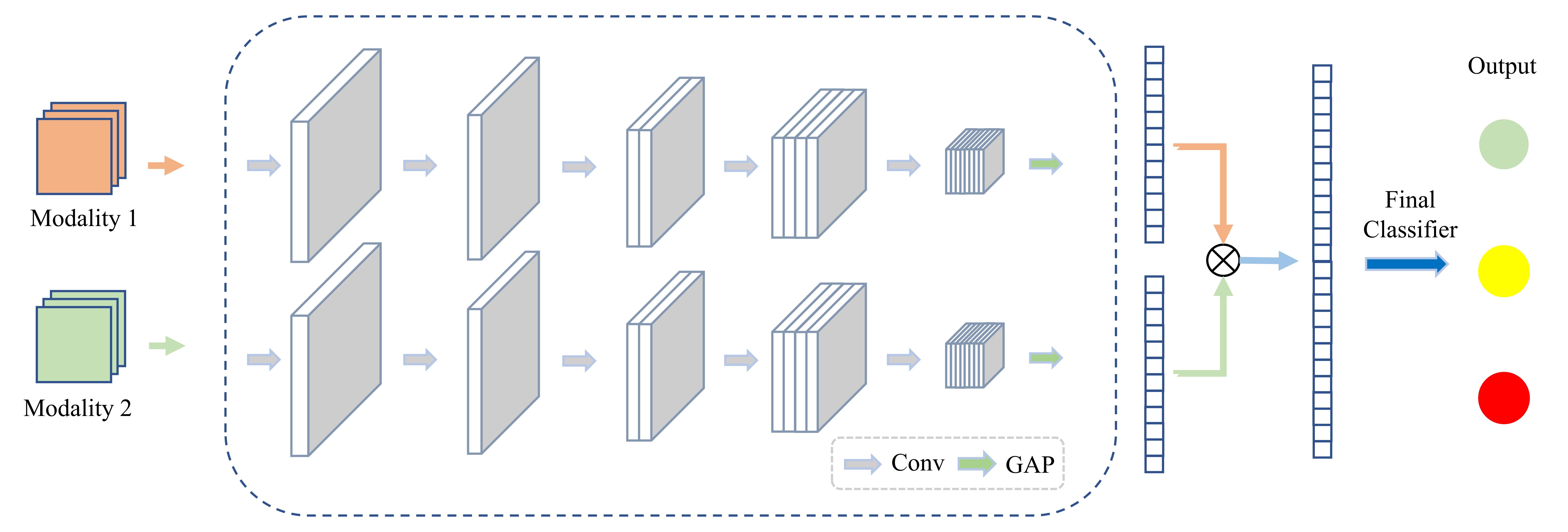}
\caption{Schematic diagram of the network architecture for classic single-level fusion. Information fusion method: Concatenation (Classic).} \label{fig_inter1_ex}
\end{figure}

(1) The most common single-level fusion architecture is to extract features from multimodal data by using different branches, then fuse these features and feed them to the final classifier \changed{\citep{suk2013deep,suk2014hierarchical,suk2015latent,xu2016multimodal,zou20173d,yang2017co,joo2021multimodal,ye2017glioma,punjabi2019neuroimaging,yap2018multimodal,rahaman2021multi,xiong2022multimodal,qin2020fine,liu2023improving,kollias2023btdnet,kadri2023efficient,saponaro2024deep}}. A schematic diagram of its network architecture is shown in Fig.~\ref{fig_inter1_ex}. After preprocessing the data, the architecture \citep{zou20173d} extracted low-level 3D features from fMRI and sMRI to classify Attention Deficit Hyperactivity Disorder (ADHD) automatically. As soon as the features are concatenated, softmax classifiers are used to differentiate ADHD cases from typically developing children (TDC) cases. In order to diagnose breast cancer, \citet{joo2021multimodal} fused MRI (T1, T2) and clinical information. Two 3D ResNet-50 were used to extract features from contrast-enhanced T1 subtraction MR images and T2 MR images, while the FC layer provided clinical inputs. For the prediction of pathological complete response, the outputs of each 3D ResNet-50 and FC layer were concatenated, and the final FC layer with sigmoid activation function was used. Likewise, 
\citet{yap2018multimodal} employed ResNet and FC layers to extract features from DSC, Clinical Image, and metadata, then applied FC layers for skin lesion classification. Aside from these methods of concatenating modal features, complex computations can also be used to merge features. \citet{xiong2022multimodal} used visual field (VF) and OCT for the diagnosis of glaucoma. VFNet and OCTNet were used to extract features from the VF and OCT modes, respectively. A weighted average was used to obtain an aggregated representation from bimodal features using an attention module. Each modal feature was assigned a weight using a fully connected layer, followed by a sigmoid function to calculate a scalar value (0-1) indicating the feature's relative contribution to the aggregate representation. To aggregate all features, a global average pooling layer was also used. The results of glaucoma diagnosis were predicted using three FC layers and a softmax layer. For CT and PET modalities, \citet{qin2020fine} extracted features using CNN networks, merged the features using gated multimodal units (GMU), and classified lung cancer using FC layers. GMU, unlike the widely used connection operation, allows for the learning of intermediate representations of multimodality features using hidden structures and gate controls, thus enabling the prediction layer to assign weights more effectively to intrinsically associated features.

\begin{figure}[!t]
\includegraphics[width=0.45\textwidth]{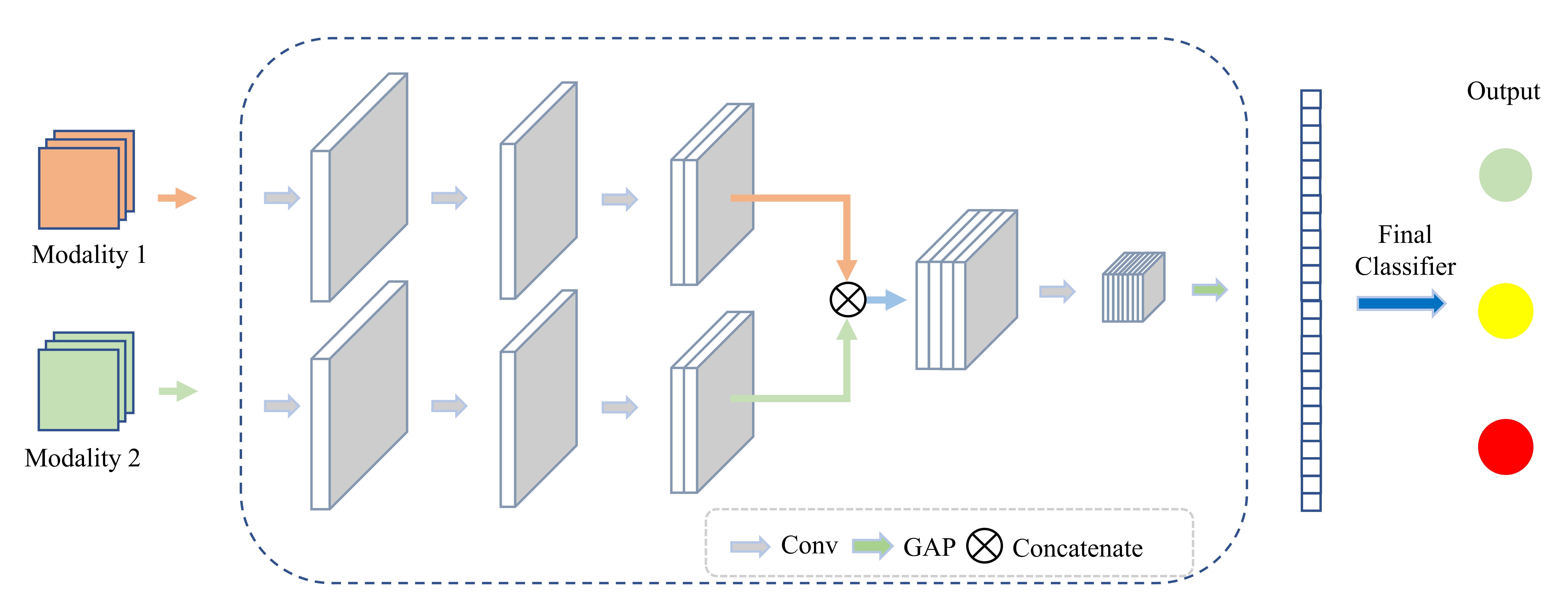}
\caption{Schematic diagram of the network architecture for single-level network fusion. Information fusion method: Merge (Network).} \label{fig_inter2_ex}
\end{figure}

(2) Two stages can be described as the single-level fusion architecture for network fusion. The first stage involves extracting single-level features separately from different modalities using DL backbones, followed by the second stage of information fusion which involves utilizing an additional DL backbone to extract high-level features from the fused features \changed{\citep{shi2017multimodal,zhou2017feature,cheng2017cnns,kim2018identification,rahaman2022two,jin2022hybrid,leng2023multimodal,lu2024hierarchical}}. Lastly, the extracted high-level features are used in the final classification process. Fig.~\ref{fig_inter2_ex} illustrates a typical network fusion architecture. \citet{cheng2017cnns} used cascaded CNN for the multimodal fusion of MRI and PET to diagnose Alzheimer's disease. They proposed a 2D CNN to combine the multimodality features and make the final classification. After 3D CNN output features are flattened to one dimension, the 1D feature vectors of MRI and PET are combined to produce a two-dimensional feature map for 2D CNN analysis. \citet{kim2018identification} developed a multimodal architecture for combining MRI, PET, and CSF features. Each modality's individual representation of high-level features is calculated using the stacked sparse extreme learning machine auto-encoder (sELM‐AE). Another stacked sELM‐AE is used to get the joint features from the high‐level MRI, PET, and CSF features. The kernel-based extreme learning machine classifies the joint feature representation. With multimodality neuroimaging and genetic data, \citet{zhou2019effective} proposed a three-stage deep feature learning DNN framework for Alzheimer's disease classification. Each modality's latent representation is learned in the first stage, then each pair of modalities' joint latent representations are learned in the second stage, and in the third stage, each pair of modalities' joint latent representations are used to create the classification model. \citet{rahaman2022two} classified schizophrenia using sMRI, fMRI, and single nucleotide polymorphisms.  The latent representations for the static functional network connectivity (sFNC), sMRI, and single nucleotide polymorphism (SNP) are learned using an autoencoder, multi-layered perceptron, and bi-directional long short-term memory (LSTM). The Multimodal Bottleneck Attention Module performs the fusion of the embeddings and then sends the combined embeddings to a variational autoencoder for encoding, followed by a Softmax layer for classification.

The single-level fusion method is currently used to merge multiple medical modalities for classification tasks and can be applied to the fusion of different medical modalities. The method does not require a specific format for the data as it extracts features from modalities using different branches and fuses data at a high-dimensional feature level. In this regard, single-level fusion is a suitable solution for unregistered or different dimensional data. Due to the fact that information fusion occurs only at the end of the network architecture, single-level fusion is not capable of analyzing low-dimensional features jointly.

\subsection{Hierarchical fusion networks}

Hierarchical fusion extends single-level fusion further in order to further exploit the complementary information between multimodal data. The hierarchical fusion process involves the fusion of different dimensional features and the classification of these jointly represented features through the process of fusion \changed{\citep{mahmood2018multimodal,zhang2020multi,zhou2021use,he2021hierarchical,li2022multimodal,wu2023aggn,omeroglu2023novel,xu2023multi,tu2024multimodal,miao2024mmtfn,xu2024cross}}. There are two ways to implement hierarchical fusion: by using additional branches for multimodal feature fusion or by using fusion blocks for joining features from different modalities.

\begin{figure}[!t]
\includegraphics[width=0.5\textwidth]{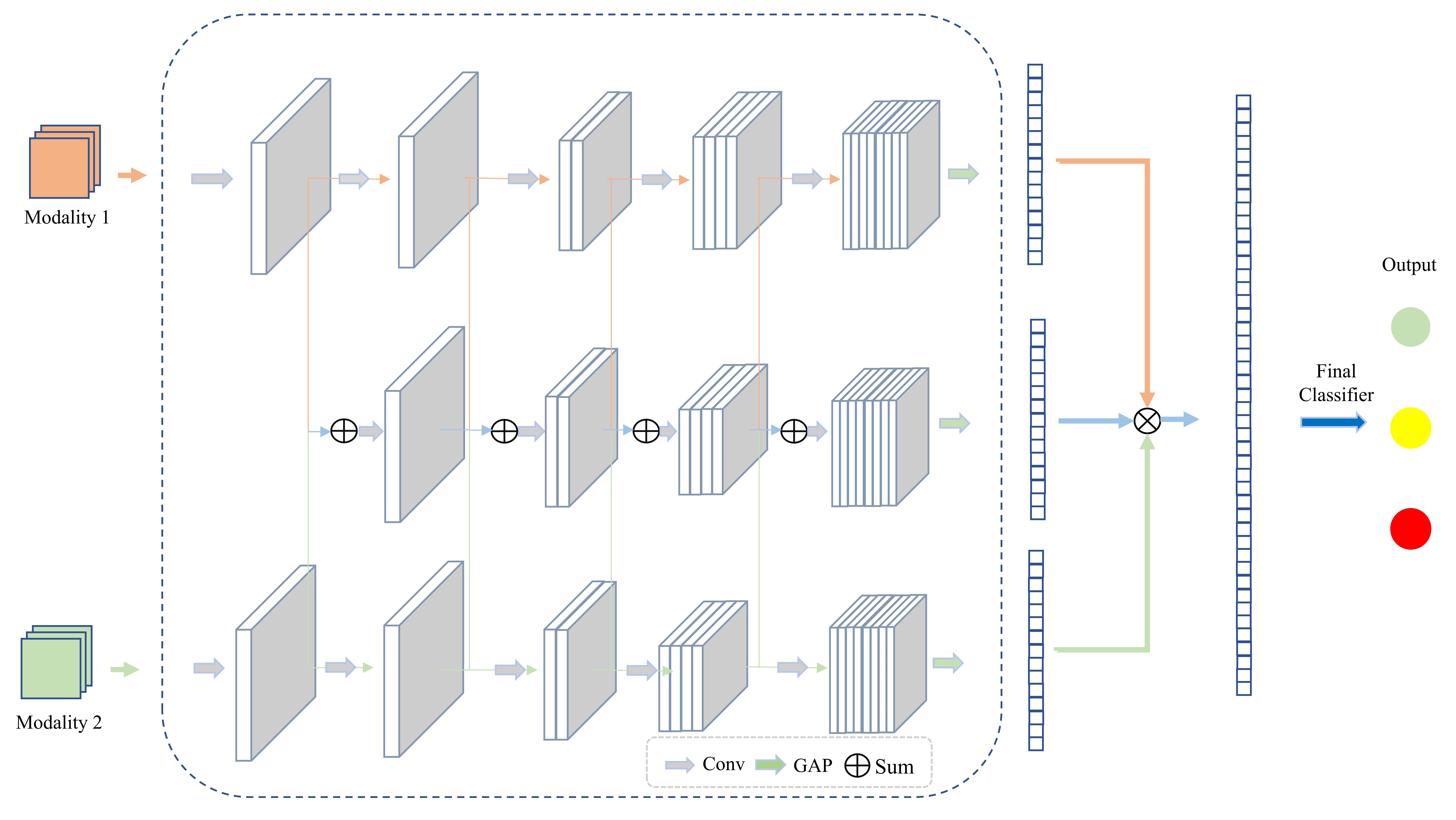}
\caption{Schematic diagram of the network architecture for hierarchical fusion. Information fusion method: Merge (Network) and Concatenation (Classic).} \label{fig_hirar_ex}
\end{figure}

(1) The common hierarchical fusion architecture involves extracting different modalities via different branches and simultaneously combining multimodal features of different dimensions via another parallel branch. Finally, the high-dimensional features from the fusion branch and each modal branch are combined for classification. Fig.~\ref{fig_hirar_ex} shows a typical network architecture for hierarchical fusion,  \citet{zhou2021use,zhang2020multi,li2022multimodal} utilized this network architecture. \citet{zhou2021use} utilized three sparse-response Deep Belief Network (DBN) branches to extract features from PET/MRI modalities, fuse them, and then employed an Extreme Learning Machine (ELM) to classify the fused features for brain diseases. \citet{zhang2020multi} used a deep multi-modal fusion network (DMFNet) to fuse PET and MRI data for the diagnosis of Alzheimer's disease. Three branches are present in DMFNet, two of which extract features from the MRI and PET scans, respectively. A channel attention model is used to extract the features from each branch and merge the reweighted feature maps. In the third branch, fused features are further extracted. \citet{li2022multimodal} used a three-branch CNN network to combine 2D fundus images with 3D OCT images in order to classify glaucoma and diabetic retinopathy. The fusion of 2D and 3D data features on the fusion pointers was achieved by changing the dimensionality of the data features using a transformation layer.

\begin{figure}[!t]
\includegraphics[width=0.5\textwidth]{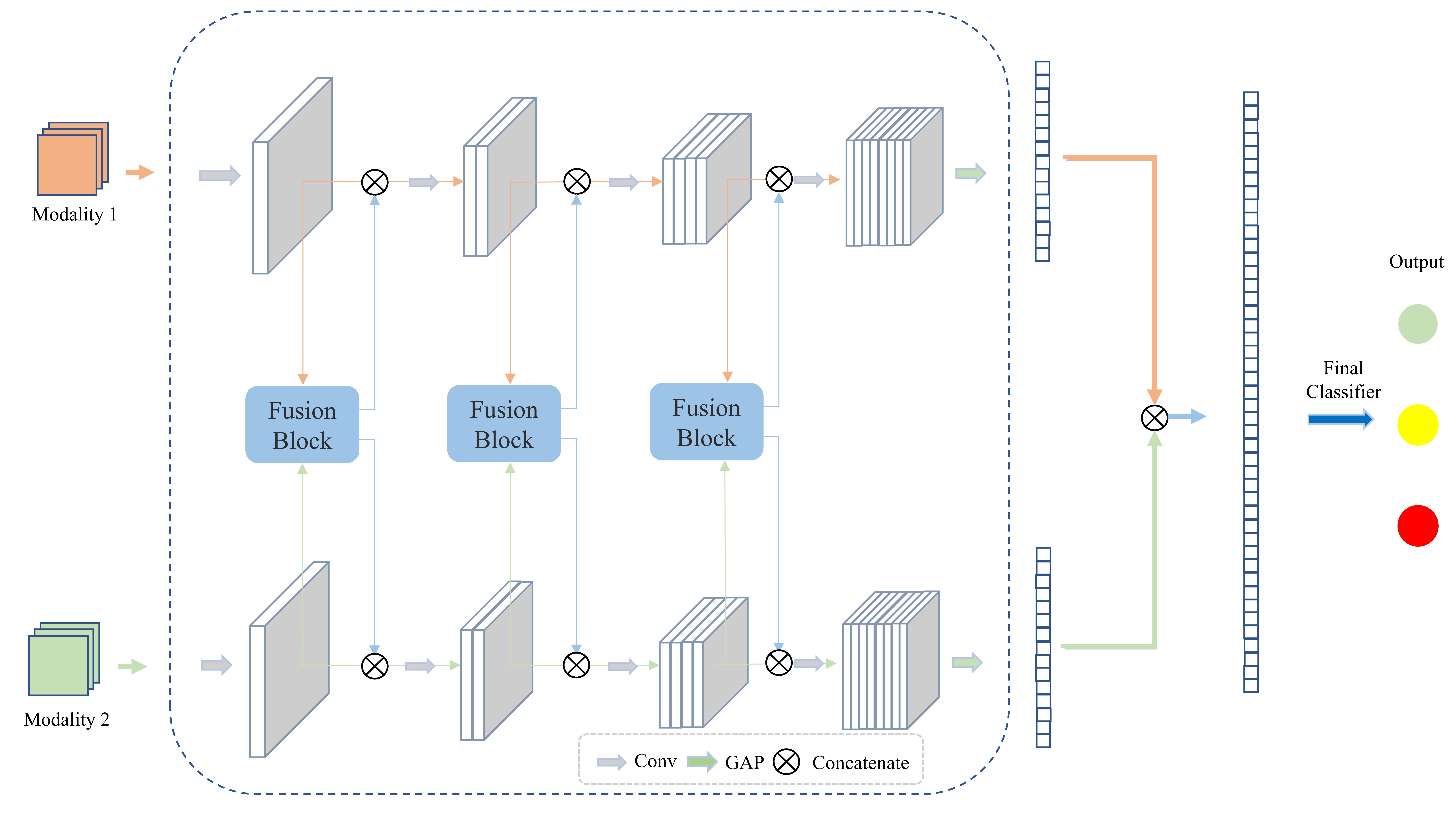}
\caption{Schematic diagram of another network architecture for hierarchical fusion. Information fusion method: Merge (Network) and Concatenation (Classic).} \label{fig_hirar_ex2}
\end{figure}

(2) Hierarchical fusion can also be structured in another way by extracting features using different branches and fusing them in different dimensions by using fusion blocks, which are then returned to each modality branch for further fusion. The design of such a network structure can reduce the number of model parameters while fusing features at multiple levels. Fig.~\ref{fig_hirar_ex2} illustrates a typical network architecture, \citet{gao2021task,he2021hierarchical} utilized this network architecture. To classify brain diseases, \citet{gao2021task} proposed a pathwise transfer deep convolution network that gradually learned and combined the multi-level and multimodal features of MRI and PET. The pathwise transfer blocks are designed to fully utilize complementary information from different imaging modalities. Pathwise transfer blocks are used to communicate information across PET and MRI, which helps to improve the classification model's performance. \citet{he2021hierarchical} proposed a multimodal MRI hierarchical-order multimodal interaction fusion network (HOMIF) to diagnose gliomas. There are two branch networks for each modality, several multimodal interaction modules with different scales and orderings, diverse learning constraints, and a predictive subnet in the framework. Each branch network has three CNN blocks with multiscale inputs and an arm with diverse high-order multimodal interaction (HOMI) modules to integrate and interact deeply with the multiscale features.

The multi-level feature fusion allows hierarchical fusion to explore more fully the complex and complementary information between modalities. Learning the synergy of multimodal data while maintaining the features of the modalities improves the model's classification performance \citep{zhang2020multi}. However, as it involves the fusion of low-dimensional features, the registration of multimodal data may affect the classification performance of hierarchical fusion.

\subsection{Attention-based fusion networks}
\label{sec:Attention-based fusion networks}

As attentional mechanisms \citep{vaswani2017attention} have been proposed and developed, more and more studies are beginning to incorporate attentional mechanisms into network architectures. Some of the network architectures mentioned above also included attention mechanisms in order to enhance the performance of the models. \citet{zhang2020multi} added attention modules to reweight the modal features. \citet{xing2022advit} use a vision transformer (ViT) to extract the modal features and fuse them. These studies, however, only operate on unimodal modalities and do not utilize the attention mechanism for multimodal interactions. Recently, some studies have used the attention mechanism to extract and combine features \changed{\citep{dai2021transmed,qiu2023hierarchical,zhang2022multimodal,liu2023cascaded,li2022attention,dai2022mutual,zuo2023alzheimer,chen2023multimodal,gao2023multimodal,bi2023multimodal,wei2024multi}}. This network architecture is called attention-based fusion, which is not related to any of the previous fusion architectures.

In the study of \citet{dai2021transmed}, they propose TransMed, which combines CNN and transformer to capture high-level cross-modalities and low-level features. First, TransMed sends the multimodal images to CNN, where they are processed as sequences, then transformers learn the relationships between them and predict the end result. TransMed is more efficient and accurate than existing multimodal fusion methods because it effectively models the global features of multimodal images.

Attention-based Hierarchical Multimodal Fusion (AHM-Fusion) is a novel fusion module \citet{qiu2023hierarchical} designed. The system includes both an early feature guidance module and a late feature fusion module, capturing deep interaction information between different multimodal features. In the early stage of feature aggregation, the early feature guidance module is used to capture multimodal interactions. To obtain classification results, late feature fusion modules based on attention mechanisms are used. Through cascading double attention layers in the late feature fusion module, the deep interaction information is further captured. Then, they used a gating-based attention mechanism to decrease the impact of insignificant features in each modality.

\citet{zhang2022multimodal} proposed a multimodal Medical Information Fusion (MMIF) framework that combines the Category Constrained-Parallel ViT framework (CCPViT) and the multimodal Representation Alignment Network (MRAN) as backbones, enabling the modeling of images and texts as unimodal features, as well as cross-modal features. CCPViT is proposed as a tool for learning key features of different modalities and for solving unaligned multimodal tasks. hen in MRAN, Cross-attention was used to cascade encoded images and decoded texts to explore deep-level interactive representations of cross-modal data, assisting with modal alignment and identifying abnormalities. MMIF is an image-text foundation modeling that could contribute to a much higher-precision classification model when compared with unimodal models.

Multimodal Mixing Transformer (3MT) was presented \citet{liu2023cascaded} as a novel technique to classify diseases. Based on neuroimaging data, gender, age, and the Mini-Mental State Examination (MMSE), They tested it for Alzheimer's Disease classification. Multimodal information is incorporated through a Cascaded Modality Transformers architecture with cross-attention. Different embedding layers are used to obtain Key (K) and Value (V) from imaging features and clinical data. K and V are then placed into a cross-attention layer with a latent code known as Query (Q). 3MT allows mixing an unlimited number of modalities and formats and full data utilization. 

\added{\citet{zuo2023alzheimer} has introduced a novel Swapping Bi-Attention Mechanism (SBM) designed for the diagnosis of Alzheimer's disease through the amalgamation of structural-functional brain images. The proposed model capitalizes on the transformer's bi-attention mechanism to explore mutually beneficial information inherent in both structural and functional images. Historically, transformers have been investigated solely within the confines of single-modality brain regions, neglecting the potential for leveraging complementary information across modalities. SBM, however, implements token exchange between the two modalities and adaptive fusion of intermediate features during the application of the Transformer for feature extraction from diverse modalities. This process facilitates the collaborative exchange of information embedded in bimodal images, enhancing the transparency of feature alignment and fusion. The incorporation of SBM results in a more lucid understanding of the influence of different modalities on feature extraction and multimodal fusion.}

Research is increasingly incorporating attention mechanisms, particularly Transformer structures, into multimodal classification tasks. While performing cross-modal attention computation, a multi-level fusion of multimodal features is achieved. Furthermore, the Transformer structure is well-suited for joining modalities of different dimensions. Nevertheless, Transformer research in medical tasks is still in its infancy, and various studies tend to focus on solving particular problems, making it difficult to conclude a general multimodal classification architecture. A further important point to be noted is that while the success of Transformer is accompanied by pre-training on large datasets, the number of samples in medical datasets is often not sufficient to achieve the good training effect of Transformer. As a result, it is recommended that Transformer and CNN are used together in a hybrid fashion.

\subsection{Output fusion networks}

\begin{figure}[!t]
\includegraphics[width=0.5\textwidth]{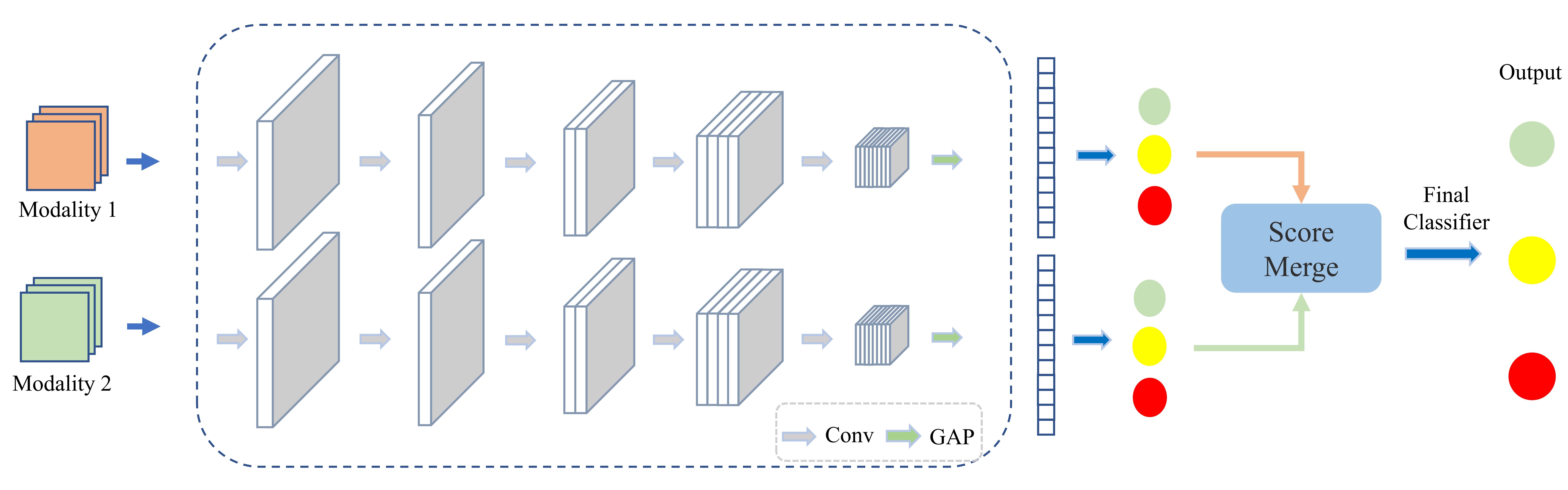}
\caption{Schematic diagram of the network architecture for output fusion. Information fusion method: Merge (Outputs).} \label{fig_late_ex}
\end{figure}

In output fusion, each modality uses a separate DL backbone to extract features and make decisions, and the results are merged into one final decision. Fig.~\ref{fig_late_ex} shows a typical network architecture for output fusion. The final Classifier of decision fusion can be achieved by simple operations \citep{moon2020computer,guo2022multimodal} such as voting, weighting, and averaging, or by classifiers \changed{\citep{abdolmaleki2020brain,fang2020ensemble,yoo2022deeppdt,kwon2022diagnosis,qiu2022multimodal}} such as SVM, extreme gradient boosting (XGBoost), adaptive boosting (AdaBoost), \added{Categorical Boosting (CatBoost),} Decision Tree, and K-nearest neighbor (KNN). \citet{moon2020computer} used unweighted average, weighted average, weighted voting, and stacking to fusion the classification results from different modalities of the US to identify breast tumors.  \citet{guo2022multimodal} applied a linear weighted module to assemble the predicted probabilities of the pre-trained models based on the 4 MRI modalities for the classification of gliomas. In order to achieve the diagnosis of early glottic cancer, \citet{kwon2022diagnosis} used decision trees to combine the classification results from the sound data and the image data. \citet{abdolmaleki2020brain} used SVM, KNN, and linear discriminant analysis (LDA) to fuse the classification results of fMRI and sMRI to diagnose ADHD.

The output fusion process involves combining unimodal results from different modalities. As a result, it is relatively easy to implement and generally does not require additional training. It is, however, difficult to exploit the complementary information between different modalities because there is no feature fusion. Furthermore, output fusion may not improve classification performance if there are large differences in classification performance between different modalities.

\section{Discussion}
\label{sec: Discussion}

\subsection{Which fusion method is the best?}

The choice of a fusion method is crucial when dealing with multimodal medical classification problems. Fortunately, many fusion architectures have been evaluated on the same dataset: ADNI. Based on quantitative results reported by the authors, comparisons between fusion architectures is possible to some extent. We consider studies performed on ADNI where MRI and PET are used to diagnose Alzheimer's disease. There are three stages in the progression of Alzheimer's disease: normal cognition (NC), mild cognitive impairment (MCI), and Alzheimer's disease (AD). In spite of the fact that MCI does not significantly interfere with daily activities, a high risk of AD progression has been consistently demonstrated in patients with MCI \citep{dubois2007research}. MCI subjects can be classified into MCI converters (cMCI) and MCI non-converters (ncMCI) to predict the transition risk of MCI. Tab.~\ref{tab_results} reports the results of the different fusion methods obtained by their authors for different classification tasks. The experiments were not replicated: when comparing these results, it should be noted that each paper relies on a different subset of patients, although the number of subjects was similar.

\begin{table*}[!htb]
\caption{Comparison of the results of different fusion methods on ADNI dataset. In the multi-classification task, 3 classes is NC vs. MCI vs. AD and 4 classes is NC vs. ncMCI vs. cMCI vs. AD. Unit:\%.}\label{tab_results}
\centering
\begin{tabular}{|l|l|l|l|l|c|c|}
\hline
Research &  Year & \makecell[c]{Fusion \\Methods} & \makecell[c]{Dataset} & NC vs. AD & NC vs MCI & \makecell[c]{Multi-classification}\\
\hline
\makecell[c]{\cite{liu2014multimodal}} & 2015 & \makecell[c]{Input \\ Fusion} & \makecell[l]{331 subjects: \\77 NC, \\102 ncMCI, \\67 cMCI, \\85 AD} & \makecell[c]{ACC: 91.40\\SEN: 92.32\\SPE: 90.42} & \makecell[c]{ACC: 82.10\\SEN: 60.00\\SPE: 92.32} & \makecell[c]{4 Classes\\ACC: 53.79 \\SEN: 52.14\\SPE: 86.98}\\
\hline
\makecell[c]{\cite{song2021effective}} & 2021 & \makecell[c]{Input \\ Fusion} & \makecell[l]{381 subjects: \\126 NC, \\160 MCI, \\95 AD} & \makecell[c]{ACC: 94.11\\SEN: 93.33\\SPE: 94.27} & \makecell[c]{ACC: 85.00\\SEN: 84.69\\SPE: 85.60} & \makecell[c]{3 Classes\\ACC: 71.52 \\SEN: 55.67\\SPE: 83.40}\\
\hline
\makecell[c]{\cite{kong2022multi}} & 2022 & \makecell[c]{Input \\ Fusion} & \makecell[l]{370 subjects: \\130 NC, \\129 MCI, \\111 AD} & \makecell[c]{ACC: 93.21\\SEN: 91.43\\SPE: 95.42} & \makecell[c]{ACC: 86.52\\SEN: 94.34\\SPE: 81.64} & \makecell[c]{3 Classes\\ACC: 87.67}\\
\hline
\makecell[c]{\cite{suk2014hierarchical}} & 2014 & \makecell[c]{Single-level \\ Fusion} & \makecell[l]{398 subjects: \\101 NC, \\128 ncMCI, \\76 cMCI, \\93 AD} & \makecell[c]{ACC: 95.35\\SEN: 94.65\\SPE: 95.22} & \makecell[c]{ACC: 85.67\\SEN: 95.37\\SPE: 65.87} & -\\
\hline
\makecell[c]{\cite{shi2017multimodal}} & 2017 & \makecell[c]{Single-level \\ Fusion} & \makecell[l]{202 subjects: \\52 NC, \\56 ncMCI, \\43 cMCI, \\51 AD} & \makecell[c]{ACC: 97.13\\SEN: 95.93\\SPE: 98.53} & \makecell[c]{ACC: 87.24\\SEN: 97.91\\SPE:  67.04} & \makecell[c]{4 Classes\\ACC: 57.00 \\SEN: 53.65\\SPE: 85.05}\\
\hline
\makecell[c]{\cite{zhang2019multi}} & 2019 & \makecell[c]{Single-level \\ Fusion} & \makecell[l]{392 subjects: \\101 NC, \\200 MCI, \\91 AD} & \makecell[c]{ACC: 98.47\\SEN: 96.58\\SPE: 95.39} & \makecell[c]{ACC: 85.74\\SEN: 90.11\\SPE:  91.82} & -\\
\hline
\makecell[c]{\cite{abdelaziz2022fusing}} & 2022 & \makecell[c]{Single-level \\ Fusion} & \makecell[l]{959 subjects: \\264 NC, \\273 ncMCI, \\ 204 cMCI,\\ 218 AD} & \makecell[c]{ACC: 98.24\\SEN: 98.82\\SPE: 97.52} & \makecell[c]{ACC: 94.59\\SEN: 90.26\\SPE:  96.98} & -\\
\hline
\makecell[c]{\cite{zhang2020multi}} & 2020 & \makecell[c]{Hierarchical \\ Fusion} & \makecell[l]{500 subjects: \\163 NC, \\113 ncMCI, \\ 105cMCI,\\ 119 AD} & \makecell[c]{ACC: 95.21\\SEN: 93.56\\SPE: 97.48} & - & \makecell[c]{4 Classes\\ACC: 86.15 }\\
\hline
\makecell[c]{\cite{fang2020ensemble}} & 2020 & \makecell[c]{Output \\ Fusion} & \makecell[l]{398 subjects: \\101 NC, \\204 MCI, \\ 93 AD} & \makecell[c]{ACC: 99.27\\SEN: 95.89\\SPE: 98.72} & \makecell[c]{ACC: 90.35\\SEN: 88.36\\SPE:  92.56} & -\\
\hline
\end{tabular}
\end{table*}

In general, we believe that deep multi-level fusion can better exploit the synergy of multimodal data to produce better classification results. This is further supported by the results in Tab.~\ref{tab_results}. Compared with the input fusion, the single-level fusion has a more robust feature fusion, which improves the overall ACC of the middle fusion. Hierarchical fusion utilizing multi-level feature fusion did not significantly improve the performance of dichotomous classification but performed well for four-class classifications. Generally, a complex model does not improve performance much when applied to a simple classification task. The more complex the network, the better it is at solving complex classification problems. When the number of categories for multi-category classification increases from two to four, the hierarchical fusion classification accuracy improves significantly. Last but not least, we note that the output fusion achieves excellent results on NC versus AD classification, thanks to the pre-training of different modal branches. With output fusion, DL backbones can be pre-trained on a large number of unimodal datasets and then fine-tuned on the multimodal datasets. Similar results are reported on other datasets. ABIDE data was combined with sMRI and fMRI to diagnose autism spectrum disorders. It was found that the hierarchical fusion \citep{liu2022attention} result was 87.2\%, which was better than the input fusion \citep{akhavan2018combination} result of 65.5\%. \citet{rahaman2022two} used the COBRE dataset for the diagnosis of schizophrenia, and the accuracy of input fusion, output fusion, and single-level fusion was 70\%, 78\%, and 95\%, respectively. Based on the GAMMA dataset, \citet{li2022multimodal} achieved 63\% accuracy in input fusion, 72\% accuracy in single-level fusion, and 80\% accuracy in hierarchical fusion when they used the same dataset for glaucoma diagnosis.

It is difficult to determine a unified solution for a wide variety of multimodal fusion medical image fusion tasks. In spite of this, we can draw some preliminary conclusions from the above analysis. For medical modalities with similar structures, modal registration is easier, so input fusion, single-level fusion, and hierarchical fusion are all network structures worth investigating. Generally, single-level fusion and hierarchical fusion fuse deeper features, which will improve the classification performance. When data have a wide range of structures or dimensions, single-level fusion and attention-based fusion are preferable solutions, as they are capable of handling a wide range of modal feature fusion scenarios. Lastly, if we have a large number of unimodal datasets for each modality in multimodal data, output fusion will perform well.

In addition to using a single multimodal fusion method, multiple fusion methods can be combined \citep{tang2022fusionm4net,li2023diagnostics,hu2020deep,wang2022combining}. \citet{tang2022fusionm4net} achieved the classification of skin lesions using a combination of single-level fusion and output fusion. For the classification of Diabetic Retinopathy, \citet{li2023diagnostics} utilized different configurations of OCT Angiography data. Their approach combined hierarchical fusion for registered modalities with late fusion for unregistered modalities. In order to improve the diagnosis of breast cancer, \citet{hu2020deep} fused multi-parametric MRI data at three levels: input, feature (intermediate), and decision (output). Combining different fusion methods can cumulate their advantages, allowing data from various perspectives to be fused and improving classification performance to some extent. It is one of the promising strategies that can be used when performing multimodal medical classification.

\subsection{How to find the best architecture?}

During the investigation of multimodal approaches, we have found that researchers need not only many tests to compare various fusion methods but also a large number of hyperparametric tests to determine the best network architecture for each fusion method. It takes a great deal of time and labor to conduct these extensive tests. There have been many recent studies that have applied Neural Architecture Search (NAS) to multimodal networks that can integrate various fusion techniques to determine the best architecture for a given dataset. The use of these methods is widespread in the fields of diagnosing dementia \citep{chatzianastasis2023neural}, gliomas segmentation \citep{wang2020neural}, multimodal action recognition \citep{perez2019mfas}, visual question answering\cite{yu2020deep}, multimodal damage identification \citep{singh2022neural}, multimodal gesture recognition \citep{yin2022bm}, etc. We believe that this approach has the potential to be explored in the future for multimodal classification in medicine.

\subsection{How to manage incomplete multimodal data?}
\label{sec:manage-incomplete-data}

The problem of modality incompleteness is one of the most pressing challenges in multimodality medical research. The high cost and potentially harmful effects of medical images may lead many patients to refuse being scanned with multiple imaging modalities for clinical diagnosis \citep{pan2020spatially}. In the ADNI dataset, all subjects had MRI data; however, only about half of the subjects had PET scans \citep{pan2020spatially}. The most common approach to solving the modality incomplete problem is to discard the modality incomplete subjects \citep{liu2014multimodal,calhoun2016multimodal,suk2014hierarchical,shi2017multimodal}, but this approach reduces the number of trainable subjects for the deep learning model, resulting in reduced classification performance. There is also the option of estimating the features of the missing subjects \citep{donders2006gentle,sterne2009multiple} based on, for example, the mean or median of the subjects with complete modal data, although this requires some prior knowledge and may introduce errors.

Generative Adversarial Networks (GAN) \citep{goodfellow2020generative} is a type of generative model used to produce data of a modality from another modality \citep{https://doi.org/10.48550/arxiv.1406.2661}. With the development of GAN, more and more fields are using this technology to generate images. The modal incompleteness problem has recently been solved through the use of GAN in many studies \changed{\citep{lin2021bidirectional,zhang2022bpgan,pan2018synthesizing,pan2020spatially,pan2021disease,jin2022hybrid,gao2023multimodal,tu2024multimodal}}. The GAN is used to generate the missing data and then the generated data is used for multimodal classification. It provides a significant increase in the number of subjects in the dataset, improves the model's classification performance, and is an effective solution when dealing with multimodal incompleteness. GAN-based solutions are currently the most promising.

\added{
Recent research endeavors have sought to confront the difficulties associated with addressing unbalanced datasets and incomplete modalities through the implementation of a distinctive fusion network design and training strategy. This approach aims to optimize the utilization of available dataset information while mitigating bias introduced by independently generating missing modalities. \citet{gravina2024multi} introduced a Multi-Input–Multi-Output 3D CNN designed for the assessment of dementia severity, specifically tailored for scenarios involving incomplete multimodal brain MRI and PET data. In alignment with our hierarchical fusion architecture, they incorporated a fusion branch named PAIRED-NET during feature extraction, employing distinct CNN branches for MRI and PET modalities, each capable of producing independent outputs. Simultaneous parameter updates for all three branches occurred during training when the patient sample encompassed the full modality set. In instances where a modality was missing, only the parameters of the branch corresponding to the available modality were updated. This methodology facilitates comprehensive network training using the entire dataset, enabling classification of instances with a single missing modality during testing. \citet{liu2023cascaded} introduced a cascaded Multi-Modal Mixing Transformer (3MT)  designed for the classification of Alzheimer's Disease with incomplete data. The architecture of 3MT comprises a sequence of Cascaded Modality Transformers (CMTs), each incorporating features from a specific modality. At the conclusion of this sequence, a more informed class prediction is obtained by aggregating the extracted multi-modal features. In scenarios involving missing data, the CMTs corresponding to the absent modalities receive zero embeddings, indicating "not available" to the model. This training approach equips the model with prior knowledge for handling diverse missing data scenarios.}

\subsection{Does multimodal fusion always performs better?}

Multimodal data not only contain complementary information but may also contain a great deal of redundant information. One study found that multimodal fusion did not enhance classification performance \citep{khagi20203d}. In one sense, this relates to the design of the network, and in another sense, multimodal fusion may not improve the performance of classification if the information in the modalities is relatively similar or if a particular modality does not accurately define the target class. Before starting a multimodal fusion project and gathering multimodal data, it is advisable to conduct a redundancy analysis \citep{salvador2019multimodal}.

Alternately, \citet{narazani2022pet} questioned the multimodal diagnostic objectives. Clinical studies primarily aim to determine the type of dementia, whereas studies on DL focused on only one type, AD. multimodal fusion studies have performed well in terms of classifying NC versus AD, but the clinical goal is the classification of AD at multiple levels. As a result of their tests, multimodal fusion networks did not improve the multi-level classification. Therefore, multimodal fusion classification studies should be conducted in conjunction with clinical needs.

\subsection{Can we advantageously combine multimodal image classification with other tasks?}

We should point out that input fusion \citep{pereira2016brain,isensee2018brain,isensee2019no,cui2018automatic,kamnitsas2017efficient}, intermediate fusion \citep{dolz2018hyperdense,dolz2019ivd,chen2018mri,andrade2022pure,li2022cross}, and output fusion \citep{kamnitsas2018ensembles,aygun2018multi} methods can also be applied to medical image segmentation, medical image fusion, and similar tasks. A notable trend is to incorporate the fusion network into the feature extraction process, thereby enabling the creation of multi-task multimodal networks. \citet{cheng2022fully} proposed an end-to-end multi-task learning network for simultaneous glioma segmentation and IDH genotyping based on the sharing of spatial and global feature representations extracted from the hybrid CNN-Transformer encoder. The performance of both classification and segmentation can be enhanced through the use of a joint network.

\subsection{\added{Can we advantageously combine images with contextual data in a classification pipeline?}}

\added{In this survey, our main focus was on integrating data from various imaging modalities. However, out of 114 reviewed papers, 27 incorporate non-image contextual information in various forms, often structured demographic and clinical metadata extracted from electronic health records \citep{huang2020multimodal, zhou2021cohesive, yan2021richer, venugopalan2021multimodal, prabhu2022multi}. Clinical metadata includes the result of physical tests, like visual acuity or refraction for chronic central serous chorioretinopathy diagnosis \citep{yoo2022deeppdt}, the result of chemical tests, like Pap or HPV for cervical dysplasia diagnosis \citep{xu2016multimodal}, and the result of cognitive tests, like executive functioning (ADNI-EF) and memory (ADNI-MEM) test for Alzheimer's disease diagnosis \citep{lee2019predicting}. Sometimes, images are combined with more complex contextual data like voice signals \citep{kwon2022diagnosis}, free-form text \citep{zhang2022multimodal}, or genomic data \citep{rahaman2021multi, venugopalan2021multimodal, rahaman2022two, qiu2023hierarchical}. Image and contextual metadata have no geometrical relationship, so input and hierarchical fusion are not relevant in this scenario. Earlier solutions thus relied on single-level intermediate fusion or output fusion. Recent studies tend to use attention-based intermediate fusion \citep{zhang2022multimodal, qiu2023hierarchical, chen2023multimodal}. One challenge with contextual data is that they are often incomplete: the reader is referred to section \ref{sec:manage-incomplete-data} for this challenge. Nevertheless, all these studies report increased classification performance when using contextual data.}

\subsection{Trends for the future}

As the network structure evolves and hardware devices become increasingly available, there has been a growing interest in multimodality research. From Table~\ref{tab_results}, it is evident that the rapid development of multimodal fusion has led to significant improvements in classification results. In fact, multimodal networks based on deep learning exhibit greater potential for development than unimodal networks.

\begin{figure}[!t]
\includegraphics[width=0.5\textwidth]{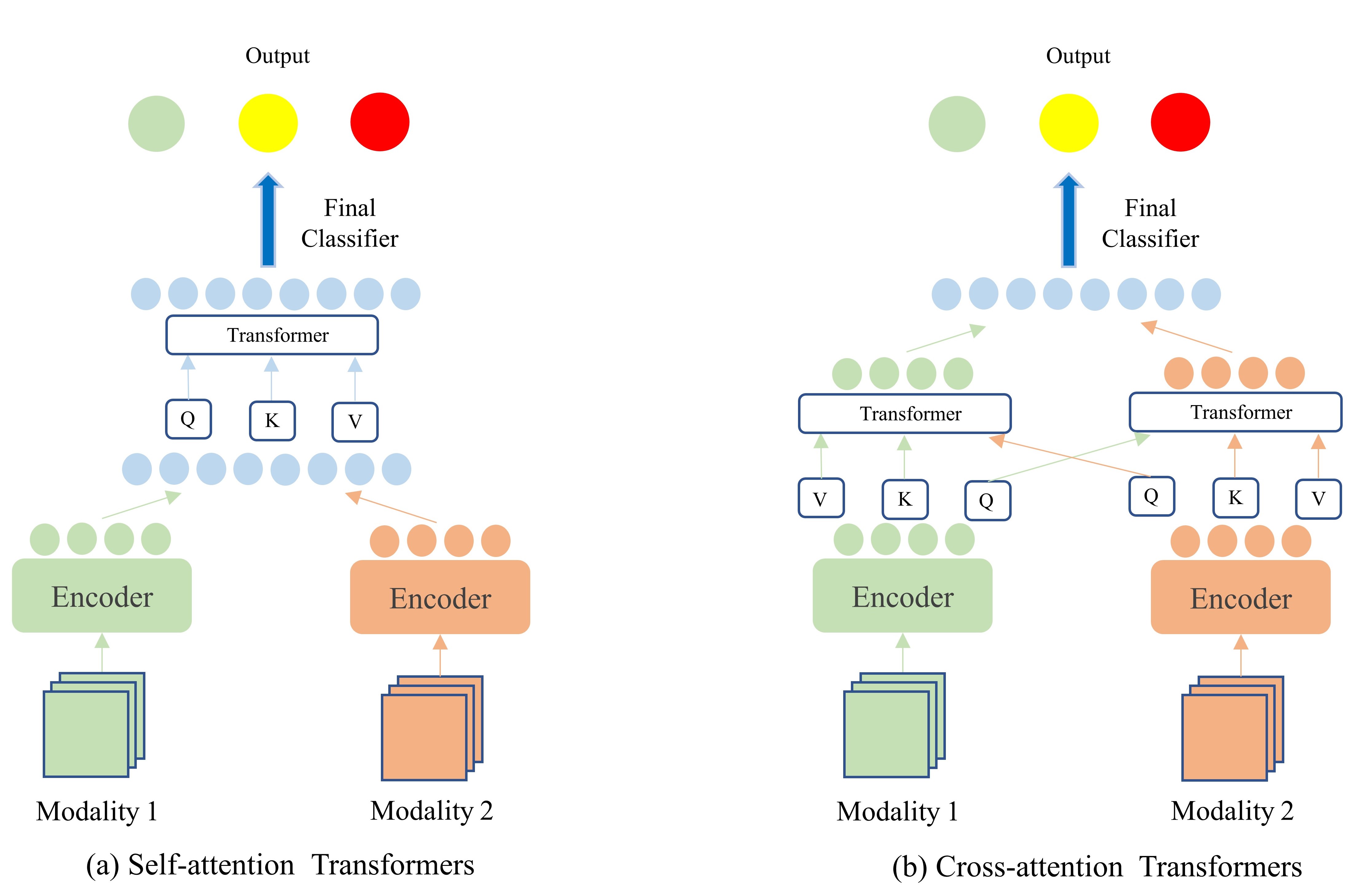}
\caption{Two architectures of Transformer-based fusion.} \label{fig_transformer}
\end{figure}

Transformer is one of the most popular network architectures, and multimodal fusion based on Transformer has developed rapidly in the past two years. In particular, for visual-language tasks, Transformer can handle the fusion of images, languages, and text very effectively. Based on the research conducted in different fields, we classify Transformer-based multimodal fusion networks into self-attention Transformers \citep{akbari2021vatt,nagrani2021attention,shi2022learning,li2021ai,pashevich2021episodic,appalaraju2021docformer,steitz2022txt,wu2022marmot} and cross-attention Transformers \citep{lu2019vilbert,chen2021multimodal,tan2019lxmert,zhu2020actbert,ramesh2021vset,rahman2021tribert,chen2021multimodal,chen2021history,li2022deepfusion}, as shown in Fig.~\ref{fig_transformer}. Following the extraction of features using encoders, self-attention Transformers concatenate features from different modalities and compute the attention relationship between the fused features using Transformer blocks. Alternatively, cross-attentional Transformers compute the attentional relationships among different modalities in order to achieve information fusion. Nowadays, these two architectures are the most popular multimodal fusion networks.

Compared to CNNs, Transformers have the advantage of efficiently identifying long-range relationships between sequences.  In medical images, most visual representations are ordered due to the similarity of human organs. Medical images contain more information regarding sequence relationships than natural images \citep{dai2021transmed}. This indicates that Transformer-based multimodal medical image fusion is a promising approach, and the above two network architectures are worth exploring. \added{While recent medical research has employed analogous structures \citep{gao2023multimodal,zuo2023alzheimer,bi2023multimodal}, it is imperative to undertake broader investigations and validations to extend the applicability of these findings.}

In addition to these developments, the field is also witnessing an emerging trend in the exploration of representation learning for multimodal data using techniques such as pretext tasks or contrastive learning. These methods aim to learn robust and transferable representations that can be applied to downstream tasks of a classification nature or directly in parallel with the classification task \citep{wei2022multimodal,MohitPrabhushankar2022,Cai2022,Gutiérrez2022,Xing2022,Taleb_2022_CVPR,hager2023best}. This field is new and a lot of research questions are emerging. How to learn an aligned representation across modalities? Can we learn one representation space for the different modalities? The research community is actively moving in this direction to address challenges associated with representation learning.

\added{In response to the challenge posed by the limited scale of medical datasets and the laborious nature of manual labeling, certain investigations have resorted to harnessing information from image-text pairs available on the web. This approach facilitates the construction of transfer learning multimodal models, which have demonstrated notable efficacy in subsequent fine-tuning tasks and zero-shot classification endeavors. \citet{zhang2022contrastive} introduced ConVIRT, an unsupervised methodology for acquiring medical visual representations through the analysis of naturally paired images and text. Their approach involves evaluating image representations in conjunction with paired text representations through a bidirectional objective, surpassing alternative methods in performance across various downstream medical classification tasks. Drawing inspiration from the Contrastive Language–Image Pre-training (CLIP) approach \citep{radford2021learning}, \citet{huang2023visual} undertook fine-tuning specifically tailored for medical applications, resulting in the formulation of Pathology Language–Image Pretraining (PLIP). Exploiting numerous de-identified images and abundant textual data disseminated by clinicians on public platforms like Twitter, PLIP introduces a multimodal, unsupervised, Transformer-based transfer learning model. This model effectively classifies new pathological images across four external datasets, exhibiting state-of-the-art performance. These initiatives highlight the immense potential of publicly shared medical information as a valuable resource for developing multimodal medical AI systems, thereby enhancing the landscape of medical diagnosis.}

In support to these advancements and needs, the TorchMultimodal library\footnote[15]{\url{https://github.com/facebookresearch/multimodal}} has been created as a framework for training state-of-the-art multimodal multi-task models at scale by Meta Research using PyTorch framework. This library is the result of concerted community efforts, reflecting the growing focus on multimodal tasks and methodologies, as well as representation learning for multimodal data. \added{Furthermore, in alignment with the increasing enthusiasm for multimodal learning, libraries such as Transformers \footnote[16]{\url{https://huggingface.co/docs/transformers/index}} by Hugging Face offer robust resources for constructing and training such models. These Transformers present pretrained models that can handle various modalities, such as text and images, simplifying tasks like image categorization or answering questions with multimodal data. This emphasis on multimodal functionalities underscores the growing demand for models that possess the ability to comprehend and analyze diverse types of data. Although Transformers are not exclusively designed for the medical domain, they can serve as a beneficial resource for scholars aiming to create and distribute code for tasks related to multimodal classification of medical images.}

As illustrated in Tab.~\ref{tab_paper}, very few multimodal image classification papers in the medical field are associated with public code. We hope TorchMultimodal, \added{Transformers} or similar libraries will facilitate code sharing.

\section{Conclusion}
\label{sec: Conclusion}

In this paper, we conducted a comprehensive review of the development of deep learning-based multimodal medical classification tasks over the past few years. We examined the complementary relationships among several common clinical modalities and delved into five \changed{types of} architectures for deep learning multimodal classification networks: input fusion, single-level fusion, hierarchical fusion, attention-based fusion, and output fusion. Our study covered a wide range of multimodal fusion scenarios in medical classification and the application domains for which different network architectures are most suitable.

Additionally, we discussed emerging trends and challenges in the field, including the exploration of representation learning techniques and the development of dedicated frameworks like TorchMultimodal. These advancements provide efficient tools for training state-of-the-art multimodal multi-task models at scale. In particular, we highlighted the advantages of using Transformer-based multimodal fusion architectures, particularly in medical imaging applications, where sequence relationships are more prevalent. This demonstrates the potential of these architectures in advancing the field of multimodal medical classification tasks.

Looking forward, we encourage the research community to continue investigating novel fusion techniques, optimization methods, and network architectures to further enhance the performance of multimodal classification tasks. Developing interpretable models, addressing data imbalance and scarcity, and exploring unsupervised and semi-supervised learning approaches are other areas worth investigating. Additionally, we recommend future research focus on the application of multimodal fusion in emerging areas such as genomics, proteomics, and patient-centered care, where the integration of diverse data types can potentially lead to significant improvements in diagnostic and therapeutic outcomes.

\section*{Declaration of Competing Interest}
The authors declare that they have no known competing financial interests or personal relationships that could have appeared to influence the work reported in this paper.

\section*{Acknowledgments}
The work takes place in the framework of the ANR RHU project Evired. This work benefits from State aid managed by the French National Research Agency under the ``Investissement d'Avenir'' program bearing the reference ANR-18-RHUS-0008. It was also funded in part by the ANR under the LabCom program (ANR-19-LCV2-0005 - ADMIRE project).

\clearpage
\onecolumn
\centering
\begin{longtable}{|l|l|}
\caption{List of Terms.}\label{tab_terms}\\
\hline
Term &  Description \\
\hline
ADC & Apparent Diffusion Coefficient\\
\hline
AD & Alzheimer's Disease\\
\hline
AdaBoost & Adaptive Boosting\\
\hline
ADHD & Attention Deficit Hyperactivity Disorder\\
\hline
AE & Auto-Encoder\\
\hline
AGDAF & Attention Guided Discriminative and Adaptive Fusion\\
\hline
ASD & Autism Spectrum Disorder\\
\hline
\added{CatBoost} & \added{Categorical Boosting}\\
\hline
CFP & Color Fundus Photographs\\
\hline
CLIMAT & Clinically-Inspired Multi-Agent Transformers\\
\hline
CMR & Cardiac Magnetic Resonance\\
\hline
CNN & Convolutional Neural Network\\
\hline
CSF & CerebroSpinal Fluid\\
\hline
CT & Computed Tomography\\
\hline
DBM & Deep Boltzmann Machine\\
\hline
DBN & Deep Belief Network\\
\hline
DCE & Dynamic Contrast-Enhanced\\
\hline
DL & Deep Learning\\
\hline
DNN & Deep Neural Network\\
\hline
Dsc & Dermatoscopic Image\\
\hline
DWI & Diffusion-Weighted Imaging\\
\hline
DXA & Dual-energy X-ray Absorptiometry\\
\hline
EHR & Electronic Health Records\\
\hline
EMR & Electronic Medical Record\\
\hline
FA & Fractional Anisotropy\\
\hline
FC Layer & Fully Connected Layer\\
\hline
FCNN & Fully Connected Neural Network\\
\hline
FHR & Fetal Heart Rate\\ 
\hline
FLAIR & Fluid Attenuated Inversion Recovery\\
\hline
FFN & Feed-Forward Network\\
\hline
fMRI & functional Magnetic Resonance Imaging\\
\hline
GAN & Generative Adversarial Network\\
\hline
GRU & Gated Recurrent Unit\\
\hline
HoFN & High-order Factorization Network\\
\hline
HPV & Human Papillomavirus\\
 \hline
KELM & Kernel‐based Extreme Learning Machine\\
 \hline
 KNN & K-Nearest Neighbor \\
 \hline
 LDA&Linear Discriminant Analysis\\
 \hline
LR & Logistic Regression\\
 \hline
LSTM & Long-Short Term Memory\\
 \hline
MD & Mean Diffusivity\\
 \hline
mDSNet & multimodality Disease-Image-Specific Network\\
 \hline
MGF & Multiheaded Gating Fusion\\
 \hline
MLP & Multilayer Perceptron\\
 \hline
MMNet & Multimodal 3D medical image classification Network\\
\hline
MM-SDPN & Multimodal Stacked Deep Polynomial Networks\\
 \hline
MRI & Magnetic Resonance Imaging\\
 \hline
NBI & Narrow Band Imaging\\
 \hline
PCA & Principal Component Analysis\\
 \hline
PEI & Positive Enhancement Integral\\
\hline
PET & Positron Emission Tomography\\
\hline
RBM & Restricted Boltzmann Machine\\
 \hline
ROI & Regions-Of-Interest\\
\hline
SAE & Stacked Auto-Encoder \\
\hline
\added{SBM} & \added{Swapping Bi-Attention Mechanism}\\
 \hline
sELM‐AE & stacked sparse Extreme Learning Machine Auto‐Encoder\\
 \hline
sMRI & structural Magnetic Resonance Imaging\\
\hline
SNP & Single Nucleotide Polymorphism\\
 \hline
SVM & Support Vector Machine\\
 \hline
 T1 & T1-Weighted \\
 \hline
T1C & T1-Contrast\\
 \hline
T1-Gd & T1-weighted Gadolinium Contrasted\\
 \hline
T2 & T2-Weighted\\
\hline
T2WI & T2-Weighted MP-MRI Image\\
\hline
TSI & Tumor Shape Image\\
\hline
 TWIST& Time-resolved Angiography with Interleaved Stochastic Trajectories\\
  \hline
OCT & Optical Coherence Tomography\\
  \hline
UC & Uterine Contraction\\ 
\hline
US & Ultrasound\\
  \hline
VF & Visual Field\\
  \hline
ViT & Vision Transformer\\
   \hline
 WSI& Whole Slide Image\\
\hline
XGBoost & Extreme Gradient Boosting\\
\hline
\end{longtable}

\footnotesize
\begin{longtable}{|p{1.7cm}|p{1.4cm}|p{0.65cm}|p{1.7cm}|p{1.0cm}|p{1.0cm}|p{0.8cm}|p{1.1cm}|p{0.72cm}|p{0.5cm}|p{0.45cm}|}
\caption{List of publications for different fusion networks. IF: Input Fusion; SLF: Single-level Fusion; HF: Hierarchical Fusion; ABF: Attention-based Fusion; OF: Output Fusion.  }
\label{tab_paper}\\
\hline
Research work  & Multimodal combination & Fusion Methods& Information Fusion Technique & DL Backbone & Final Classifier & Body Organ & Dataset& Patients & Class & Code \\
\hline
\cite{li2015robust}  &MRI, PET, CSF&IF & Concatenation (Inputs) & RBM&SVM &Brain & ADNI& 202&2&N/A\\
\hline
\cite{liu2014multimodal}  & MRI, PET&IF & Concatenation (Inputs)& Manual & Softmax&Brain & ADNI&331& 2,4 & N/A\\
\hline
\cite{akhavan2018combination}  &sMRI, fMRI &IF & Concatenation (Inputs) & DBN& FC Layer&Brain & ABIDE I, ABIDE II & 185 & 2 &  N/A\\
\hline
\cite{aldoj2020semi}  & MRI (ADC, DWI, T2)&IF & Concatenation (Inputs)&CNN & FC Layer&Prostate & TCIA (PROSTATEx)& 200 & 2 &N/A\\
\hline
\cite{qian2020combined}  & US (US B-mode, US color
Doppler) &IF & Concatenation (Inputs)&CNN &FC Layer &Breast & Private Data &59959 & 4 & N/A\\
\hline
\cite{zong2020deep}  &MRI (ADC, DWI, T2)&IF & Concatenation (Inputs)& CNN& FC Layer&Prostate & TCIA (PROSTATEx) &201 &2 & N/A\\
\hline
\cite{sanford2020deep}  &MRI (T2, ADC, High-b)&IF & Merge (Inputs)& Resnet& FC Layer& Prostate&Private Data, TCIA (PROSTATEx) & 687 & 2 & N/A\\
\hline
\cite{lin2021bidirectional}  &MRI, PET&IF & Concatenation (Inputs)&CNN &FC Layer &Brain & ADNI& 1086&2 &N/A\\
\hline
\cite{song2021effective}  &MRI, PET&IF & Merge (Inputs)& CNN& FC Layer&Brain & ADNI & 381 & 2,3 &  N/A\\
\hline
\cite{decuyper2021automated}  &MRI (T1, T1C, T2M FLAIR)&IF & Merge (Inputs)& CNN& FC Layer& Brain&TCIA (GBM, LGG), BraTS 2019 & 628 & 2 & N/A\\
\hline
\cite{kong2022multi}  &MRI, PET&IF & Merge (Inputs) &CNN & FC Layer& Brain& ADNI & 370 & 2,3 & N/A\\
\hline
\cite{zhang2022bpgan}  &MRI, PET&IF & Concatenation (Inputs) & Resnet&FC Layer & Brain& ADNI & 873 & 2,3 & N/A\\
\hline
\cite{zhou2023prediction}  &MRI (PEI, DWI)&IF & Concatenation (Inputs)& CNN& FC Layer&Breast & Private Data & 210 & 2 & N/A\\
\hline
\added{\cite{rallabandi2023deep}}  &\added{MRI, PET}&\added{IF} & \added{Merge (Inputs)} & \added{CNN} & \added{SoftMax}&\added{Brain} & \added{OASIS-3} & \added{1098} & \added{2} & \added{N/A}\\
\hline
\added{\cite{odusami2023explainable}}  &\added{MRI, PET}&\added{IF} & \added{Concatenation (Inputs)} & \added{ResNet}& \added{FC Layer}&\added{Brain} & \added{ADNI} & \added{412} & \added{2} & \added{N/A}\\
\hline
\cite{suk2013deep}  &MRI, PET, CSF&SLF & Concatenation (Classic)& SAE&SVM &Brain &ADNI & 202 & 2 & N/A\\
\hline
\cite{suk2014hierarchical}  &MRI, PET&SLF & Concatenation (Classic) &DBM & SVM& Brain& ADNI & 398 & 2 & N/A\\
\hline
\cite{suk2015latent}  &MRI, PET&SLF & Concatenation (Classic)& SAE& SVM&Brain & ADNI & 202 & 2 & N/A \\
\hline
\cite{xu2016multimodal}  &Photograph of the cervix, Pap tests, HPV tests&SLF & Concatenation (Classic) &CNN &FC Layer &Cervix & TCIA (Guanacaste Project) & 690 & 2 & N/A \\
\hline
\cite{mehrtash2017classification}  &MRI (ADC, DWI, DCE)&SLF & Concatenation (Classic) & CNN& FC Layer&Prostate &TCIA (PROSTATEx) & 201 & 2 & N/A   \\
\hline
\cite{vu2017multimodal}  &MRI, PET&SLF & Concatenation (Classic) &CNN &FC Layer &Brain & ADNI & 317 & 2 & N/A\\
\hline
\cite{zou20173d}  &sMRI, fMRI&SLF & Concatenation (Classic) &CNN &FC Layer &Brain & ADHD-200 & 730 & 2 & N/A\\
\hline
\cite{yang2017co}  &MRI (ADC, T2)&SLF & Merge (Classic) & CNN& FC Layer& Prostate& Private Data & 160 & 2 &\footnote[16]{\url{https://github.com/Andysis/co-trained-CADx}}\\
\hline
\cite{shi2017multimodal}  &MRI, PET&SLF & Concatenation (Network) & MM-SDPN&FC Layer &Brain & ADNI & 202 & 2,4 & N/A\\
\hline
\cite{zhou2017feature}  &MRI, PET&SLF & Concatenation (Network) &DNN &Score Merge &Brain & ADNI & 805 & 3,4 & N/A\\
\hline
\cite{ye2017glioma}  &MRI (TI, T2, T1C, FLAIR)&SLF& Merge (Classic)&CNN &FC Layer &Brain & BraTS 2015 & 274  & 2 &N/A \\
\hline
\cite{cheng2017cnns}  &MRI, PET&SLF & Concatenation (Network) & CNN&FC Layer &Brain &ADNI & 193 & 2 & N/A \\
\hline
\cite{le2017automated}  &MRI (ADC, T2WI)&SLF & Merge (Classic) &CNN & SVM&Prostate & TCIA, Private Data & 364 & 2 & N/A\\
\hline
\cite{pan2018synthesizing}  &MRI, PET&SLF& Concatenation (Classic) &CNN & FC Layer&Brain &ADNI & 1457 & 2 & N/A \\
\hline
\cite{ge2018deep}  &MRI (TI, T2, FLAIR)&SLF & Merge (Classic)& CNN& FC Layer&Brain & BraTS 2017, TCIA (LGG) & 444 & 2 & N/A\\
\hline
\cite{liu2018multi}  &MRI, PET&SLF & Concatenation (Classic)& CNN& FC Layer& Brain& ADNI & 397 & 2 & N/A\\
\hline
\cite{wang2018automated}  &MRI (ADC, T2)&SLF & Concatenation (Classic)& CNN&Softmax &Prostate &Private Data, TCIA (PROSTATEx) & 360 & 2& N/A\\
\hline
\cite{kim2018identification}  &MRI, PET, CSF&SLF & Concatenation (Network)& sELM‐AE&KELM & Brain& ADNI & 202 & 2 & N/A\\
\hline
\cite{kawahara2018seven}  &Dsc, Clinical Image, Metadata&SLF& Concatenation (Classic)&CNN & FC Layer&Skin & SPC & 1011 & 2 & N/A\\
\hline
\cite{lu2018multimodal}  &MRI, PET&SLF & Concatenation (Network)& DNN& Score Merge& Brain& ADNI & 626 & 2 & N/A\\
\hline
\cite{zhou2019effective}  &MRI, PET, SNP&SLF & Concatenation (Network)&DNN & Score Merge& Brain& ADNI & 360 & 2 & N/A\\
\hline
\cite{feng2019deep}  &MRI, PET&SLF & Concatenation (Classic)& CNN, LSTM& Softmax& Brain& ADNI & 397 & 2 & N/A\\
\hline
\cite{dalmis2019artificial}  &MRI (ADC, T2, TWIST)&SLF & Concatenation (Classic)& CNN&Random Forest &Breast & Private Data & 576 & 2 & N/A\\
\hline
\cite{huang2019diagnosis}  &MRI, PET&SLF & Concatenation (Classic)&VGG &FC Layer & Brain&ADNI& 1512& 2 & N/A \\
\hline
\cite{lee2019predicting}  &MRI, CSF, Demographic Information, Cognitive Performance&SLF & Concatenation (Classic) &GRU &LR &Brain & ADNI & 1618 & 2 & N/A  \\
\hline
\cite{punjabi2019neuroimaging}  &MRI, PET&SLF & Merge (Classic)& CNN& FC Layer&Brain & ADNI & 723 
 & 2 & N/A\\
\hline
\cite{zhang2019multi}  &MRI, PET&SLF & Concatenation (Classic)&CNN &Softmax &Brain & ADNI & 392 & 2 & N/A\\
\hline
\cite{qin2020fine}  &PET, CT&SLF & Merge (Classic) & CNN& FC Layer& Lung&Private Data& 397 & 3&N/A  \\
\hline
\cite{pan2020spatially}  &MRI, PET&SLF & Concatenation (Classic)&CNN &FC Layer &Brain &ADNI & 2355 & 2 &  N/A\\
\hline
\cite{el2020multimodal}  &MRI, PET, Clinical Datas&SLF & Concatenation (Classic) &CNN, LSTM & FC Layer&Brain &ADNI & 1536 & 4 & N/A\\
\hline
\cite{venugopalan2021multimodal}  &MRI, EHR, SNP&SLF & Concatenation (Classic)&CNN, SAE &FC Layer &Brain & ADNI & 808 & 3 & N/A\\
\hline
\cite{qian2021prospective}  &US (US B-mode, US color
Doppler, US elastography images)&SLF & Concatenation (Classic) &CNN &FC Layer & Breast&Private Data & 775 &2& N/A \\
\hline
\cite{massalimova2021input}  &MRI (T1, FA, MD)&SLF & Concatenation (Classic)&ResNet &FC Layer &Brain & OASIS-3 & 1098  & 3 & N/A\\
\hline
\cite{abdelaziz2021alzheimer}  &MRI, PET, SNP&SLF & Concatenation (Classic) & CNN&FC Layer & Brain& ADNI & 805 & 2&N/A \\
\hline
\cite{zhou2021cohesive}  &CT, EMR&SLF& Concatenation (Classic) & CNN, HoFN&FC Layer &Lung &Private Data & 733 & 4 & N/A \\
\hline
\cite{pan2021disease}  &MRI, PET&SLF & Concatenation (Classic) &mDSNet &FC Layer & Brain&ADNI & 1455&  2 & N/A\\
\hline
\cite{yan2021richer}  &EMR, Pathological images&SLF & Concatenation (Classic) &VGG, AE & FC Layer&Breast & Private Data  & 185 & 2 & N/A\\
\hline
\cite{zhang2021multimodal}  &MRI, PET&SLF & Concatenation (Network) &CNN & Score Merge&Brain & ADNI & 875 & 2 & N/A\\
\hline
\cite{joo2021multimodal}  &MRI (T1, T2), Clinical Information&SLF & Concatenation (Classic) &ResNet &FC Layer &Breast & Private Data & 536 & 2 & N/A\\
\hline
\cite{rahaman2021multi}  &sMRI, fMRI, SNP&SLF & Concatenation (Classic)&DNN &FC Layer &Brain & COBRE & 437 & 2 & N/A\\
\hline
\cite{abdelaziz2022fusing}  &MRI, PET&SLF & Concatenation (Network)& CNN& FC Layer& Brain& ADNI& 959  & 2 & N/A\\
\hline
\cite{puyol2022multimodal}  & Echocardio-graphy, CMR&SLF & Merge (Network)& CNN& SVM& Heart& Private Data & 50 & 2 & N/A\\
\hline
\cite{al2022cardiovascular}  &DXA, CFP&SLF & Concatenation (Network)& CNN& FC Layer& Heart& Private Data & 483 & 2 & N/A\\
\hline
\cite{rahaman2022two}  &sMRI, fMRI, Genomic Sequence&SLF & Merge (Network)& AE, MLP, LSTM& Softmax& Brain& COBRE & 437 & 2 & N/A\\
\hline
\cite{wu2022gamma}  &CFP, OCT&SLF& Concatenation (Classic) &CNN & FC Layer&Eye &GAMMA & 300 & 3 & N/A\\
\hline
\cite{jin2022hybrid}  &MRI, PET&SLF & Concatenation (Network) &ResNet & FC Layer& Brain&ADNI & 360 & 2 & N/A\\
\hline
\cite{xiong2022multimodal}  &VF, OCT&SLF & Merge (Classic) & CNN& FC Layer&Eye & Private Data & 1083 & 2 & N/A\\
\hline
\cite{huang2022detecting}  &VF, CFP&SLF & Concatenation (Classic) &CNN & FC Layer&Eye & Private Data &  1027 & 2 & N/A\\
\hline
\cite{xing2022advit}  &PET-AV45, PET-FDG&SLF & Concatenation (Classic)& ViT& FC Layer&Brain &ADNI & 381 & 2 & N/A\\
\hline
\cite{dolci2022deep}  &sMRI, fMRI, SNP&SLF & Concatenation (Classic)& CNN& FC Layer&Brain & ADNI& 788 & 2 & N/A\\
\hline
\added{\cite{tu2022alzheimer}}  &\added{MRI, Metadata}&\added{SLF} & \added{Concatenation (Classic)}& \added{CNN}& \added{FC Layer}&\added{Brain} & \added{ADNI}& \added{1461} & \added{2} & \added{N/A}\\
\hline
\added{\cite{leng2023multimodal}}  &\added{MRI, PET}&\added{SLF} & \added{Merge (Network)}& \added{CNN}& \added{FC Layer}&\added{Brain} & \added{ADNI}& \added{536} & \added{2} & \added{N/A}\\
\hline
\added{\cite{liu2023improving}}  &\added{MRI, Metadata}&\added{SLF} & \added{Concatenation (Classic)}& \added{VGG}& \added{CatBoost}&\added{Brain} & \added{ADNI}& \added{242} & \added{2} & \added{N/A}\\
\hline
\added{\cite{kollias2023btdnet}}  &\added{MRI (T1, T2, T1C, FLAIR)} &\added{SLF} & \added{Concatenation (Classic)}& \added{CNN, RNN}& \added{FC Layer}&\added{Brain} & \added{BraTS 2021}& \added{585} & \added{2} & \added{N/A}\\
\hline
\added{\cite{kadri2023efficient}}  &\added{PET, MRI, CT}&\added{SLF} & \added{Concatenation (Classic)}& \added{ViT, CNN}& \added{FC Layer}&\added{Brain} & \added{OASIS-1, OASIS-3, ADNI}& \added{N/A} & \added{2,3,4} & \added{N/A}\\
\hline
\added{\cite{abbas2023deepmnf}}  &\added{sMRI, fMRI}&\added{SLF} & \added{Concatenation (Classic)}& \added{CNN}& \added{FC Layer}&\added{Brain} & \added{ABIDE I} & \added{855} & \added{2} & \added{N/A}\\
\hline
\added{\cite{lu2024hierarchical}}  &\added{MRI, SNP, Clinical data}&\added{SLF} & \added{Merge (Network)}& \added{CNN}& \added{FC Layer}&\added{Brain} & \added{ADNI}& \added{577} & \added{2} & \added{N/A}\\
\hline
\added{\cite{saponaro2024deep}}  &\added{sMRI, fMRI}&\added{SLF} & \added{Concatenation (Classic)}& \added{DNN}& \added{FC Layer}&\added{Brain} & \added{ABIDE I, ABIDE II} & \added{1383} & \added{2} & \added{N/A}\\
\hline
\added{\cite{gravina2024multi}}  &\added{MRI, PET}&\added{SLF} & \added{Concatenation (Classic)}& \added{CNN}& \added{FC Layer}&\added{Brain} & \added{OASIS-3}& \added{1025} & \added{2,3} & \added{N/A}\\
\hline
\cite{mahmood2018multimodal} &White Light RGB, NBI&HF & Merge (Network), Concatenation (Classic)& CNN& FC Layer& Digestive tract & ISIT-UMR & 76 & 3 & N/A\\
\hline
\cite{zhang2020multi} &MRI, PET&HF& Merge (Network), Merge (Classic)& CNN& Softmax& Brain& ADNI & 500 & 2,4 & N/A\\
\hline
\cite{zhou2021use} &MRI, PET&HF & Merge (Network),Merge (Classic)&RBM &Softmax &Brain & ADNI & 340 & 2 &N/A \\
\hline
\cite{he2021hierarchical} &MRI (T1C, FLAIR)&HF & Merge (Network), Merge (Classic) & CNN& FC Layer& Brain& TCIA, BraTS 2017 & 499 & 2 & N/A\\
\hline
\cite{gao2021task} &MRI, PET&HF  & Concatenation (Network), Concatenation (Classic)&CNN &FC Layer &Brain & ADNI & 977 & 2 & N/A\\
\hline
\cite{li2022multimodal} &CFP, OCT&HF  & Merge (Network), Concatenation (Classic)& CNN& FC Layer& Eye& GAMMA, Private Data & 264 & 2,3 & N/A\\
\hline
\cite{liu2022attention}  &sMRI, fMRI&HF & Merge (Network), Concatenation (Classic)&MGF &FC Layer &Brain & ABIDE, ADHD-200, COBRE & 2120 & 2 & N/A \\
\hline
\added{\cite{xu2023multi}}  &\added{MRI, PET}&\added{HF} & \added{Merge (Network),Concatenation (Classic)}&\added{CNN} &\added{FC Layer} &\added{Brain} & \added{ADNI} & \added{579} & \added{2} & \added{N/A} \\
\hline
\added{\cite{wu2023aggn}}  &\added{MRI (T1, T2, T1C, FLAIR)} &\added{HF} & \added{Merge (Network), Concatenation (Classic)}&\added{CNN} &\added{FC Layer}&\added{Brain} & \added{BraTS 2018, BraTS 2019} & \added{326} & \added{2} & \added{N/A} \\
\hline
\added{\cite{omeroglu2023novel}}  &\added{Dsc, Clinical Image, Metadata} &\added{HF} & \added{Merge (Network), Concatenation (Classic)}&\added{CNN} &\added{FC Layer}&\added{Skin} & \added{SPC} & \added{1011} & \added{2,3,5} & \added{N/A} \\
\hline
\added{\cite{miao2024mmtfn}}  &\added{MRI, PET}&\added{HF} & \added{Merge (Network), Merge (Classic)}&\added{CNN, ViT} &\added{FC Layer} &\added{Brain} & \added{ADNI} & \added{720} & \added{2} & \added{N/A} \\
\hline
\added{\cite{xu2024cross}}  &\added{MRI (T1, T2, T1C, FLAIR)} &\added{HF} & \added{Concatenation (Network), Concatenation (Classic)}&\added{CNN} &\added{SoftMax} &\added{Brain} & \added{BraTS 2018, BraTS 2019} & \added{620} & \added{2} & \added{N/A} \\
\hline
\added{\cite{tu2024multimodal}}  &\added{MRI, PET}&\added{HF} & \added{Merge (Network), Concatenation (Classic)}&\added{CNN} &\added{FC Layer} &\added{Brain} & \added{ADNI} & \added{821} & \added{2} & \added{N/A} \\
\hline
\cite{dai2021transmed} &MRI (T1,T2)& ABF&Attention Fusion&TransMed&FC Layer&Parotid, Knee & MRNet & 344 & 2 & N/A\\
\hline
\cite{hoang2022clinically} & Visual Data, Clinical Data&ABF&Attention Fusion&CLIMAT&FFN&Brain, Knee & ADNI, OAI \citep{nevitt2006osteoarthritis} & 10260& 5 & \footnote[17]{\url{https://github.com/oulu-imeds/climatv2}}\\
\hline
\cite{liu2023cascaded} & Neuro-imaging Data, Clinical Data&ABF&Attention Fusion&3MT&FC Layer&Brain& ADNI & 816  & 2 &  N/A\\
\hline
\cite{zhang2022multimodal} & Image, Text&ABF&Attention Fusion&MMIF&FC Layer&Uterus& CTU-UHB& 160 & 2 &  N/A\\
\hline
\cite{li2022attention} & Multi-parametric MRI&ABF&Attention Fusion&AGDAF&FC Layer&Liver& Private Data & 112 & 2 & N/A\\
\hline
\cite{dai2022mutual} &MRI (T1,T2)&ABF&Attention Fusion&MMNet&FC Layer&Parotid, Prostate& MRNet, TCIA (PROSTATEx) & 2089 & 2 & N/A\\
\hline
\cite{qiu2023hierarchical} &Genomic Data, Pathology Data&ABF&Attention Fusion&AHM-Fusion&FC Layer&Brain, Lung& TCGA (LUAD/ LUSC, GBM/ LGG) & 1670 & 2,3 & N/A \\
\hline
\added{\cite{zuo2023alzheimer}} &\added{sMRI, fMRI}&\added{ABF}&\added{Attention Fusion}&\added{SBM}&\added{FC Layer}&\added{Brain}& \added{ADNI} & \added{268} & \added{2} & \added{N/A} \\
\hline
\added{\cite{chen2023multimodal}} &\added{MRI, PET, Metadata}&\added{ABF}&\added{Attention Fusion}&\added{CBAM}&\added{FC Layer}&\added{Brain}& \added{ADNI} & \added{227} & \added{2,3} & \added{N/A} \\
\hline
\added{\cite{gao2023multimodal}} &\added{MRI, PET}&\added{ABF}&\added{Attention Fusion, Concatenation (Classic)}&\added{CNN, ViT}&\added{FC Layer}&\added{Brain}& \added{ADNI, OASIS-3} & \added{1700} & \added{2} & \added{N/A} \\
\hline
\added{\cite{bi2023multimodal}} &\added{sMRI, fMRI}&\added{ABF}&\added{Attention Fusion}&\added{ViT, CNN}&\added{FC Layer}&\added{Brain}& \added{COBRE} & \added{827} & \added{2,3,5} & \added{N/A} \\
\hline
\added{\cite{wei2024multi}} &\added{Dsc, Clinical Image}&\added{ABF}&\added{Attention Fusion, Concatenation (Classic)}&\added{CNN}&\added{FC Layer}&\added{Skin}& \added{SPC} & \added{1011} & \added{2,3,5} & \added{N/A} \\
\hline
\cite{abdolmaleki2020brain} &sMRI, fMRI&OF & Merge (Outputs) & CNN& SVM, KNN,  LDA&Brain & ADHD-200 & 730 & 2 & N/A\\
\hline
\cite{xi2020deep} &MRI (T1C, T2), Clinical Data&OF & Merge (Outputs)& ResNet& Bagging& Kidney& Private Data & 1162 & 2 & N/A\\
\hline
\cite{moon2020computer} &US (ROI, Tumor image, TSI, Fused image)&OF & Merge (Outputs)&CNN &Score Merge & Breast& BUSI \citep{al2019dataset} & 697  & 2 & N/A \\
\hline
\cite{fang2020ensemble} &MRI, PET&OF & Merge (Outputs)& CNN& AdaBoost& Brain& ADNI & 398 & 2 & N/A\\
\hline
\cite{huang2020multimodal} &CT, EMR&OF & Merge (Outputs)& DNN, CNN& Score Merge&Lung &Private Data & 1794 & 2 & N/A\\
\hline
\cite{ying2021multi} &MRI, SNP&OF & Merge (Outputs)& CNN, MLP& Ensemble Gate&Brain & ADNI & 100  & 2 & N/A\\
\hline
\cite{yoo2022deeppdt} &CFP, Clinical Data&OF & Merge (Outputs)& ResNet& XGBoost&Eye & Private Data & 166 & 2 & N/A\\
\hline
\cite{prabhu2022multi} &MRI, EHR&OF & Merge (Outputs)& CNN, AE&Score Merge & Brain& ADNI& 3996   &2,3 & N/A\\
\hline
\cite{guo2022multimodal} &MRI (TI, T2, T1C, FLAIR)&OF & Merge (Outputs)& CNN& Score Merge& Brain& CPM-RadPath 2020& 221 & 3 & N/A\\
\hline
\cite{kwon2022diagnosis} &Laryngeal image, Voice&OF & Merge (Outputs)& CNN&Decision Tree & Larynx& Private Data & 431 & 2 & N/A\\
\hline
\added{\cite{qiu2022multimodal}} &\added{MRI, Metadata}&\added{OF} & \added{Merge (Outputs)}& \added{CNN}&\added{CatBoost} & \added{Brain}& \added{OASIS-3} & \added{666} & \added{2} & \added{\footnote[18]{\url{https://github.com/vkola-lab/ncomms2022}}}\\
\hline
\cite{hu2020deep} &MRI (DCE, T2)& IF, SLF, OF  & Merge (Input), Merge (Classic), Merge (Outputs)& VGG& Score Merge& Breast& Private Data & 616 & 2 &N/A \\
\hline
\cite{wang2022combining} &WSI, MRI (T1, T1-Gd, T2, FLAIR)& IF,
OF  & Concatenation (Input), Merge (Outputs)& CNN& Score Merge& Brain& CPM-RadPath 2020&378 & 3& N/A\\
\hline
\cite{tang2022fusionm4net} &Dsc, Clinical Image, Metadata& SLF, OF  & Concatenation (Classic), Merge (Outputs)& CNN& Score Merge& Skin& SPC & 1011& 2,3,5  & N/A\\
\hline
\added{\cite{mustafa2023diagnosing}} &\added{CT, MRI}& \added{IF, OF}  & \added{Merge (Input), Merge (Outputs)} & \added{CNN}& \added{Score Merge}& \added{Brain}& \added{OASIS-3} & \added{1377}& \added{4}  & \added{N/A}\\
\hline
\end{longtable}
\clearpage
\twocolumn

\bibliographystyle{model2-names.bst}\biboptions{authoryear}
\bibliography{refs}

\begin{thebibliography}{268}
\expandafter\ifx\csname natexlab\endcsname\relax\def\natexlab#1{#1}\fi
\providecommand{\url}[1]{\texttt{#1}}
\providecommand{\href}[2]{#2}
\providecommand{\path}[1]{#1}
\providecommand{\DOIprefix}{doi:}
\providecommand{\ArXivprefix}{arXiv:}
\providecommand{\URLprefix}{URL: }
\providecommand{\Pubmedprefix}{pmid:}
\providecommand{\doi}[1]{\href{http://dx.doi.org/#1}{\path{#1}}}
\providecommand{\Pubmed}[1]{\href{pmid:#1}{\path{#1}}}
\providecommand{\bibinfo}[2]{#2}
\ifx\xfnm\relax \def\xfnm[#1]{\unskip,\space#1}\fi
\bibitem[{Abbas et~al.(2023)Abbas, Chi and Chen}]{abbas2023deepmnf}
\bibinfo{author}{Abbas, S.Q.}, \bibinfo{author}{Chi, L.}, \bibinfo{author}{Chen, Y.P.P.}, \bibinfo{year}{2023}.
\newblock \bibinfo{title}{Deepmnf: Deep multimodal neuroimaging framework for diagnosing autism spectrum disorder}.
\newblock \bibinfo{journal}{Artificial Intelligence in Medicine} \bibinfo{volume}{136}, \bibinfo{pages}{102475}.
\bibitem[{Abdelaziz et~al.(2021)Abdelaziz, Wang and Elazab}]{abdelaziz2021alzheimer}
\bibinfo{author}{Abdelaziz, M.}, \bibinfo{author}{Wang, T.}, \bibinfo{author}{Elazab, A.}, \bibinfo{year}{2021}.
\newblock \bibinfo{title}{Alzheimer’s disease diagnosis framework from incomplete multimodal data using convolutional neural networks}.
\newblock \bibinfo{journal}{Journal of Biomedical Informatics} \bibinfo{volume}{121}, \bibinfo{pages}{103863}.
\bibitem[{Abdelaziz et~al.(2022)Abdelaziz, Wang and Elazab}]{abdelaziz2022fusing}
\bibinfo{author}{Abdelaziz, M.}, \bibinfo{author}{Wang, T.}, \bibinfo{author}{Elazab, A.}, \bibinfo{year}{2022}.
\newblock \bibinfo{title}{Fusing multimodal and anatomical volumes of interest features using convolutional auto-encoder and convolutional neural networks for alzheimer’s disease diagnosis}.
\newblock \bibinfo{journal}{Frontiers in Aging Neuroscience} \bibinfo{volume}{14}.
\bibitem[{Abdelgawad et~al.(2020)Abdelgawad, Abu-samra, Abdelhay and Abdel-Azeem}]{abdelgawad2020b}
\bibinfo{author}{Abdelgawad, E.A.}, \bibinfo{author}{Abu-samra, M.F.}, \bibinfo{author}{Abdelhay, N.M.}, \bibinfo{author}{Abdel-Azeem, H.M.}, \bibinfo{year}{2020}.
\newblock \bibinfo{title}{B-mode ultrasound, color doppler, and sonoelastography in differentiation between benign and malignant cervical lymph nodes with special emphasis on sonoelastography}.
\newblock \bibinfo{journal}{Egyptian Journal of Radiology and Nuclear Medicine} \bibinfo{volume}{51}, \bibinfo{pages}{1--10}.
\bibitem[{Abdolmaleki and Abadeh(2020)}]{abdolmaleki2020brain}
\bibinfo{author}{Abdolmaleki, S.}, \bibinfo{author}{Abadeh, M.S.}, \bibinfo{year}{2020}.
\newblock \bibinfo{title}{Brain mr image classification for adhd diagnosis using deep neural networks}, in: \bibinfo{booktitle}{2020 international conference on machine vision and image processing (MVIP)}, \bibinfo{organization}{IEEE}. pp. \bibinfo{pages}{1--5}.
\bibitem[{Akbari et~al.(2021)Akbari, Yuan, Qian, Chuang, Chang, Cui and Gong}]{akbari2021vatt}
\bibinfo{author}{Akbari, H.}, \bibinfo{author}{Yuan, L.}, \bibinfo{author}{Qian, R.}, \bibinfo{author}{Chuang, W.H.}, \bibinfo{author}{Chang, S.F.}, \bibinfo{author}{Cui, Y.}, \bibinfo{author}{Gong, B.}, \bibinfo{year}{2021}.
\newblock \bibinfo{title}{Vatt: Transformers for multimodal self-supervised learning from raw video, audio and text}.
\newblock \bibinfo{journal}{Advances in Neural Information Processing Systems} \bibinfo{volume}{34}, \bibinfo{pages}{24206--24221}.
\bibitem[{Akhavan~Aghdam et~al.(2018)Akhavan~Aghdam, Sharifi and Pedram}]{akhavan2018combination}
\bibinfo{author}{Akhavan~Aghdam, M.}, \bibinfo{author}{Sharifi, A.}, \bibinfo{author}{Pedram, M.M.}, \bibinfo{year}{2018}.
\newblock \bibinfo{title}{Combination of rs-fmri and smri data to discriminate autism spectrum disorders in young children using deep belief network}.
\newblock \bibinfo{journal}{Journal of digital imaging} \bibinfo{volume}{31}, \bibinfo{pages}{895--903}.
\bibitem[{Al-Absi et~al.(2022)Al-Absi, Islam, Refaee, Chowdhury and Alam}]{al2022cardiovascular}
\bibinfo{author}{Al-Absi, H.R.}, \bibinfo{author}{Islam, M.T.}, \bibinfo{author}{Refaee, M.A.}, \bibinfo{author}{Chowdhury, M.E.}, \bibinfo{author}{Alam, T.}, \bibinfo{year}{2022}.
\newblock \bibinfo{title}{Cardiovascular disease diagnosis from dxa scan and retinal images using deep learning}.
\newblock \bibinfo{journal}{Sensors} \bibinfo{volume}{22}, \bibinfo{pages}{4310}.
\bibitem[{Al-Dhabyani et~al.(2019)Al-Dhabyani, Gomaa, Khaled and Fahmy}]{al2019dataset}
\bibinfo{author}{Al-Dhabyani, W.}, \bibinfo{author}{Gomaa, M.}, \bibinfo{author}{Khaled, H.}, \bibinfo{author}{Fahmy, A.}, \bibinfo{year}{2019}.
\newblock \bibinfo{title}{Dataset of breast ultrasound images. data brief 28, 104863 (2020)}.
\bibitem[{Aldoj et~al.(2020)Aldoj, Lukas, Dewey and Penzkofer}]{aldoj2020semi}
\bibinfo{author}{Aldoj, N.}, \bibinfo{author}{Lukas, S.}, \bibinfo{author}{Dewey, M.}, \bibinfo{author}{Penzkofer, T.}, \bibinfo{year}{2020}.
\newblock \bibinfo{title}{Semi-automatic classification of prostate cancer on multi-parametric mr imaging using a multi-channel 3d convolutional neural network}.
\newblock \bibinfo{journal}{European radiology} \bibinfo{volume}{30}, \bibinfo{pages}{1243--1253}.
\bibitem[{Andrade-Miranda et~al.(2022)Andrade-Miranda, Jaouen, Bourbonne, Lucia, Visvikis and Conze}]{andrade2022pure}
\bibinfo{author}{Andrade-Miranda, G.}, \bibinfo{author}{Jaouen, V.}, \bibinfo{author}{Bourbonne, V.}, \bibinfo{author}{Lucia, F.}, \bibinfo{author}{Visvikis, D.}, \bibinfo{author}{Conze, P.H.}, \bibinfo{year}{2022}.
\newblock \bibinfo{title}{Pure versus hybrid transformers for multi-modal brain tumor segmentation: a comparative study}, in: \bibinfo{booktitle}{2022 IEEE International Conference on Image Processing (ICIP)}, \bibinfo{organization}{IEEE}. pp. \bibinfo{pages}{1336--1340}.
\bibitem[{Andrearczyk et~al.(2022)Andrearczyk, Oreiller, Boughdad, Rest, Elhalawani, Jreige, Prior, Valli{\`e}res, Visvikis, Hatt et~al.}]{andrearczyk2022overview}
\bibinfo{author}{Andrearczyk, V.}, \bibinfo{author}{Oreiller, V.}, \bibinfo{author}{Boughdad, S.}, \bibinfo{author}{Rest, C.C.L.}, \bibinfo{author}{Elhalawani, H.}, \bibinfo{author}{Jreige, M.}, \bibinfo{author}{Prior, J.O.}, \bibinfo{author}{Valli{\`e}res, M.}, \bibinfo{author}{Visvikis, D.}, \bibinfo{author}{Hatt, M.}, et~al., \bibinfo{year}{2022}.
\newblock \bibinfo{title}{Overview of the hecktor challenge at miccai 2021: automatic head and neck tumor segmentation and outcome prediction in pet/ct images}, in: \bibinfo{booktitle}{Head and Neck Tumor Segmentation and Outcome Prediction: Second Challenge, HECKTOR 2021, Held in Conjunction with MICCAI 2021, Strasbourg, France, September 27, 2021, Proceedings}. \bibinfo{publisher}{Springer}, pp. \bibinfo{pages}{1--37}.
\bibitem[{Appalaraju et~al.(2021)Appalaraju, Jasani, Kota, Xie and Manmatha}]{appalaraju2021docformer}
\bibinfo{author}{Appalaraju, S.}, \bibinfo{author}{Jasani, B.}, \bibinfo{author}{Kota, B.U.}, \bibinfo{author}{Xie, Y.}, \bibinfo{author}{Manmatha, R.}, \bibinfo{year}{2021}.
\newblock \bibinfo{title}{Docformer: End-to-end transformer for document understanding}, in: \bibinfo{booktitle}{Proceedings of the IEEE/CVF international conference on computer vision}, pp. \bibinfo{pages}{993--1003}.
\bibitem[{Ayg{\"u}n et~al.(2018)Ayg{\"u}n, {\c{S}}ahin and {\"U}nal}]{aygun2018multi}
\bibinfo{author}{Ayg{\"u}n, M.}, \bibinfo{author}{{\c{S}}ahin, Y.H.}, \bibinfo{author}{{\"U}nal, G.}, \bibinfo{year}{2018}.
\newblock \bibinfo{title}{Multi modal convolutional neural networks for brain tumor segmentation}.
\newblock \bibinfo{journal}{arXiv preprint arXiv:1809.06191} .
\bibitem[{Azam et~al.(2021)Azam, Khan, Ahmad and Mazzara}]{azam2021multimodal}
\bibinfo{author}{Azam, M.A.}, \bibinfo{author}{Khan, K.B.}, \bibinfo{author}{Ahmad, M.}, \bibinfo{author}{Mazzara, M.}, \bibinfo{year}{2021}.
\newblock \bibinfo{title}{Multimodal medical image registration and fusion for quality enhancement}.
\newblock \bibinfo{journal}{Computers, Materials \& Continua} \bibinfo{volume}{68}, \bibinfo{pages}{821--840}.
\bibitem[{Azam et~al.(2022)Azam, Khan, Salahuddin, Rehman, Khan, Khan, Kadry and Gandomi}]{azam2022review}
\bibinfo{author}{Azam, M.A.}, \bibinfo{author}{Khan, K.B.}, \bibinfo{author}{Salahuddin, S.}, \bibinfo{author}{Rehman, E.}, \bibinfo{author}{Khan, S.A.}, \bibinfo{author}{Khan, M.A.}, \bibinfo{author}{Kadry, S.}, \bibinfo{author}{Gandomi, A.H.}, \bibinfo{year}{2022}.
\newblock \bibinfo{title}{A review on multimodal medical image fusion: Compendious analysis of medical modalities, multimodal databases, fusion techniques and quality metrics}.
\newblock \bibinfo{journal}{Computers in biology and medicine} \bibinfo{volume}{144}, \bibinfo{pages}{105253}.
\bibitem[{Bailey et~al.(2005)Bailey, Maisey, Townsend and Valk}]{bailey2005positron}
\bibinfo{author}{Bailey, D.L.}, \bibinfo{author}{Maisey, M.N.}, \bibinfo{author}{Townsend, D.W.}, \bibinfo{author}{Valk, P.E.}, \bibinfo{year}{2005}.
\newblock \bibinfo{title}{Positron emission tomography}. volume~\bibinfo{volume}{2}.
\newblock \bibinfo{publisher}{Springer}.
\bibitem[{Ballard(1987)}]{ballard1987modular}
\bibinfo{author}{Ballard, D.H.}, \bibinfo{year}{1987}.
\newblock \bibinfo{title}{Modular learning in neural networks.}, in: \bibinfo{booktitle}{Aaai}, pp. \bibinfo{pages}{279--284}.
\bibitem[{Baltru{\v{s}}aitis et~al.(2018)Baltru{\v{s}}aitis, Ahuja and Morency}]{baltruvsaitis2018multimodal}
\bibinfo{author}{Baltru{\v{s}}aitis, T.}, \bibinfo{author}{Ahuja, C.}, \bibinfo{author}{Morency, L.P.}, \bibinfo{year}{2018}.
\newblock \bibinfo{title}{Multimodal machine learning: A survey and taxonomy}.
\newblock \bibinfo{journal}{IEEE transactions on pattern analysis and machine intelligence} \bibinfo{volume}{41}, \bibinfo{pages}{423--443}.
\bibitem[{Bandettini(2012)}]{bandettini2012twenty}
\bibinfo{author}{Bandettini, P.A.}, \bibinfo{year}{2012}.
\newblock \bibinfo{title}{Twenty years of functional mri: the science and the stories}.
\newblock \bibinfo{journal}{Neuroimage} \bibinfo{volume}{62}, \bibinfo{pages}{575--588}.
\bibitem[{Bashir et~al.(2019)Bashir, Junejo, Qadri, Fleury and Qadri}]{bashir2019swt}
\bibinfo{author}{Bashir, R.}, \bibinfo{author}{Junejo, R.}, \bibinfo{author}{Qadri, N.N.}, \bibinfo{author}{Fleury, M.}, \bibinfo{author}{Qadri, M.Y.}, \bibinfo{year}{2019}.
\newblock \bibinfo{title}{Swt and pca image fusion methods for multi-modal imagery}.
\newblock \bibinfo{journal}{Multimedia Tools and Applications} \bibinfo{volume}{78}, \bibinfo{pages}{1235--1263}.
\bibitem[{Beckett et~al.(2015)Beckett, Donohue, Wang, Aisen, Harvey, Saito, Initiative et~al.}]{beckett2015alzheimer}
\bibinfo{author}{Beckett, L.A.}, \bibinfo{author}{Donohue, M.C.}, \bibinfo{author}{Wang, C.}, \bibinfo{author}{Aisen, P.}, \bibinfo{author}{Harvey, D.J.}, \bibinfo{author}{Saito, N.}, \bibinfo{author}{Initiative, A.D.N.}, et~al., \bibinfo{year}{2015}.
\newblock \bibinfo{title}{The alzheimer's disease neuroimaging initiative phase 2: Increasing the length, breadth, and depth of our understanding}.
\newblock \bibinfo{journal}{Alzheimer's \& Dementia} \bibinfo{volume}{11}, \bibinfo{pages}{823--831}.
\bibitem[{Besenczi et~al.(2016)Besenczi, T{\'o}th and Hajdu}]{besenczi2016review}
\bibinfo{author}{Besenczi, R.}, \bibinfo{author}{T{\'o}th, J.}, \bibinfo{author}{Hajdu, A.}, \bibinfo{year}{2016}.
\newblock \bibinfo{title}{A review on automatic analysis techniques for color fundus photographs}.
\newblock \bibinfo{journal}{Computational and structural biotechnology journal} \bibinfo{volume}{14}, \bibinfo{pages}{371--384}.
\bibitem[{Bhat and Koundal(2021)}]{bhat2021multi}
\bibinfo{author}{Bhat, S.}, \bibinfo{author}{Koundal, D.}, \bibinfo{year}{2021}.
\newblock \bibinfo{title}{Multi-focus image fusion techniques: a survey}.
\newblock \bibinfo{journal}{Artificial Intelligence Review} \bibinfo{volume}{54}, \bibinfo{pages}{5735--5787}.
\bibitem[{Bhatnagar et~al.(2013)Bhatnagar, Wu and Liu}]{bhatnagar2013directive}
\bibinfo{author}{Bhatnagar, G.}, \bibinfo{author}{Wu, Q.J.}, \bibinfo{author}{Liu, Z.}, \bibinfo{year}{2013}.
\newblock \bibinfo{title}{Directive contrast based multimodal medical image fusion in nsct domain}.
\newblock \bibinfo{journal}{IEEE transactions on multimedia} \bibinfo{volume}{15}, \bibinfo{pages}{1014--1024}.
\bibitem[{Bi et~al.(2023)Bi, Abrol, Fu and Calhoun}]{bi2023multimodal}
\bibinfo{author}{Bi, Y.}, \bibinfo{author}{Abrol, A.}, \bibinfo{author}{Fu, Z.}, \bibinfo{author}{Calhoun, V.}, \bibinfo{year}{2023}.
\newblock \bibinfo{title}{A multimodal vision transformer for interpretable fusion of functional and structural neuroimaging data}.
\newblock \bibinfo{journal}{bioRxiv} , \bibinfo{pages}{2023--07}.
\bibitem[{Bien et~al.(2018)Bien, Rajpurkar, Ball, Irvin, Park, Jones, Bereket, Patel, Yeom, Shpanskaya et~al.}]{bien2018deep}
\bibinfo{author}{Bien, N.}, \bibinfo{author}{Rajpurkar, P.}, \bibinfo{author}{Ball, R.L.}, \bibinfo{author}{Irvin, J.}, \bibinfo{author}{Park, A.}, \bibinfo{author}{Jones, E.}, \bibinfo{author}{Bereket, M.}, \bibinfo{author}{Patel, B.N.}, \bibinfo{author}{Yeom, K.W.}, \bibinfo{author}{Shpanskaya, K.}, et~al., \bibinfo{year}{2018}.
\newblock \bibinfo{title}{Deep-learning-assisted diagnosis for knee magnetic resonance imaging: development and retrospective validation of mrnet}.
\newblock \bibinfo{journal}{PLoS medicine} \bibinfo{volume}{15}, \bibinfo{pages}{e1002699}.
\bibitem[{Boulahia et~al.(2021)Boulahia, Amamra, Madi and Daikh}]{boulahia2021early}
\bibinfo{author}{Boulahia, S.Y.}, \bibinfo{author}{Amamra, A.}, \bibinfo{author}{Madi, M.R.}, \bibinfo{author}{Daikh, S.}, \bibinfo{year}{2021}.
\newblock \bibinfo{title}{Early, intermediate and late fusion strategies for robust deep learning-based multimodal action recognition}.
\newblock \bibinfo{journal}{Machine Vision and Applications} \bibinfo{volume}{32}, \bibinfo{pages}{121}.
\bibitem[{Buzug(2011)}]{buzug2011computed}
\bibinfo{author}{Buzug, T.M.}, \bibinfo{year}{2011}.
\newblock \bibinfo{title}{Computed tomography}.
\newblock \bibinfo{publisher}{Springer}.
\bibitem[{Cai et~al.(2022)Cai, Lin, He and Tang}]{Cai2022}
\bibinfo{author}{Cai, Z.}, \bibinfo{author}{Lin, L.}, \bibinfo{author}{He, H.}, \bibinfo{author}{Tang, X.}, \bibinfo{year}{2022}.
\newblock \bibinfo{title}{Corolla: An efficient multi-modality fusion framework with supervised contrastive learning for glaucoma grading}, in: \bibinfo{booktitle}{2022 IEEE 19th International Symposium on Biomedical Imaging (ISBI)}, pp. \bibinfo{pages}{1--4}.
\newblock \DOIprefix\doi{10.1109/ISBI52829.2022.9761712}.
\bibitem[{Calhoun and Sui(2016)}]{calhoun2016multimodal}
\bibinfo{author}{Calhoun, V.D.}, \bibinfo{author}{Sui, J.}, \bibinfo{year}{2016}.
\newblock \bibinfo{title}{Multimodal fusion of brain imaging data: a key to finding the missing link (s) in complex mental illness}.
\newblock \bibinfo{journal}{Biological psychiatry: cognitive neuroscience and neuroimaging} \bibinfo{volume}{1}, \bibinfo{pages}{230--244}.
\bibitem[{Calhoun et~al.(2012)Calhoun, Sui, Kiehl, Turner, Allen and Pearlson}]{calhoun2012exploring}
\bibinfo{author}{Calhoun, V.D.}, \bibinfo{author}{Sui, J.}, \bibinfo{author}{Kiehl, K.}, \bibinfo{author}{Turner, J.}, \bibinfo{author}{Allen, E.}, \bibinfo{author}{Pearlson, G.}, \bibinfo{year}{2012}.
\newblock \bibinfo{title}{Exploring the psychosis functional connectome: aberrant intrinsic networks in schizophrenia and bipolar disorder}.
\newblock \bibinfo{journal}{Frontiers in psychiatry} \bibinfo{volume}{2}, \bibinfo{pages}{75}.
\bibitem[{Chatzianastasis et~al.(2023)Chatzianastasis, Ilias, Askounis and Vazirgiannis}]{chatzianastasis2023neural}
\bibinfo{author}{Chatzianastasis, M.}, \bibinfo{author}{Ilias, L.}, \bibinfo{author}{Askounis, D.}, \bibinfo{author}{Vazirgiannis, M.}, \bibinfo{year}{2023}.
\newblock \bibinfo{title}{Neural architecture search with multimodal fusion methods for diagnosing dementia}.
\newblock \bibinfo{journal}{arXiv preprint arXiv:2302.05894} .
\bibitem[{Chen et~al.(2023)Chen, Guo, Xing, Chen, Yuan, Zhang and Zhang}]{chen2023multimodal}
\bibinfo{author}{Chen, H.}, \bibinfo{author}{Guo, H.}, \bibinfo{author}{Xing, L.}, \bibinfo{author}{Chen, D.}, \bibinfo{author}{Yuan, T.}, \bibinfo{author}{Zhang, Y.}, \bibinfo{author}{Zhang, X.}, \bibinfo{year}{2023}.
\newblock \bibinfo{title}{Multimodal predictive classification of alzheimer's disease based on attention-combined fusion network: Integrated neuroimaging modalities and medical examination data}.
\newblock \bibinfo{journal}{IET Image Processing} \bibinfo{volume}{17}, \bibinfo{pages}{3153--3164}.
\bibitem[{Chen et~al.(2018)Chen, Wu, DSouza, Abidin, Wism{\"u}ller and Xu}]{chen2018mri}
\bibinfo{author}{Chen, L.}, \bibinfo{author}{Wu, Y.}, \bibinfo{author}{DSouza, A.M.}, \bibinfo{author}{Abidin, A.Z.}, \bibinfo{author}{Wism{\"u}ller, A.}, \bibinfo{author}{Xu, C.}, \bibinfo{year}{2018}.
\newblock \bibinfo{title}{Mri tumor segmentation with densely connected 3d cnn}, in: \bibinfo{booktitle}{Medical Imaging 2018: Image Processing}, \bibinfo{organization}{SPIE}. pp. \bibinfo{pages}{357--364}.
\bibitem[{Chen et~al.(2021a)Chen, Lu, Weng, Chen, Williamson, Manz, Shady and Mahmood}]{chen2021multimodal}
\bibinfo{author}{Chen, R.J.}, \bibinfo{author}{Lu, M.Y.}, \bibinfo{author}{Weng, W.H.}, \bibinfo{author}{Chen, T.Y.}, \bibinfo{author}{Williamson, D.F.}, \bibinfo{author}{Manz, T.}, \bibinfo{author}{Shady, M.}, \bibinfo{author}{Mahmood, F.}, \bibinfo{year}{2021}a.
\newblock \bibinfo{title}{Multimodal co-attention transformer for survival prediction in gigapixel whole slide images}, in: \bibinfo{booktitle}{Proceedings of the IEEE/CVF International Conference on Computer Vision}, pp. \bibinfo{pages}{4015--4025}.
\bibitem[{Chen et~al.(2021b)Chen, Guhur, Schmid and Laptev}]{chen2021history}
\bibinfo{author}{Chen, S.}, \bibinfo{author}{Guhur, P.L.}, \bibinfo{author}{Schmid, C.}, \bibinfo{author}{Laptev, I.}, \bibinfo{year}{2021}b.
\newblock \bibinfo{title}{History aware multimodal transformer for vision-and-language navigation}.
\newblock \bibinfo{journal}{Advances in neural information processing systems} \bibinfo{volume}{34}, \bibinfo{pages}{5834--5847}.
\bibitem[{Cheng and Liu(2017)}]{cheng2017cnns}
\bibinfo{author}{Cheng, D.}, \bibinfo{author}{Liu, M.}, \bibinfo{year}{2017}.
\newblock \bibinfo{title}{Cnns based multi-modality classification for ad diagnosis}, in: \bibinfo{booktitle}{2017 10th international congress on image and signal processing, biomedical engineering and informatics (CISP-BMEI)}, \bibinfo{organization}{IEEE}. pp. \bibinfo{pages}{1--5}.
\bibitem[{Cheng et~al.(2022)Cheng, Liu, Kuang and Wang}]{cheng2022fully}
\bibinfo{author}{Cheng, J.}, \bibinfo{author}{Liu, J.}, \bibinfo{author}{Kuang, H.}, \bibinfo{author}{Wang, J.}, \bibinfo{year}{2022}.
\newblock \bibinfo{title}{A fully automated multimodal mri-based multi-task learning for glioma segmentation and idh genotyping}.
\newblock \bibinfo{journal}{IEEE Transactions on Medical Imaging} \bibinfo{volume}{41}, \bibinfo{pages}{1520--1532}.
\bibitem[{Chudacek et~al.(2014)Chudacek, Spilka, Bursa, Janku, Hruban, Huptych and Lhotska}]{chudavcek2014open}
\bibinfo{author}{Chudacek, V.}, \bibinfo{author}{Spilka, J.}, \bibinfo{author}{Bursa, M.}, \bibinfo{author}{Janku, P.}, \bibinfo{author}{Hruban, L.}, \bibinfo{author}{Huptych, M.}, \bibinfo{author}{Lhotska, L.}, \bibinfo{year}{2014}.
\newblock \bibinfo{title}{Open access intrapartum ctg database}.
\newblock \bibinfo{journal}{BMC pregnancy and childbirth} \bibinfo{volume}{14}, \bibinfo{pages}{1--12}.
\bibitem[{Clark et~al.(2013)Clark, Vendt, Smith, Freymann, Kirby, Koppel, Moore, Phillips, Maffitt, Pringle et~al.}]{clark2013cancer}
\bibinfo{author}{Clark, K.}, \bibinfo{author}{Vendt, B.}, \bibinfo{author}{Smith, K.}, \bibinfo{author}{Freymann, J.}, \bibinfo{author}{Kirby, J.}, \bibinfo{author}{Koppel, P.}, \bibinfo{author}{Moore, S.}, \bibinfo{author}{Phillips, S.}, \bibinfo{author}{Maffitt, D.}, \bibinfo{author}{Pringle, M.}, et~al., \bibinfo{year}{2013}.
\newblock \bibinfo{title}{The cancer imaging archive (tcia): maintaining and operating a public information repository}.
\newblock \bibinfo{journal}{Journal of digital imaging} \bibinfo{volume}{26}, \bibinfo{pages}{1045--1057}.
\bibitem[{consortium(2012)}]{adhd2012adhd}
\bibinfo{author}{consortium, A..}, \bibinfo{year}{2012}.
\newblock \bibinfo{title}{The adhd-200 consortium: a model to advance the translational potential of neuroimaging in clinical neuroscience}.
\newblock \bibinfo{journal}{Frontiers in systems neuroscience} \bibinfo{volume}{6}, \bibinfo{pages}{62}.
\bibitem[{Cui et~al.(2018)Cui, Mao, Jiang, Liu and Xiong}]{cui2018automatic}
\bibinfo{author}{Cui, S.}, \bibinfo{author}{Mao, L.}, \bibinfo{author}{Jiang, J.}, \bibinfo{author}{Liu, C.}, \bibinfo{author}{Xiong, S.}, \bibinfo{year}{2018}.
\newblock \bibinfo{title}{Automatic semantic segmentation of brain gliomas from mri images using a deep cascaded neural network}.
\newblock \bibinfo{journal}{Journal of healthcare engineering} \bibinfo{volume}{2018}.
\bibitem[{Cuingnet et~al.(2011)Cuingnet, Gerardin, Tessieras, Auzias, Leh{\'e}ricy, Habert, Chupin, Benali, Colliot, ADNI et~al.}]{cuingnet2011automatic}
\bibinfo{author}{Cuingnet, R.}, \bibinfo{author}{Gerardin, E.}, \bibinfo{author}{Tessieras, J.}, \bibinfo{author}{Auzias, G.}, \bibinfo{author}{Leh{\'e}ricy, S.}, \bibinfo{author}{Habert, M.O.}, \bibinfo{author}{Chupin, M.}, \bibinfo{author}{Benali, H.}, \bibinfo{author}{Colliot, O.}, \bibinfo{author}{ADNI}, et~al., \bibinfo{year}{2011}.
\newblock \bibinfo{title}{Automatic classification of patients with alzheimer's disease from structural mri: a comparison of ten methods using the adni database}.
\newblock \bibinfo{journal}{neuroimage} \bibinfo{volume}{56}, \bibinfo{pages}{766--781}.
\bibitem[{Dai et~al.(2021)Dai, Gao and Liu}]{dai2021transmed}
\bibinfo{author}{Dai, Y.}, \bibinfo{author}{Gao, Y.}, \bibinfo{author}{Liu, F.}, \bibinfo{year}{2021}.
\newblock \bibinfo{title}{Transmed: Transformers advance multi-modal medical image classification}.
\newblock \bibinfo{journal}{Diagnostics} \bibinfo{volume}{11}, \bibinfo{pages}{1384}.
\bibitem[{Dai et~al.(2022)Dai, Gao, Liu and Fu}]{dai2022mutual}
\bibinfo{author}{Dai, Y.}, \bibinfo{author}{Gao, Y.}, \bibinfo{author}{Liu, F.}, \bibinfo{author}{Fu, J.}, \bibinfo{year}{2022}.
\newblock \bibinfo{title}{Mutual attention-based hybrid dimensional network for multimodal imaging computer-aided diagnosis}.
\newblock \bibinfo{journal}{arXiv preprint arXiv:2201.09421} .
\bibitem[{Dalmis et~al.(2019)Dalmis, Gubern-M{\'e}rida, Vreemann, Bult, Karssemeijer, Mann and Teuwen}]{dalmis2019artificial}
\bibinfo{author}{Dalmis, M.U.}, \bibinfo{author}{Gubern-M{\'e}rida, A.}, \bibinfo{author}{Vreemann, S.}, \bibinfo{author}{Bult, P.}, \bibinfo{author}{Karssemeijer, N.}, \bibinfo{author}{Mann, R.}, \bibinfo{author}{Teuwen, J.}, \bibinfo{year}{2019}.
\newblock \bibinfo{title}{Artificial intelligence--based classification of breast lesions imaged with a multiparametric breast mri protocol with ultrafast dce-mri, t2, and dwi}.
\newblock \bibinfo{journal}{Investigative radiology} \bibinfo{volume}{54}, \bibinfo{pages}{325--332}.
\bibitem[{Das and Kundu(2013)}]{das2013neuro}
\bibinfo{author}{Das, S.}, \bibinfo{author}{Kundu, M.K.}, \bibinfo{year}{2013}.
\newblock \bibinfo{title}{A neuro-fuzzy approach for medical image fusion}.
\newblock \bibinfo{journal}{IEEE transactions on biomedical engineering} \bibinfo{volume}{60}, \bibinfo{pages}{3347--3353}.
\bibitem[{Davatzikos et~al.(2011)Davatzikos, Bhatt, Shaw, Batmanghelich and Trojanowski}]{davatzikos2011prediction}
\bibinfo{author}{Davatzikos, C.}, \bibinfo{author}{Bhatt, P.}, \bibinfo{author}{Shaw, L.M.}, \bibinfo{author}{Batmanghelich, K.N.}, \bibinfo{author}{Trojanowski, J.Q.}, \bibinfo{year}{2011}.
\newblock \bibinfo{title}{Prediction of mci to ad conversion, via mri, csf biomarkers, and pattern classification}.
\newblock \bibinfo{journal}{Neurobiology of aging} \bibinfo{volume}{32}, \bibinfo{pages}{2322--e19}.
\bibitem[{Decuyper et~al.(2021)Decuyper, Bonte, Deblaere and Van~Holen}]{decuyper2021automated}
\bibinfo{author}{Decuyper, M.}, \bibinfo{author}{Bonte, S.}, \bibinfo{author}{Deblaere, K.}, \bibinfo{author}{Van~Holen, R.}, \bibinfo{year}{2021}.
\newblock \bibinfo{title}{Automated mri based pipeline for segmentation and prediction of grade, idh mutation and 1p19q co-deletion in glioma}.
\newblock \bibinfo{journal}{Computerized Medical Imaging and Graphics} \bibinfo{volume}{88}, \bibinfo{pages}{101831}.
\bibitem[{Deng et~al.(2009)Deng, Dong, Socher, Li, Li and Fei-Fei}]{deng2009imagenet}
\bibinfo{author}{Deng, J.}, \bibinfo{author}{Dong, W.}, \bibinfo{author}{Socher, R.}, \bibinfo{author}{Li, L.J.}, \bibinfo{author}{Li, K.}, \bibinfo{author}{Fei-Fei, L.}, \bibinfo{year}{2009}.
\newblock \bibinfo{title}{Imagenet: A large-scale hierarchical image database}, in: \bibinfo{booktitle}{2009 IEEE conference on computer vision and pattern recognition}, \bibinfo{organization}{Ieee}. pp. \bibinfo{pages}{248--255}.
\bibitem[{Di~Martino et~al.(2017)Di~Martino, O’connor, Chen, Alaerts, Anderson, Assaf, Balsters, Baxter, Beggiato, Bernaerts et~al.}]{di2017enhancing}
\bibinfo{author}{Di~Martino, A.}, \bibinfo{author}{O’connor, D.}, \bibinfo{author}{Chen, B.}, \bibinfo{author}{Alaerts, K.}, \bibinfo{author}{Anderson, J.S.}, \bibinfo{author}{Assaf, M.}, \bibinfo{author}{Balsters, J.H.}, \bibinfo{author}{Baxter, L.}, \bibinfo{author}{Beggiato, A.}, \bibinfo{author}{Bernaerts, S.}, et~al., \bibinfo{year}{2017}.
\newblock \bibinfo{title}{Enhancing studies of the connectome in autism using the autism brain imaging data exchange ii}.
\newblock \bibinfo{journal}{Scientific data} \bibinfo{volume}{4}, \bibinfo{pages}{1--15}.
\bibitem[{Di~Martino et~al.(2014)Di~Martino, Yan, Li, Denio, Castellanos, Alaerts, Anderson, Assaf, Bookheimer, Dapretto et~al.}]{di2014autism}
\bibinfo{author}{Di~Martino, A.}, \bibinfo{author}{Yan, C.G.}, \bibinfo{author}{Li, Q.}, \bibinfo{author}{Denio, E.}, \bibinfo{author}{Castellanos, F.X.}, \bibinfo{author}{Alaerts, K.}, \bibinfo{author}{Anderson, J.S.}, \bibinfo{author}{Assaf, M.}, \bibinfo{author}{Bookheimer, S.Y.}, \bibinfo{author}{Dapretto, M.}, et~al., \bibinfo{year}{2014}.
\newblock \bibinfo{title}{The autism brain imaging data exchange: towards a large-scale evaluation of the intrinsic brain architecture in autism}.
\newblock \bibinfo{journal}{Molecular psychiatry} \bibinfo{volume}{19}, \bibinfo{pages}{659--667}.
\bibitem[{Dolci et~al.(2022)Dolci, Rahaman, Chen, Duan, Fu, Abrol, Menegaz and Calhoun}]{dolci2022deep}
\bibinfo{author}{Dolci, G.}, \bibinfo{author}{Rahaman, M.A.}, \bibinfo{author}{Chen, J.}, \bibinfo{author}{Duan, K.}, \bibinfo{author}{Fu, Z.}, \bibinfo{author}{Abrol, A.}, \bibinfo{author}{Menegaz, G.}, \bibinfo{author}{Calhoun, V.D.}, \bibinfo{year}{2022}.
\newblock \bibinfo{title}{A deep generative multimodal imaging genomics framework for alzheimer's disease prediction}, in: \bibinfo{booktitle}{2022 IEEE 22nd International Conference on Bioinformatics and Bioengineering (BIBE)}, \bibinfo{organization}{IEEE}. pp. \bibinfo{pages}{41--44}.
\bibitem[{Dolz et~al.(2019)Dolz, Desrosiers and Ben~Ayed}]{dolz2019ivd}
\bibinfo{author}{Dolz, J.}, \bibinfo{author}{Desrosiers, C.}, \bibinfo{author}{Ben~Ayed, I.}, \bibinfo{year}{2019}.
\newblock \bibinfo{title}{Ivd-net: Intervertebral disc localization and segmentation in mri with a multi-modal unet}, in: \bibinfo{booktitle}{Computational Methods and Clinical Applications for Spine Imaging: 5th International Workshop and Challenge, CSI 2018, Held in Conjunction with MICCAI 2018, Granada, Spain, September 16, 2018, Revised Selected Papers}, \bibinfo{organization}{Springer}. pp. \bibinfo{pages}{130--143}.
\bibitem[{Dolz et~al.(2018)Dolz, Gopinath, Yuan, Lombaert, Desrosiers and Ayed}]{dolz2018hyperdense}
\bibinfo{author}{Dolz, J.}, \bibinfo{author}{Gopinath, K.}, \bibinfo{author}{Yuan, J.}, \bibinfo{author}{Lombaert, H.}, \bibinfo{author}{Desrosiers, C.}, \bibinfo{author}{Ayed, I.B.}, \bibinfo{year}{2018}.
\newblock \bibinfo{title}{Hyperdense-net: a hyper-densely connected cnn for multi-modal image segmentation}.
\newblock \bibinfo{journal}{IEEE transactions on medical imaging} \bibinfo{volume}{38}, \bibinfo{pages}{1116--1126}.
\bibitem[{Donders et~al.(2006)Donders, Van Der~Heijden, Stijnen and Moons}]{donders2006gentle}
\bibinfo{author}{Donders, A.R.T.}, \bibinfo{author}{Van Der~Heijden, G.J.}, \bibinfo{author}{Stijnen, T.}, \bibinfo{author}{Moons, K.G.}, \bibinfo{year}{2006}.
\newblock \bibinfo{title}{A gentle introduction to imputation of missing values}.
\newblock \bibinfo{journal}{Journal of clinical epidemiology} \bibinfo{volume}{59}, \bibinfo{pages}{1087--1091}.
\bibitem[{Dubois et~al.(2007)Dubois, Feldman, Jacova, DeKosky, Barberger-Gateau, Cummings, Delacourte, Galasko, Gauthier, Jicha et~al.}]{dubois2007research}
\bibinfo{author}{Dubois, B.}, \bibinfo{author}{Feldman, H.H.}, \bibinfo{author}{Jacova, C.}, \bibinfo{author}{DeKosky, S.T.}, \bibinfo{author}{Barberger-Gateau, P.}, \bibinfo{author}{Cummings, J.}, \bibinfo{author}{Delacourte, A.}, \bibinfo{author}{Galasko, D.}, \bibinfo{author}{Gauthier, S.}, \bibinfo{author}{Jicha, G.}, et~al., \bibinfo{year}{2007}.
\newblock \bibinfo{title}{Research criteria for the diagnosis of alzheimer's disease: revising the nincds--adrda criteria}.
\newblock \bibinfo{journal}{The Lancet Neurology} \bibinfo{volume}{6}, \bibinfo{pages}{734--746}.
\bibitem[{El-Gamal et~al.(2016)El-Gamal, Elmogy and Atwan}]{el2016current}
\bibinfo{author}{El-Gamal, F.E.Z.A.}, \bibinfo{author}{Elmogy, M.}, \bibinfo{author}{Atwan, A.}, \bibinfo{year}{2016}.
\newblock \bibinfo{title}{Current trends in medical image registration and fusion}.
\newblock \bibinfo{journal}{Egyptian Informatics Journal} \bibinfo{volume}{17}, \bibinfo{pages}{99--124}.
\bibitem[{El~Habib~Daho et~al.(2023)El~Habib~Daho, Li, Zeghlache, Atse, Le~Boit{\'e}, Bonnin, Cosette, Deman, Borderie, Lepicard, Tadayoni, Cochener, Conze, Lamard and Quellec}]{elhabibdaho2023OMIA}
\bibinfo{author}{El~Habib~Daho, M.}, \bibinfo{author}{Li, Y.}, \bibinfo{author}{Zeghlache, R.}, \bibinfo{author}{Atse, Y.C.}, \bibinfo{author}{Le~Boit{\'e}, H.}, \bibinfo{author}{Bonnin, S.}, \bibinfo{author}{Cosette, D.}, \bibinfo{author}{Deman, P.}, \bibinfo{author}{Borderie, L.}, \bibinfo{author}{Lepicard, C.}, \bibinfo{author}{Tadayoni, R.}, \bibinfo{author}{Cochener, B.}, \bibinfo{author}{Conze, P.H.}, \bibinfo{author}{Lamard, M.}, \bibinfo{author}{Quellec, G.}, \bibinfo{year}{2023}.
\newblock \bibinfo{title}{Improved automatic diabetic retinopathy severity classification using deep multimodal fusion of uwf-cfp and octa images}, in: \bibinfo{editor}{Antony, B.}, \bibinfo{editor}{Chen, H.}, \bibinfo{editor}{Fang, H.}, \bibinfo{editor}{Fu, H.}, \bibinfo{editor}{Lee, C.S.}, \bibinfo{editor}{Zheng, Y.} (Eds.), \bibinfo{booktitle}{Ophthalmic Medical Image Analysis}, \bibinfo{publisher}{Springer Nature Switzerland}, \bibinfo{address}{Cham}. pp. \bibinfo{pages}{11--20}.
\bibitem[{El-Sappagh et~al.(2020)El-Sappagh, Abuhmed, Islam and Kwak}]{el2020multimodal}
\bibinfo{author}{El-Sappagh, S.}, \bibinfo{author}{Abuhmed, T.}, \bibinfo{author}{Islam, S.R.}, \bibinfo{author}{Kwak, K.S.}, \bibinfo{year}{2020}.
\newblock \bibinfo{title}{Multimodal multitask deep learning model for alzheimer’s disease progression detection based on time series data}.
\newblock \bibinfo{journal}{Neurocomputing} \bibinfo{volume}{412}, \bibinfo{pages}{197--215}.
\bibitem[{Fang et~al.(2020)Fang, Liu and Xu}]{fang2020ensemble}
\bibinfo{author}{Fang, X.}, \bibinfo{author}{Liu, Z.}, \bibinfo{author}{Xu, M.}, \bibinfo{year}{2020}.
\newblock \bibinfo{title}{Ensemble of deep convolutional neural networks based multi-modality images for alzheimer's disease diagnosis}.
\newblock \bibinfo{journal}{IET Image Processing} \bibinfo{volume}{14}, \bibinfo{pages}{318--326}.
\bibitem[{Feng et~al.(2019)Feng, Elazab, Yang, Wang, Zhou, Hu, Xiao and Lei}]{feng2019deep}
\bibinfo{author}{Feng, C.}, \bibinfo{author}{Elazab, A.}, \bibinfo{author}{Yang, P.}, \bibinfo{author}{Wang, T.}, \bibinfo{author}{Zhou, F.}, \bibinfo{author}{Hu, H.}, \bibinfo{author}{Xiao, X.}, \bibinfo{author}{Lei, B.}, \bibinfo{year}{2019}.
\newblock \bibinfo{title}{Deep learning framework for alzheimer’s disease diagnosis via 3d-cnn and fsbi-lstm}.
\newblock \bibinfo{journal}{IEEE Access} \bibinfo{volume}{7}, \bibinfo{pages}{63605--63618}.
\bibitem[{Gao et~al.(2021)Gao, Shi, Shen and Liu}]{gao2021task}
\bibinfo{author}{Gao, X.}, \bibinfo{author}{Shi, F.}, \bibinfo{author}{Shen, D.}, \bibinfo{author}{Liu, M.}, \bibinfo{year}{2021}.
\newblock \bibinfo{title}{Task-induced pyramid and attention gan for multimodal brain image imputation and classification in alzheimer's disease}.
\newblock \bibinfo{journal}{IEEE journal of biomedical and health informatics} \bibinfo{volume}{26}, \bibinfo{pages}{36--43}.
\bibitem[{Gao et~al.(2023)Gao, Shi, Shen and Liu}]{gao2023multimodal}
\bibinfo{author}{Gao, X.}, \bibinfo{author}{Shi, F.}, \bibinfo{author}{Shen, D.}, \bibinfo{author}{Liu, M.}, \bibinfo{year}{2023}.
\newblock \bibinfo{title}{Multimodal transformer network for incomplete image generation and diagnosis of alzheimer’s disease}.
\newblock \bibinfo{journal}{Computerized Medical Imaging and Graphics} \bibinfo{volume}{110}, \bibinfo{pages}{102303}.
\bibitem[{Ge et~al.(2018)Ge, Gu, Jakola and Yang}]{ge2018deep}
\bibinfo{author}{Ge, C.}, \bibinfo{author}{Gu, I.Y.H.}, \bibinfo{author}{Jakola, A.S.}, \bibinfo{author}{Yang, J.}, \bibinfo{year}{2018}.
\newblock \bibinfo{title}{Deep learning and multi-sensor fusion for glioma classification using multistream 2d convolutional networks}, in: \bibinfo{booktitle}{2018 40th Annual international conference of the IEEE engineering in medicine and biology society (EMBC)}, \bibinfo{organization}{IEEE}. pp. \bibinfo{pages}{5894--5897}.
\bibitem[{Ge et~al.(2017)Ge, Demyanov, Chakravorty, Bowling and Garnavi}]{ge2017skin}
\bibinfo{author}{Ge, Z.}, \bibinfo{author}{Demyanov, S.}, \bibinfo{author}{Chakravorty, R.}, \bibinfo{author}{Bowling, A.}, \bibinfo{author}{Garnavi, R.}, \bibinfo{year}{2017}.
\newblock \bibinfo{title}{Skin disease recognition using deep saliency features and multimodal learning of dermoscopy and clinical images}, in: \bibinfo{booktitle}{Medical Image Computing and Computer Assisted Intervention- MICCAI 2017: 20th International Conference, Quebec City, QC, Canada, September 11-13, 2017, Proceedings, Part III 20}, \bibinfo{organization}{Springer}. pp. \bibinfo{pages}{250--258}.
\bibitem[{Goodfellow et~al.(2016)Goodfellow, Bengio and Courville}]{goodfellow2016deep}
\bibinfo{author}{Goodfellow, I.}, \bibinfo{author}{Bengio, Y.}, \bibinfo{author}{Courville, A.}, \bibinfo{year}{2016}.
\newblock \bibinfo{title}{Deep learning}.
\newblock \bibinfo{publisher}{MIT press}.
\bibitem[{Goodfellow et~al.(2020)Goodfellow, Pouget-Abadie, Mirza, Xu, Warde-Farley, Ozair, Courville and Bengio}]{goodfellow2020generative}
\bibinfo{author}{Goodfellow, I.}, \bibinfo{author}{Pouget-Abadie, J.}, \bibinfo{author}{Mirza, M.}, \bibinfo{author}{Xu, B.}, \bibinfo{author}{Warde-Farley, D.}, \bibinfo{author}{Ozair, S.}, \bibinfo{author}{Courville, A.}, \bibinfo{author}{Bengio, Y.}, \bibinfo{year}{2020}.
\newblock \bibinfo{title}{Generative adversarial networks}.
\newblock \bibinfo{journal}{Communications of the ACM} \bibinfo{volume}{63}, \bibinfo{pages}{139--144}.
\bibitem[{Goodfellow et~al.(2014)Goodfellow, Pouget-Abadie, Mirza, Xu, Warde-Farley, Ozair, Courville and Bengio}]{https://doi.org/10.48550/arxiv.1406.2661}
\bibinfo{author}{Goodfellow, I.J.}, \bibinfo{author}{Pouget-Abadie, J.}, \bibinfo{author}{Mirza, M.}, \bibinfo{author}{Xu, B.}, \bibinfo{author}{Warde-Farley, D.}, \bibinfo{author}{Ozair, S.}, \bibinfo{author}{Courville, A.}, \bibinfo{author}{Bengio, Y.}, \bibinfo{year}{2014}.
\newblock \bibinfo{title}{Generative adversarial networks}.
\newblock \URLprefix \url{https://arxiv.org/abs/1406.2661}, \DOIprefix\doi{10.48550/ARXIV.1406.2661}.
\bibitem[{Gravina et~al.(2024)Gravina, Garc{\'\i}a-Pedrero, Gonzalo-Mart{\'\i}n, Sansone and Soda}]{gravina2024multi}
\bibinfo{author}{Gravina, M.}, \bibinfo{author}{Garc{\'\i}a-Pedrero, A.}, \bibinfo{author}{Gonzalo-Mart{\'\i}n, C.}, \bibinfo{author}{Sansone, C.}, \bibinfo{author}{Soda, P.}, \bibinfo{year}{2024}.
\newblock \bibinfo{title}{Multi input--multi output 3d cnn for dementia severity assessment with incomplete multimodal data}.
\newblock \bibinfo{journal}{Artificial Intelligence in Medicine} \bibinfo{volume}{149}, \bibinfo{pages}{102774}.
\bibitem[{Guo et~al.(2022)Guo, Wang, Chen, Wang, Zhang and Zhu}]{guo2022multimodal}
\bibinfo{author}{Guo, S.}, \bibinfo{author}{Wang, L.}, \bibinfo{author}{Chen, Q.}, \bibinfo{author}{Wang, L.}, \bibinfo{author}{Zhang, J.}, \bibinfo{author}{Zhu, Y.}, \bibinfo{year}{2022}.
\newblock \bibinfo{title}{Multimodal mri image decision fusion-based network for glioma classification}.
\newblock \bibinfo{journal}{Frontiers in Oncology} \bibinfo{volume}{12}.
\bibitem[{Gutiérrez et~al.(2022)Gutiérrez, Arevalo and Martánez}]{Gutiérrez2022}
\bibinfo{author}{Gutiérrez, Y.}, \bibinfo{author}{Arevalo, J.}, \bibinfo{author}{Martánez, F.}, \bibinfo{year}{2022}.
\newblock \bibinfo{title}{Multimodal contrastive supervised learning to classify clinical significance mri regions on prostate cancer}, in: \bibinfo{booktitle}{2022 44th Annual International Conference of the IEEE Engineering in Medicine \& Biology Society (EMBC)}, pp. \bibinfo{pages}{1682--1685}.
\newblock \DOIprefix\doi{10.1109/EMBC48229.2022.9871243}.
\bibitem[{Hager et~al.(2023)Hager, Menten and Rueckert}]{hager2023best}
\bibinfo{author}{Hager, P.}, \bibinfo{author}{Menten, M.J.}, \bibinfo{author}{Rueckert, D.}, \bibinfo{year}{2023}.
\newblock \bibinfo{title}{Best of both worlds: Multimodal contrastive learning with tabular and imaging data}.
\newblock \href{http://arxiv.org/abs/2303.14080}{\tt arXiv:2303.14080}.
\bibitem[{He et~al.(2010)He, Liu, Li and Wang}]{he2010multimodal}
\bibinfo{author}{He, C.}, \bibinfo{author}{Liu, Q.}, \bibinfo{author}{Li, H.}, \bibinfo{author}{Wang, H.}, \bibinfo{year}{2010}.
\newblock \bibinfo{title}{Multimodal medical image fusion based on ihs and pca}.
\newblock \bibinfo{journal}{Procedia Engineering} \bibinfo{volume}{7}, \bibinfo{pages}{280--285}.
\bibitem[{He et~al.(2015)He, Zhang, Ren and Sun}]{https://doi.org/10.48550/arxiv.1512.03385}
\bibinfo{author}{He, K.}, \bibinfo{author}{Zhang, X.}, \bibinfo{author}{Ren, S.}, \bibinfo{author}{Sun, J.}, \bibinfo{year}{2015}.
\newblock \bibinfo{title}{Deep residual learning for image recognition}.
\newblock \URLprefix \url{https://arxiv.org/abs/1512.03385}, \DOIprefix\doi{10.48550/ARXIV.1512.03385}.
\bibitem[{He et~al.(2021)He, Han, Zhang and Chen}]{he2021hierarchical}
\bibinfo{author}{He, M.}, \bibinfo{author}{Han, K.}, \bibinfo{author}{Zhang, Y.}, \bibinfo{author}{Chen, W.}, \bibinfo{year}{2021}.
\newblock \bibinfo{title}{Hierarchical-order multimodal interaction fusion network for grading gliomas}.
\newblock \bibinfo{journal}{Physics in Medicine \& Biology} \bibinfo{volume}{66}, \bibinfo{pages}{215016}.
\bibitem[{Hecht et~al.(2001)Hecht, Fellner, Fellner, Hilz, Heuss and Neund{\"o}rfer}]{hecht2001mri}
\bibinfo{author}{Hecht, M.}, \bibinfo{author}{Fellner, F.}, \bibinfo{author}{Fellner, C.}, \bibinfo{author}{Hilz, M.}, \bibinfo{author}{Heuss, D.}, \bibinfo{author}{Neund{\"o}rfer, B.}, \bibinfo{year}{2001}.
\newblock \bibinfo{title}{Mri-flair images of the head show corticospinal tract alterations in als patients more frequently than t2-, t1-and proton-density-weighted images}.
\newblock \bibinfo{journal}{Journal of the neurological sciences} \bibinfo{volume}{186}, \bibinfo{pages}{37--44}.
\bibitem[{Hermessi et~al.(2021)Hermessi, Mourali and Zagrouba}]{hermessi2021multimodal}
\bibinfo{author}{Hermessi, H.}, \bibinfo{author}{Mourali, O.}, \bibinfo{author}{Zagrouba, E.}, \bibinfo{year}{2021}.
\newblock \bibinfo{title}{Multimodal medical image fusion review: Theoretical background and recent advances}.
\newblock \bibinfo{journal}{Signal Processing} \bibinfo{volume}{183}, \bibinfo{pages}{108036}.
\bibitem[{Hoang~Nguyen et~al.(2022)Hoang~Nguyen, Blaschko, Saarakkala and Tiulpin}]{hoang2022clinically}
\bibinfo{author}{Hoang~Nguyen, H.}, \bibinfo{author}{Blaschko, M.B.}, \bibinfo{author}{Saarakkala, S.}, \bibinfo{author}{Tiulpin, A.}, \bibinfo{year}{2022}.
\newblock \bibinfo{title}{Clinically-inspired multi-agent transformers for disease trajectory forecasting from multimodal data}.
\newblock \bibinfo{journal}{arXiv e-prints} , \bibinfo{pages}{arXiv--2210}.
\bibitem[{Hsu et~al.(2022)Hsu, Guo, Pei, Chiang, Li, Hsiao, Colen and Liu}]{hsu2022weakly}
\bibinfo{author}{Hsu, W.W.}, \bibinfo{author}{Guo, J.M.}, \bibinfo{author}{Pei, L.}, \bibinfo{author}{Chiang, L.A.}, \bibinfo{author}{Li, Y.F.}, \bibinfo{author}{Hsiao, J.C.}, \bibinfo{author}{Colen, R.}, \bibinfo{author}{Liu, P.}, \bibinfo{year}{2022}.
\newblock \bibinfo{title}{A weakly supervised deep learning-based method for glioma subtype classification using wsi and mpmris}.
\newblock \bibinfo{journal}{Scientific Reports} \bibinfo{volume}{12}, \bibinfo{pages}{6111}.
\bibitem[{Hu et~al.(2020)Hu, Whitney and Giger}]{hu2020deep}
\bibinfo{author}{Hu, Q.}, \bibinfo{author}{Whitney, H.M.}, \bibinfo{author}{Giger, M.L.}, \bibinfo{year}{2020}.
\newblock \bibinfo{title}{A deep learning methodology for improved breast cancer diagnosis using multiparametric mri}.
\newblock \bibinfo{journal}{Scientific reports} \bibinfo{volume}{10}, \bibinfo{pages}{10536}.
\bibitem[{Huang et~al.(1991)Huang, Swanson, Lin, Schuman, Stinson, Chang, Hee, Flotte, Gregory, Puliafito et~al.}]{huang1991optical}
\bibinfo{author}{Huang, D.}, \bibinfo{author}{Swanson, E.A.}, \bibinfo{author}{Lin, C.P.}, \bibinfo{author}{Schuman, J.S.}, \bibinfo{author}{Stinson, W.G.}, \bibinfo{author}{Chang, W.}, \bibinfo{author}{Hee, M.R.}, \bibinfo{author}{Flotte, T.}, \bibinfo{author}{Gregory, K.}, \bibinfo{author}{Puliafito, C.A.}, et~al., \bibinfo{year}{1991}.
\newblock \bibinfo{title}{Optical coherence tomography}.
\newblock \bibinfo{journal}{science} \bibinfo{volume}{254}, \bibinfo{pages}{1178--1181}.
\bibitem[{Huang et~al.(2016)Huang, Liu, van~der Maaten and Weinberger}]{https://doi.org/10.48550/arxiv.1608.06993}
\bibinfo{author}{Huang, G.}, \bibinfo{author}{Liu, Z.}, \bibinfo{author}{van~der Maaten, L.}, \bibinfo{author}{Weinberger, K.Q.}, \bibinfo{year}{2016}.
\newblock \bibinfo{title}{Densely connected convolutional networks}.
\newblock \URLprefix \url{https://arxiv.org/abs/1608.06993}, \DOIprefix\doi{10.48550/ARXIV.1608.06993}.
\bibitem[{Huang et~al.(2020)Huang, Pareek, Zamanian, Banerjee and Lungren}]{huang2020multimodal}
\bibinfo{author}{Huang, S.C.}, \bibinfo{author}{Pareek, A.}, \bibinfo{author}{Zamanian, R.}, \bibinfo{author}{Banerjee, I.}, \bibinfo{author}{Lungren, M.P.}, \bibinfo{year}{2020}.
\newblock \bibinfo{title}{Multimodal fusion with deep neural networks for leveraging ct imaging and electronic health record: a case-study in pulmonary embolism detection}.
\newblock \bibinfo{journal}{Scientific reports} \bibinfo{volume}{10}, \bibinfo{pages}{1--9}.
\bibitem[{Huang et~al.(2022)Huang, Sun, Gupta, Montesano, Crabb, Garway-Heath, Brusini, Lanzetta, Oddone, Turpin et~al.}]{huang2022detecting}
\bibinfo{author}{Huang, X.}, \bibinfo{author}{Sun, J.}, \bibinfo{author}{Gupta, K.}, \bibinfo{author}{Montesano, G.}, \bibinfo{author}{Crabb, D.P.}, \bibinfo{author}{Garway-Heath, D.F.}, \bibinfo{author}{Brusini, P.}, \bibinfo{author}{Lanzetta, P.}, \bibinfo{author}{Oddone, F.}, \bibinfo{author}{Turpin, A.}, et~al., \bibinfo{year}{2022}.
\newblock \bibinfo{title}{Detecting glaucoma from multi-modal data using probabilistic deep learning}.
\newblock \bibinfo{journal}{Frontiers in Medicine} \bibinfo{volume}{9}.
\bibitem[{Huang et~al.(2019)Huang, Xu, Zhou, Tong, Zhuang and ADNI}]{huang2019diagnosis}
\bibinfo{author}{Huang, Y.}, \bibinfo{author}{Xu, J.}, \bibinfo{author}{Zhou, Y.}, \bibinfo{author}{Tong, T.}, \bibinfo{author}{Zhuang, X.}, \bibinfo{author}{ADNI}, \bibinfo{year}{2019}.
\newblock \bibinfo{title}{Diagnosis of alzheimer’s disease via multi-modality 3d convolutional neural network}.
\newblock \bibinfo{journal}{Frontiers in neuroscience} \bibinfo{volume}{13}, \bibinfo{pages}{509}.
\bibitem[{Huang et~al.(2023)Huang, Bianchi, Yuksekgonul, Montine and Zou}]{huang2023visual}
\bibinfo{author}{Huang, Z.}, \bibinfo{author}{Bianchi, F.}, \bibinfo{author}{Yuksekgonul, M.}, \bibinfo{author}{Montine, T.J.}, \bibinfo{author}{Zou, J.}, \bibinfo{year}{2023}.
\newblock \bibinfo{title}{A visual--language foundation model for pathology image analysis using medical twitter}.
\newblock \bibinfo{journal}{Nature medicine} \bibinfo{volume}{29}, \bibinfo{pages}{2307--2316}.
\bibitem[{Isensee et~al.(2018)Isensee, Kickingereder, Wick, Bendszus and Maier-Hein}]{isensee2018brain}
\bibinfo{author}{Isensee, F.}, \bibinfo{author}{Kickingereder, P.}, \bibinfo{author}{Wick, W.}, \bibinfo{author}{Bendszus, M.}, \bibinfo{author}{Maier-Hein, K.H.}, \bibinfo{year}{2018}.
\newblock \bibinfo{title}{Brain tumor segmentation and radiomics survival prediction: Contribution to the brats 2017 challenge}, in: \bibinfo{booktitle}{Brainlesion: Glioma, Multiple Sclerosis, Stroke and Traumatic Brain Injuries: Third International Workshop, BrainLes 2017, Held in Conjunction with MICCAI 2017, Quebec City, QC, Canada, September 14, 2017, Revised Selected Papers 3}, \bibinfo{organization}{Springer}. pp. \bibinfo{pages}{287--297}.
\bibitem[{Isensee et~al.(2019)Isensee, Kickingereder, Wick, Bendszus and Maier-Hein}]{isensee2019no}
\bibinfo{author}{Isensee, F.}, \bibinfo{author}{Kickingereder, P.}, \bibinfo{author}{Wick, W.}, \bibinfo{author}{Bendszus, M.}, \bibinfo{author}{Maier-Hein, K.H.}, \bibinfo{year}{2019}.
\newblock \bibinfo{title}{No new-net}, in: \bibinfo{booktitle}{Brainlesion: Glioma, Multiple Sclerosis, Stroke and Traumatic Brain Injuries: 4th International Workshop, BrainLes 2018, Held in Conjunction with MICCAI 2018, Granada, Spain, September 16, 2018, Revised Selected Papers, Part II 4}, \bibinfo{organization}{Springer}. pp. \bibinfo{pages}{234--244}.
\bibitem[{Jin et~al.(2022)Jin, Zhao, Zhao, Che and Li}]{jin2022hybrid}
\bibinfo{author}{Jin, L.}, \bibinfo{author}{Zhao, K.}, \bibinfo{author}{Zhao, Y.}, \bibinfo{author}{Che, T.}, \bibinfo{author}{Li, S.}, \bibinfo{year}{2022}.
\newblock \bibinfo{title}{A hybrid deep learning method for early and late mild cognitive impairment diagnosis with incomplete multimodal data}.
\newblock \bibinfo{journal}{Frontiers in Neuroinformatics} \bibinfo{volume}{16}.
\bibitem[{Joo et~al.(2021)Joo, Ko, Kwon, Jeon, Jung, Kim, Chung and Im}]{joo2021multimodal}
\bibinfo{author}{Joo, S.}, \bibinfo{author}{Ko, E.S.}, \bibinfo{author}{Kwon, S.}, \bibinfo{author}{Jeon, E.}, \bibinfo{author}{Jung, H.}, \bibinfo{author}{Kim, J.Y.}, \bibinfo{author}{Chung, M.J.}, \bibinfo{author}{Im, Y.H.}, \bibinfo{year}{2021}.
\newblock \bibinfo{title}{Multimodal deep learning models for the prediction of pathologic response to neoadjuvant chemotherapy in breast cancer}.
\newblock \bibinfo{journal}{Scientific reports} \bibinfo{volume}{11}, \bibinfo{pages}{18800}.
\bibitem[{Kadri et~al.(2023)Kadri, Bouaziz, Tmar and Gargouri}]{kadri2023efficient}
\bibinfo{author}{Kadri, R.}, \bibinfo{author}{Bouaziz, B.}, \bibinfo{author}{Tmar, M.}, \bibinfo{author}{Gargouri, F.}, \bibinfo{year}{2023}.
\newblock \bibinfo{title}{Efficient multimodel method based on transformers and coatnet for alzheimer's diagnosis}.
\newblock \bibinfo{journal}{Digital Signal Processing} \bibinfo{volume}{143}, \bibinfo{pages}{104229}.
\bibitem[{Kamnitsas et~al.(2018)Kamnitsas, Bai, Ferrante, McDonagh, Sinclair, Pawlowski, Rajchl, Lee, Kainz, Rueckert et~al.}]{kamnitsas2018ensembles}
\bibinfo{author}{Kamnitsas, K.}, \bibinfo{author}{Bai, W.}, \bibinfo{author}{Ferrante, E.}, \bibinfo{author}{McDonagh, S.}, \bibinfo{author}{Sinclair, M.}, \bibinfo{author}{Pawlowski, N.}, \bibinfo{author}{Rajchl, M.}, \bibinfo{author}{Lee, M.}, \bibinfo{author}{Kainz, B.}, \bibinfo{author}{Rueckert, D.}, et~al., \bibinfo{year}{2018}.
\newblock \bibinfo{title}{Ensembles of multiple models and architectures for robust brain tumour segmentation}, in: \bibinfo{booktitle}{Brainlesion: Glioma, Multiple Sclerosis, Stroke and Traumatic Brain Injuries: Third International Workshop, BrainLes 2017, Held in Conjunction with MICCAI 2017, Quebec City, QC, Canada, September 14, 2017, Revised Selected Papers 3}, \bibinfo{organization}{Springer}. pp. \bibinfo{pages}{450--462}.
\bibitem[{Kamnitsas et~al.(2017)Kamnitsas, Ledig, Newcombe, Simpson, Kane, Menon, Rueckert and Glocker}]{kamnitsas2017efficient}
\bibinfo{author}{Kamnitsas, K.}, \bibinfo{author}{Ledig, C.}, \bibinfo{author}{Newcombe, V.F.}, \bibinfo{author}{Simpson, J.P.}, \bibinfo{author}{Kane, A.D.}, \bibinfo{author}{Menon, D.K.}, \bibinfo{author}{Rueckert, D.}, \bibinfo{author}{Glocker, B.}, \bibinfo{year}{2017}.
\newblock \bibinfo{title}{Efficient multi-scale 3d cnn with fully connected crf for accurate brain lesion segmentation}.
\newblock \bibinfo{journal}{Medical image analysis} \bibinfo{volume}{36}, \bibinfo{pages}{61--78}.
\bibitem[{Kawahara et~al.(2018)Kawahara, Daneshvar, Argenziano and Hamarneh}]{kawahara2018seven}
\bibinfo{author}{Kawahara, J.}, \bibinfo{author}{Daneshvar, S.}, \bibinfo{author}{Argenziano, G.}, \bibinfo{author}{Hamarneh, G.}, \bibinfo{year}{2018}.
\newblock \bibinfo{title}{Seven-point checklist and skin lesion classification using multitask multimodal neural nets}.
\newblock \bibinfo{journal}{IEEE journal of biomedical and health informatics} \bibinfo{volume}{23}, \bibinfo{pages}{538--546}.
\bibitem[{Khagi and Kwon(2020)}]{khagi20203d}
\bibinfo{author}{Khagi, B.}, \bibinfo{author}{Kwon, G.R.}, \bibinfo{year}{2020}.
\newblock \bibinfo{title}{3d cnn design for the classification of alzheimer’s disease using brain mri and pet}.
\newblock \bibinfo{journal}{IEEE Access} \bibinfo{volume}{8}, \bibinfo{pages}{217830--217847}.
\bibitem[{Kim and Lee(2018)}]{kim2018identification}
\bibinfo{author}{Kim, J.}, \bibinfo{author}{Lee, B.}, \bibinfo{year}{2018}.
\newblock \bibinfo{title}{Identification of alzheimer's disease and mild cognitive impairment using multimodal sparse hierarchical extreme learning machine}.
\newblock \bibinfo{journal}{Human brain mapping} \bibinfo{volume}{39}, \bibinfo{pages}{3728--3741}.
\bibitem[{Kline et~al.(2022)Kline, Wang, Li, Dennis, Hutch, Xu, Wang, Cheng and Luo}]{kline2022multimodal}
\bibinfo{author}{Kline, A.}, \bibinfo{author}{Wang, H.}, \bibinfo{author}{Li, Y.}, \bibinfo{author}{Dennis, S.}, \bibinfo{author}{Hutch, M.}, \bibinfo{author}{Xu, Z.}, \bibinfo{author}{Wang, F.}, \bibinfo{author}{Cheng, F.}, \bibinfo{author}{Luo, Y.}, \bibinfo{year}{2022}.
\newblock \bibinfo{title}{Multimodal machine learning in precision health: A scoping review}.
\newblock \bibinfo{journal}{npj Digital Medicine} \bibinfo{volume}{5}, \bibinfo{pages}{171}.
\bibitem[{Kohannim et~al.(2010)Kohannim, Hua, Hibar, Lee, Chou, Toga, Jack~Jr, Weiner, Thompson, ADNI et~al.}]{kohannim2010boosting}
\bibinfo{author}{Kohannim, O.}, \bibinfo{author}{Hua, X.}, \bibinfo{author}{Hibar, D.P.}, \bibinfo{author}{Lee, S.}, \bibinfo{author}{Chou, Y.Y.}, \bibinfo{author}{Toga, A.W.}, \bibinfo{author}{Jack~Jr, C.R.}, \bibinfo{author}{Weiner, M.W.}, \bibinfo{author}{Thompson, P.M.}, \bibinfo{author}{ADNI}, et~al., \bibinfo{year}{2010}.
\newblock \bibinfo{title}{Boosting power for clinical trials using classifiers based on multiple biomarkers}.
\newblock \bibinfo{journal}{Neurobiology of aging} \bibinfo{volume}{31}, \bibinfo{pages}{1429--1442}.
\bibitem[{Kollias et~al.(2023)Kollias, Vendal, Gadhavi and Russom}]{kollias2023btdnet}
\bibinfo{author}{Kollias, D.}, \bibinfo{author}{Vendal, K.}, \bibinfo{author}{Gadhavi, P.}, \bibinfo{author}{Russom, S.}, \bibinfo{year}{2023}.
\newblock \bibinfo{title}{Btdnet: A multi-modal approach for brain tumor radiogenomic classification}.
\newblock \bibinfo{journal}{Applied Sciences} \bibinfo{volume}{13}, \bibinfo{pages}{11984}.
\bibitem[{Kong et~al.(2022)Kong, Zhang, Zhu, Yi, Wang and Zhang}]{kong2022multi}
\bibinfo{author}{Kong, Z.}, \bibinfo{author}{Zhang, M.}, \bibinfo{author}{Zhu, W.}, \bibinfo{author}{Yi, Y.}, \bibinfo{author}{Wang, T.}, \bibinfo{author}{Zhang, B.}, \bibinfo{year}{2022}.
\newblock \bibinfo{title}{Multi-modal data alzheimer’s disease detection based on 3d convolution}.
\newblock \bibinfo{journal}{Biomedical Signal Processing and Control} \bibinfo{volume}{75}, \bibinfo{pages}{103565}.
\bibitem[{Krizhevsky et~al.(2017)Krizhevsky, Sutskever and Hinton}]{krizhevsky2017imagenet}
\bibinfo{author}{Krizhevsky, A.}, \bibinfo{author}{Sutskever, I.}, \bibinfo{author}{Hinton, G.E.}, \bibinfo{year}{2017}.
\newblock \bibinfo{title}{Imagenet classification with deep convolutional neural networks}.
\newblock \bibinfo{journal}{Communications of the ACM} \bibinfo{volume}{60}, \bibinfo{pages}{84--90}.
\bibitem[{Kuban et~al.(2003)Kuban, Thames, Levy, Horwitz, Kupelian, Martinez, Michalski, Pisansky, Sandler, Shipley et~al.}]{kuban2003long}
\bibinfo{author}{Kuban, D.A.}, \bibinfo{author}{Thames, H.D.}, \bibinfo{author}{Levy, L.B.}, \bibinfo{author}{Horwitz, E.M.}, \bibinfo{author}{Kupelian, P.A.}, \bibinfo{author}{Martinez, A.A.}, \bibinfo{author}{Michalski, J.M.}, \bibinfo{author}{Pisansky, T.M.}, \bibinfo{author}{Sandler, H.M.}, \bibinfo{author}{Shipley, W.U.}, et~al., \bibinfo{year}{2003}.
\newblock \bibinfo{title}{Long-term multi-institutional analysis of stage t1--t2 prostate cancer treated with radiotherapy in the psa era}.
\newblock \bibinfo{journal}{International Journal of Radiation Oncology* Biology* Physics} \bibinfo{volume}{57}, \bibinfo{pages}{915--928}.
\bibitem[{Kurc et~al.(2020)Kurc, Bakas, Ren, Bagari, Momeni, Huang, Zhang, Kumar, Thibault, Qi et~al.}]{kurc2020segmentation}
\bibinfo{author}{Kurc, T.}, \bibinfo{author}{Bakas, S.}, \bibinfo{author}{Ren, X.}, \bibinfo{author}{Bagari, A.}, \bibinfo{author}{Momeni, A.}, \bibinfo{author}{Huang, Y.}, \bibinfo{author}{Zhang, L.}, \bibinfo{author}{Kumar, A.}, \bibinfo{author}{Thibault, M.}, \bibinfo{author}{Qi, Q.}, et~al., \bibinfo{year}{2020}.
\newblock \bibinfo{title}{Segmentation and classification in digital pathology for glioma research: challenges and deep learning approaches}.
\newblock \bibinfo{journal}{Frontiers in neuroscience} \bibinfo{volume}{14}, \bibinfo{pages}{27}.
\bibitem[{Kwon et~al.(2022)Kwon, Wang, Shin, Cheon, Lee, Lee, Lim, Jo, Cho and Shin}]{kwon2022diagnosis}
\bibinfo{author}{Kwon, I.}, \bibinfo{author}{Wang, S.G.}, \bibinfo{author}{Shin, S.C.}, \bibinfo{author}{Cheon, Y.I.}, \bibinfo{author}{Lee, B.J.}, \bibinfo{author}{Lee, J.C.}, \bibinfo{author}{Lim, D.W.}, \bibinfo{author}{Jo, C.}, \bibinfo{author}{Cho, Y.}, \bibinfo{author}{Shin, B.J.}, \bibinfo{year}{2022}.
\newblock \bibinfo{title}{Diagnosis of early glottic cancer using laryngeal image and voice based on ensemble learning of convolutional neural network classifiers}.
\newblock \bibinfo{journal}{Journal of Voice} .
\bibitem[{Lalousis et~al.(2021)Lalousis, Wood, Schmaal, Chisholm, Griffiths, Reniers, Bertolino, Borgwardt, Brambilla, Kambeitz et~al.}]{lalousis2021heterogeneity}
\bibinfo{author}{Lalousis, P.A.}, \bibinfo{author}{Wood, S.J.}, \bibinfo{author}{Schmaal, L.}, \bibinfo{author}{Chisholm, K.}, \bibinfo{author}{Griffiths, S.L.}, \bibinfo{author}{Reniers, R.L.}, \bibinfo{author}{Bertolino, A.}, \bibinfo{author}{Borgwardt, S.}, \bibinfo{author}{Brambilla, P.}, \bibinfo{author}{Kambeitz, J.}, et~al., \bibinfo{year}{2021}.
\newblock \bibinfo{title}{Heterogeneity and classification of recent onset psychosis and depression: a multimodal machine learning approach}.
\newblock \bibinfo{journal}{Schizophrenia bulletin} \bibinfo{volume}{47}, \bibinfo{pages}{1130--1140}.
\bibitem[{LaMontagne et~al.(2019)LaMontagne, Benzinger, Morris, Keefe, Hornbeck, Xiong, Grant, Hassenstab, Moulder, Vlassenko et~al.}]{lamontagne2019oasis}
\bibinfo{author}{LaMontagne, P.J.}, \bibinfo{author}{Benzinger, T.L.}, \bibinfo{author}{Morris, J.C.}, \bibinfo{author}{Keefe, S.}, \bibinfo{author}{Hornbeck, R.}, \bibinfo{author}{Xiong, C.}, \bibinfo{author}{Grant, E.}, \bibinfo{author}{Hassenstab, J.}, \bibinfo{author}{Moulder, K.}, \bibinfo{author}{Vlassenko, A.G.}, et~al., \bibinfo{year}{2019}.
\newblock \bibinfo{title}{Oasis-3: longitudinal neuroimaging, clinical, and cognitive dataset for normal aging and alzheimer disease}.
\newblock \bibinfo{journal}{MedRxiv} , \bibinfo{pages}{2019--12}.
\bibitem[{Le et~al.(2017)Le, Chen, Wang, Wang, Liu, Cheng and Yang}]{le2017automated}
\bibinfo{author}{Le, M.H.}, \bibinfo{author}{Chen, J.}, \bibinfo{author}{Wang, L.}, \bibinfo{author}{Wang, Z.}, \bibinfo{author}{Liu, W.}, \bibinfo{author}{Cheng, K.T.T.}, \bibinfo{author}{Yang, X.}, \bibinfo{year}{2017}.
\newblock \bibinfo{title}{Automated diagnosis of prostate cancer in multi-parametric mri based on multimodal convolutional neural networks}.
\newblock \bibinfo{journal}{Physics in Medicine \& Biology} \bibinfo{volume}{62}, \bibinfo{pages}{6497}.
\bibitem[{Lee et~al.(2019a)Lee, Nho, Kang, Sohn and Kim}]{lee2019predicting}
\bibinfo{author}{Lee, G.}, \bibinfo{author}{Nho, K.}, \bibinfo{author}{Kang, B.}, \bibinfo{author}{Sohn, K.A.}, \bibinfo{author}{Kim, D.}, \bibinfo{year}{2019}a.
\newblock \bibinfo{title}{Predicting alzheimer’s disease progression using multi-modal deep learning approach}.
\newblock \bibinfo{journal}{Scientific reports} \bibinfo{volume}{9}, \bibinfo{pages}{1952}.
\bibitem[{Lee et~al.(2019b)Lee, Mawla, Kim, Loggia, Ortiz, Jung, Chan, Gerber, Schmithorst, Edwards et~al.}]{lee2019machine}
\bibinfo{author}{Lee, J.}, \bibinfo{author}{Mawla, I.}, \bibinfo{author}{Kim, J.}, \bibinfo{author}{Loggia, M.L.}, \bibinfo{author}{Ortiz, A.}, \bibinfo{author}{Jung, C.}, \bibinfo{author}{Chan, S.T.}, \bibinfo{author}{Gerber, J.}, \bibinfo{author}{Schmithorst, V.J.}, \bibinfo{author}{Edwards, R.R.}, et~al., \bibinfo{year}{2019}b.
\newblock \bibinfo{title}{Machine learning-based prediction of clinical pain using multimodal neuroimaging and autonomic metrics}.
\newblock \bibinfo{journal}{pain} \bibinfo{volume}{160}, \bibinfo{pages}{550}.
\bibitem[{Leighton(2007)}]{leighton2007ultrasound}
\bibinfo{author}{Leighton, T.G.}, \bibinfo{year}{2007}.
\newblock \bibinfo{title}{What is ultrasound?}
\newblock \bibinfo{journal}{Progress in biophysics and molecular biology} \bibinfo{volume}{93}, \bibinfo{pages}{3--83}.
\bibitem[{Leng et~al.(2023)Leng, Cui, Peng, Yan, Cao, Yan, Chen, Jiang, Zheng, Initiative et~al.}]{leng2023multimodal}
\bibinfo{author}{Leng, Y.}, \bibinfo{author}{Cui, W.}, \bibinfo{author}{Peng, Y.}, \bibinfo{author}{Yan, C.}, \bibinfo{author}{Cao, Y.}, \bibinfo{author}{Yan, Z.}, \bibinfo{author}{Chen, S.}, \bibinfo{author}{Jiang, X.}, \bibinfo{author}{Zheng, J.}, \bibinfo{author}{Initiative, A.D.N.}, et~al., \bibinfo{year}{2023}.
\newblock \bibinfo{title}{Multimodal cross enhanced fusion network for diagnosis of alzheimer’s disease and subjective memory complaints}.
\newblock \bibinfo{journal}{Computers in Biology and Medicine} \bibinfo{volume}{157}, \bibinfo{pages}{106788}.
\bibitem[{Li et~al.(2015)Li, Tran, Thung, Ji, Shen and Li}]{li2015robust}
\bibinfo{author}{Li, F.}, \bibinfo{author}{Tran, L.}, \bibinfo{author}{Thung, K.H.}, \bibinfo{author}{Ji, S.}, \bibinfo{author}{Shen, D.}, \bibinfo{author}{Li, J.}, \bibinfo{year}{2015}.
\newblock \bibinfo{title}{A robust deep model for improved classification of ad/mci patients}.
\newblock \bibinfo{journal}{IEEE journal of biomedical and health informatics} \bibinfo{volume}{19}, \bibinfo{pages}{1610--1616}.
\bibitem[{Li et~al.(2022a)Li, Bu and Qian}]{li2022cross}
\bibinfo{author}{Li, J.}, \bibinfo{author}{Bu, C.}, \bibinfo{author}{Qian, C.}, \bibinfo{year}{2022}a.
\newblock \bibinfo{title}{A cross-attention based image fusion network for prediction of mild cognitive impairment}, in: \bibinfo{booktitle}{Journal of Physics: Conference Series}, \bibinfo{organization}{IOP Publishing}. p. \bibinfo{pages}{012002}.
\bibitem[{Li et~al.(2017)Li, Zhou, Zhao, Hao, Yao, Zhong and Zhi}]{li2017b}
\bibinfo{author}{Li, L.}, \bibinfo{author}{Zhou, X.}, \bibinfo{author}{Zhao, X.}, \bibinfo{author}{Hao, S.}, \bibinfo{author}{Yao, J.}, \bibinfo{author}{Zhong, W.}, \bibinfo{author}{Zhi, H.}, \bibinfo{year}{2017}.
\newblock \bibinfo{title}{B-mode ultrasound combined with color doppler and strain elastography in the diagnosis of non-mass breast lesions: A prospective study}.
\newblock \bibinfo{journal}{Ultrasound in medicine \& biology} \bibinfo{volume}{43}, \bibinfo{pages}{2582--2590}.
\bibitem[{Li et~al.(2021)Li, Yang, Ross and Kanazawa}]{li2021ai}
\bibinfo{author}{Li, R.}, \bibinfo{author}{Yang, S.}, \bibinfo{author}{Ross, D.A.}, \bibinfo{author}{Kanazawa, A.}, \bibinfo{year}{2021}.
\newblock \bibinfo{title}{Ai choreographer: Music conditioned 3d dance generation with aist++}, in: \bibinfo{booktitle}{Proceedings of the IEEE/CVF International Conference on Computer Vision}, pp. \bibinfo{pages}{13401--13412}.
\bibitem[{Li et~al.(2022b)Li, Xie, Wang, Zhang and Zhou}]{li2022attention}
\bibinfo{author}{Li, S.}, \bibinfo{author}{Xie, Y.}, \bibinfo{author}{Wang, G.}, \bibinfo{author}{Zhang, L.}, \bibinfo{author}{Zhou, W.}, \bibinfo{year}{2022}b.
\newblock \bibinfo{title}{Attention guided discriminative feature learning and adaptive fusion for grading hepatocellular carcinoma with contrast-enhanced mr}.
\newblock \bibinfo{journal}{Computerized Medical Imaging and Graphics} \bibinfo{volume}{97}, \bibinfo{pages}{102050}.
\bibitem[{Li et~al.(2022c)Li, El~Habib~Daho, Conze, Al~Hajj, Bonnin, Ren, Manivannan, Magazzeni, Tadayoni, Cochener et~al.}]{li2022multimodal}
\bibinfo{author}{Li, Y.}, \bibinfo{author}{El~Habib~Daho, M.}, \bibinfo{author}{Conze, P.H.}, \bibinfo{author}{Al~Hajj, H.}, \bibinfo{author}{Bonnin, S.}, \bibinfo{author}{Ren, H.}, \bibinfo{author}{Manivannan, N.}, \bibinfo{author}{Magazzeni, S.}, \bibinfo{author}{Tadayoni, R.}, \bibinfo{author}{Cochener, B.}, et~al., \bibinfo{year}{2022}c.
\newblock \bibinfo{title}{Multimodal information fusion for glaucoma and diabetic retinopathy classification}, in: \bibinfo{booktitle}{Ophthalmic Medical Image Analysis: 9th International Workshop, OMIA 2022, Held in Conjunction with MICCAI 2022, Singapore, Singapore, September 22, 2022, Proceedings}, \bibinfo{organization}{Springer}. pp. \bibinfo{pages}{53--62}.
\bibitem[{Li et~al.(2023)Li, El~Habib~Daho, Conze, Zeghlache, Le~Boité, Bonnin, Cosette, Magazzeni, Lay, Le~Guilcher, Tadayoni, Cochener, Lamard and Quellec}]{li2023diagnostics}
\bibinfo{author}{Li, Y.}, \bibinfo{author}{El~Habib~Daho, M.}, \bibinfo{author}{Conze, P.H.}, \bibinfo{author}{Zeghlache, R.}, \bibinfo{author}{Le~Boité, H.}, \bibinfo{author}{Bonnin, S.}, \bibinfo{author}{Cosette, D.}, \bibinfo{author}{Magazzeni, S.}, \bibinfo{author}{Lay, B.}, \bibinfo{author}{Le~Guilcher, A.}, \bibinfo{author}{Tadayoni, R.}, \bibinfo{author}{Cochener, B.}, \bibinfo{author}{Lamard, M.}, \bibinfo{author}{Quellec, G.}, \bibinfo{year}{2023}.
\newblock \bibinfo{title}{Hybrid fusion of high-resolution and ultra-widefield octa acquisitions for the automatic diagnosis of diabetic retinopathy}.
\newblock \bibinfo{journal}{Diagnostics} \bibinfo{volume}{13}.
\newblock \URLprefix \url{https://www.mdpi.com/2075-4418/13/17/2770}, \DOIprefix\doi{10.3390/diagnostics13172770}.
\bibitem[{Li et~al.(2022d)Li, Yu, Meng, Caine, Ngiam, Peng, Shen, Lu, Zhou, Le et~al.}]{li2022deepfusion}
\bibinfo{author}{Li, Y.}, \bibinfo{author}{Yu, A.W.}, \bibinfo{author}{Meng, T.}, \bibinfo{author}{Caine, B.}, \bibinfo{author}{Ngiam, J.}, \bibinfo{author}{Peng, D.}, \bibinfo{author}{Shen, J.}, \bibinfo{author}{Lu, Y.}, \bibinfo{author}{Zhou, D.}, \bibinfo{author}{Le, Q.V.}, et~al., \bibinfo{year}{2022}d.
\newblock \bibinfo{title}{Deepfusion: Lidar-camera deep fusion for multi-modal 3d object detection}, in: \bibinfo{booktitle}{Proceedings of the IEEE/CVF Conference on Computer Vision and Pattern Recognition}, pp. \bibinfo{pages}{17182--17191}.
\bibitem[{Lin et~al.(2021)Lin, Lin, Chen, Zhang, Gao, Huang, Tong, Du and ADNI}]{lin2021bidirectional}
\bibinfo{author}{Lin, W.}, \bibinfo{author}{Lin, W.}, \bibinfo{author}{Chen, G.}, \bibinfo{author}{Zhang, H.}, \bibinfo{author}{Gao, Q.}, \bibinfo{author}{Huang, Y.}, \bibinfo{author}{Tong, T.}, \bibinfo{author}{Du, M.}, \bibinfo{author}{ADNI}, \bibinfo{year}{2021}.
\newblock \bibinfo{title}{Bidirectional mapping of brain mri and pet with 3d reversible gan for the diagnosis of alzheimer’s disease}.
\newblock \bibinfo{journal}{Frontiers in Neuroscience} \bibinfo{volume}{15}, \bibinfo{pages}{646013}.
\bibitem[{Lindig et~al.(2018)Lindig, Kotikalapudi, Schweikardt, Martin, Bender, Klose, Ernemann, Focke and Bender}]{lindig2018evaluation}
\bibinfo{author}{Lindig, T.}, \bibinfo{author}{Kotikalapudi, R.}, \bibinfo{author}{Schweikardt, D.}, \bibinfo{author}{Martin, P.}, \bibinfo{author}{Bender, F.}, \bibinfo{author}{Klose, U.}, \bibinfo{author}{Ernemann, U.}, \bibinfo{author}{Focke, N.K.}, \bibinfo{author}{Bender, B.}, \bibinfo{year}{2018}.
\newblock \bibinfo{title}{Evaluation of multimodal segmentation based on 3d t1-, t2-and flair-weighted images--the difficulty of choosing}.
\newblock \bibinfo{journal}{Neuroimage} \bibinfo{volume}{170}, \bibinfo{pages}{210--221}.
\bibitem[{Lipkova et~al.(2022)Lipkova, Chen, Chen, Lu, Barbieri, Shao, Vaidya, Chen, Zhuang, Williamson et~al.}]{lipkova2022artificial}
\bibinfo{author}{Lipkova, J.}, \bibinfo{author}{Chen, R.J.}, \bibinfo{author}{Chen, B.}, \bibinfo{author}{Lu, M.Y.}, \bibinfo{author}{Barbieri, M.}, \bibinfo{author}{Shao, D.}, \bibinfo{author}{Vaidya, A.J.}, \bibinfo{author}{Chen, C.}, \bibinfo{author}{Zhuang, L.}, \bibinfo{author}{Williamson, D.F.}, et~al., \bibinfo{year}{2022}.
\newblock \bibinfo{title}{Artificial intelligence for multimodal data integration in oncology}.
\newblock \bibinfo{journal}{Cancer Cell} \bibinfo{volume}{40}, \bibinfo{pages}{1095--1110}.
\bibitem[{Liu et~al.(2023a)Liu, Liu, Zhang, To, Nasrallah and Chandra}]{liu2023cascaded}
\bibinfo{author}{Liu, L.}, \bibinfo{author}{Liu, S.}, \bibinfo{author}{Zhang, L.}, \bibinfo{author}{To, X.V.}, \bibinfo{author}{Nasrallah, F.}, \bibinfo{author}{Chandra, S.S.}, \bibinfo{year}{2023}a.
\newblock \bibinfo{title}{Cascaded multi-modal mixing transformers for alzheimer’s disease classification with incomplete data}.
\newblock \bibinfo{journal}{NeuroImage} \bibinfo{volume}{277}, \bibinfo{pages}{120267}.
\bibitem[{Liu et~al.(2018)Liu, Cheng, Wang, Wang and ADNI}]{liu2018multi}
\bibinfo{author}{Liu, M.}, \bibinfo{author}{Cheng, D.}, \bibinfo{author}{Wang, K.}, \bibinfo{author}{Wang, Y.}, \bibinfo{author}{ADNI}, \bibinfo{year}{2018}.
\newblock \bibinfo{title}{Multi-modality cascaded convolutional neural networks for alzheimer’s disease diagnosis}.
\newblock \bibinfo{journal}{Neuroinformatics} \bibinfo{volume}{16}, \bibinfo{pages}{295--308}.
\bibitem[{Liu et~al.(2017)Liu, Gao, Yap and Shen}]{liu2017multi}
\bibinfo{author}{Liu, M.}, \bibinfo{author}{Gao, Y.}, \bibinfo{author}{Yap, P.T.}, \bibinfo{author}{Shen, D.}, \bibinfo{year}{2017}.
\newblock \bibinfo{title}{Multi-hypergraph learning for incomplete multimodality data}.
\newblock \bibinfo{journal}{IEEE journal of biomedical and health informatics} \bibinfo{volume}{22}, \bibinfo{pages}{1197--1208}.
\bibitem[{Liu et~al.(2014a)Liu, Zhang, Shen and ADNI}]{liu2014hierarchical}
\bibinfo{author}{Liu, M.}, \bibinfo{author}{Zhang, D.}, \bibinfo{author}{Shen, D.}, \bibinfo{author}{ADNI}, \bibinfo{year}{2014}a.
\newblock \bibinfo{title}{Hierarchical fusion of features and classifier decisions for alzheimer's disease diagnosis}.
\newblock \bibinfo{journal}{Human brain mapping} \bibinfo{volume}{35}, \bibinfo{pages}{1305--1319}.
\bibitem[{Liu et~al.(2012)Liu, Zhang, Shen, ADNI et~al.}]{liu2012ensemble}
\bibinfo{author}{Liu, M.}, \bibinfo{author}{Zhang, D.}, \bibinfo{author}{Shen, D.}, \bibinfo{author}{ADNI}, et~al., \bibinfo{year}{2012}.
\newblock \bibinfo{title}{Ensemble sparse classification of alzheimer's disease}.
\newblock \bibinfo{journal}{NeuroImage} \bibinfo{volume}{60}, \bibinfo{pages}{1106--1116}.
\bibitem[{Liu et~al.(2022)Liu, Huang, Hu, Zhu, Wong and Tan}]{liu2022attention}
\bibinfo{author}{Liu, R.}, \bibinfo{author}{Huang, Z.A.}, \bibinfo{author}{Hu, Y.}, \bibinfo{author}{Zhu, Z.}, \bibinfo{author}{Wong, K.C.}, \bibinfo{author}{Tan, K.C.}, \bibinfo{year}{2022}.
\newblock \bibinfo{title}{Attention-like multimodality fusion with data augmentation for diagnosis of mental disorders using mri}.
\newblock \bibinfo{journal}{IEEE Transactions on Neural Networks and Learning Systems} .
\bibitem[{Liu et~al.(2014b)Liu, Liu, Cai, Che, Pujol, Kikinis, Feng, Fulham et~al.}]{liu2014multimodal}
\bibinfo{author}{Liu, S.}, \bibinfo{author}{Liu, S.}, \bibinfo{author}{Cai, W.}, \bibinfo{author}{Che, H.}, \bibinfo{author}{Pujol, S.}, \bibinfo{author}{Kikinis, R.}, \bibinfo{author}{Feng, D.}, \bibinfo{author}{Fulham, M.J.}, et~al., \bibinfo{year}{2014}b.
\newblock \bibinfo{title}{Multimodal neuroimaging feature learning for multiclass diagnosis of alzheimer's disease}.
\newblock \bibinfo{journal}{IEEE transactions on biomedical engineering} \bibinfo{volume}{62}, \bibinfo{pages}{1132--1140}.
\bibitem[{Liu et~al.(2023b)Liu, Zheng, Li, Pan, Fang, Liu, Qiao, Pan, Jia and Ge}]{liu2023improving}
\bibinfo{author}{Liu, S.}, \bibinfo{author}{Zheng, Y.}, \bibinfo{author}{Li, H.}, \bibinfo{author}{Pan, M.}, \bibinfo{author}{Fang, Z.}, \bibinfo{author}{Liu, M.}, \bibinfo{author}{Qiao, Y.}, \bibinfo{author}{Pan, N.}, \bibinfo{author}{Jia, W.}, \bibinfo{author}{Ge, X.}, \bibinfo{year}{2023}b.
\newblock \bibinfo{title}{Improving alzheimer diagnoses with an interpretable deep learning framework: Including neuropsychiatric symptoms}.
\newblock \bibinfo{journal}{Neuroscience} \bibinfo{volume}{531}, \bibinfo{pages}{86--98}.
\bibitem[{Liu et~al.(2015)Liu, Liu and Wang}]{liu2015general}
\bibinfo{author}{Liu, Y.}, \bibinfo{author}{Liu, S.}, \bibinfo{author}{Wang, Z.}, \bibinfo{year}{2015}.
\newblock \bibinfo{title}{A general framework for image fusion based on multi-scale transform and sparse representation}.
\newblock \bibinfo{journal}{Information fusion} \bibinfo{volume}{24}, \bibinfo{pages}{147--164}.
\bibitem[{Liu et~al.(2010)Liu, Yang and Sun}]{liu2010pet}
\bibinfo{author}{Liu, Y.}, \bibinfo{author}{Yang, J.}, \bibinfo{author}{Sun, J.}, \bibinfo{year}{2010}.
\newblock \bibinfo{title}{Pet/ct medical image fusion algorithm based on multiwavelet transform}, in: \bibinfo{booktitle}{2010 2nd International Conference on Advanced Computer Control}, \bibinfo{organization}{IEEE}. pp. \bibinfo{pages}{264--268}.
\bibitem[{Lu et~al.(2018)Lu, Popuri, Ding, Balachandar and Beg}]{lu2018multimodal}
\bibinfo{author}{Lu, D.}, \bibinfo{author}{Popuri, K.}, \bibinfo{author}{Ding, G.W.}, \bibinfo{author}{Balachandar, R.}, \bibinfo{author}{Beg, M.F.}, \bibinfo{year}{2018}.
\newblock \bibinfo{title}{Multimodal and multiscale deep neural networks for the early diagnosis of alzheimer’s disease using structural mr and fdg-pet images}.
\newblock \bibinfo{journal}{Scientific reports} \bibinfo{volume}{8}, \bibinfo{pages}{5697}.
\bibitem[{Lu et~al.(2019)Lu, Batra, Parikh and Lee}]{lu2019vilbert}
\bibinfo{author}{Lu, J.}, \bibinfo{author}{Batra, D.}, \bibinfo{author}{Parikh, D.}, \bibinfo{author}{Lee, S.}, \bibinfo{year}{2019}.
\newblock \bibinfo{title}{Vilbert: Pretraining task-agnostic visiolinguistic representations for vision-and-language tasks}.
\newblock \bibinfo{journal}{Advances in neural information processing systems} \bibinfo{volume}{32}.
\bibitem[{Lu et~al.(2010)Lu, Pu, Gao, Zhou, Qiu and Si-Tu}]{lu2010comparative}
\bibinfo{author}{Lu, M.H.}, \bibinfo{author}{Pu, X.Y.}, \bibinfo{author}{Gao, X.}, \bibinfo{author}{Zhou, X.F.}, \bibinfo{author}{Qiu, J.G.}, \bibinfo{author}{Si-Tu, J.}, \bibinfo{year}{2010}.
\newblock \bibinfo{title}{A comparative study of clinical value of single b-mode ultrasound guidance and b-mode combined with color doppler ultrasound guidance in mini-invasive percutaneous nephrolithotomy to decrease hemorrhagic complications}.
\newblock \bibinfo{journal}{Urology} \bibinfo{volume}{76}, \bibinfo{pages}{815--820}.
\bibitem[{Lu et~al.(2024)Lu, Hu, Mitelpunkt, Bhatnagar, Lu and Liang}]{lu2024hierarchical}
\bibinfo{author}{Lu, P.}, \bibinfo{author}{Hu, L.}, \bibinfo{author}{Mitelpunkt, A.}, \bibinfo{author}{Bhatnagar, S.}, \bibinfo{author}{Lu, L.}, \bibinfo{author}{Liang, H.}, \bibinfo{year}{2024}.
\newblock \bibinfo{title}{A hierarchical attention-based multimodal fusion framework for predicting the progression of alzheimer’s disease}.
\newblock \bibinfo{journal}{Biomedical Signal Processing and Control} \bibinfo{volume}{88}, \bibinfo{pages}{105669}.
\bibitem[{MacKie et~al.(2002)MacKie, Fleming, McMahon and Jarrett}]{mackie2002use}
\bibinfo{author}{MacKie, R.}, \bibinfo{author}{Fleming, C.}, \bibinfo{author}{McMahon, A.}, \bibinfo{author}{Jarrett, P.}, \bibinfo{year}{2002}.
\newblock \bibinfo{title}{The use of the dermatoscope to identify early melanoma using the three-colour test}.
\newblock \bibinfo{journal}{British Journal of Dermatology} \bibinfo{volume}{146}, \bibinfo{pages}{481--484}.
\bibitem[{Mahmood et~al.(2018)Mahmood, Yang, Ashley and Durr}]{mahmood2018multimodal}
\bibinfo{author}{Mahmood, F.}, \bibinfo{author}{Yang, Z.}, \bibinfo{author}{Ashley, T.}, \bibinfo{author}{Durr, N.J.}, \bibinfo{year}{2018}.
\newblock \bibinfo{title}{Multimodal densenet}.
\newblock \bibinfo{journal}{arXiv preprint arXiv:1811.07407} .
\bibitem[{Marcus et~al.(2010)Marcus, Fotenos, Csernansky, Morris and Buckner}]{marcus2010open}
\bibinfo{author}{Marcus, D.S.}, \bibinfo{author}{Fotenos, A.F.}, \bibinfo{author}{Csernansky, J.G.}, \bibinfo{author}{Morris, J.C.}, \bibinfo{author}{Buckner, R.L.}, \bibinfo{year}{2010}.
\newblock \bibinfo{title}{Open access series of imaging studies: longitudinal mri data in nondemented and demented older adults}.
\newblock \bibinfo{journal}{Journal of cognitive neuroscience} \bibinfo{volume}{22}, \bibinfo{pages}{2677--2684}.
\bibitem[{Massalimova and Varol(2021)}]{massalimova2021input}
\bibinfo{author}{Massalimova, A.}, \bibinfo{author}{Varol, H.A.}, \bibinfo{year}{2021}.
\newblock \bibinfo{title}{Input agnostic deep learning for alzheimer’s disease classification using multimodal mri images}, in: \bibinfo{booktitle}{2021 43rd Annual International Conference of the IEEE Engineering in Medicine \& Biology Society (EMBC)}, \bibinfo{organization}{IEEE}. pp. \bibinfo{pages}{2875--2878}.
\bibitem[{Mehrtash et~al.(2017)Mehrtash, Sedghi, Ghafoorian, Taghipour, Tempany, Wells~III, Kapur, Mousavi, Abolmaesumi and Fedorov}]{mehrtash2017classification}
\bibinfo{author}{Mehrtash, A.}, \bibinfo{author}{Sedghi, A.}, \bibinfo{author}{Ghafoorian, M.}, \bibinfo{author}{Taghipour, M.}, \bibinfo{author}{Tempany, C.M.}, \bibinfo{author}{Wells~III, W.M.}, \bibinfo{author}{Kapur, T.}, \bibinfo{author}{Mousavi, P.}, \bibinfo{author}{Abolmaesumi, P.}, \bibinfo{author}{Fedorov, A.}, \bibinfo{year}{2017}.
\newblock \bibinfo{title}{Classification of clinical significance of mri prostate findings using 3d convolutional neural networks}, in: \bibinfo{booktitle}{Medical imaging 2017: computer-aided diagnosis}, \bibinfo{organization}{SPIE}. pp. \bibinfo{pages}{589--592}.
\bibitem[{Menze et~al.(2014)Menze, Jakab, Bauer, Kalpathy-Cramer, Farahani, Kirby, Burren, Porz, Slotboom, Wiest et~al.}]{menze2014multimodal}
\bibinfo{author}{Menze, B.H.}, \bibinfo{author}{Jakab, A.}, \bibinfo{author}{Bauer, S.}, \bibinfo{author}{Kalpathy-Cramer, J.}, \bibinfo{author}{Farahani, K.}, \bibinfo{author}{Kirby, J.}, \bibinfo{author}{Burren, Y.}, \bibinfo{author}{Porz, N.}, \bibinfo{author}{Slotboom, J.}, \bibinfo{author}{Wiest, R.}, et~al., \bibinfo{year}{2014}.
\newblock \bibinfo{title}{The multimodal brain tumor image segmentation benchmark (brats)}.
\newblock \bibinfo{journal}{IEEE transactions on medical imaging} \bibinfo{volume}{34}, \bibinfo{pages}{1993--2024}.
\bibitem[{Mesejo et~al.(2016)Mesejo, Pizarro, Abergel, Rouquette, Beorchia, Poincloux and Bartoli}]{mesejo2016computer}
\bibinfo{author}{Mesejo, P.}, \bibinfo{author}{Pizarro, D.}, \bibinfo{author}{Abergel, A.}, \bibinfo{author}{Rouquette, O.}, \bibinfo{author}{Beorchia, S.}, \bibinfo{author}{Poincloux, L.}, \bibinfo{author}{Bartoli, A.}, \bibinfo{year}{2016}.
\newblock \bibinfo{title}{Computer-aided classification of gastrointestinal lesions in regular colonoscopy}.
\newblock \bibinfo{journal}{IEEE transactions on medical imaging} \bibinfo{volume}{35}, \bibinfo{pages}{2051--2063}.
\bibitem[{Miao et~al.(2024)Miao, Xu, Li, Yang, Sheng, Liu, Bezabih and Yu}]{miao2024mmtfn}
\bibinfo{author}{Miao, S.}, \bibinfo{author}{Xu, Q.}, \bibinfo{author}{Li, W.}, \bibinfo{author}{Yang, C.}, \bibinfo{author}{Sheng, B.}, \bibinfo{author}{Liu, F.}, \bibinfo{author}{Bezabih, T.T.}, \bibinfo{author}{Yu, X.}, \bibinfo{year}{2024}.
\newblock \bibinfo{title}{Mmtfn: Multi-modal multi-scale transformer fusion network for alzheimer's disease diagnosis}.
\newblock \bibinfo{journal}{International Journal of Imaging Systems and Technology} \bibinfo{volume}{34}, \bibinfo{pages}{e22970}.
\bibitem[{Mishra and Palkar(2015)}]{mishra2015image}
\bibinfo{author}{Mishra, D.}, \bibinfo{author}{Palkar, B.}, \bibinfo{year}{2015}.
\newblock \bibinfo{title}{Image fusion techniques: a review}.
\newblock \bibinfo{journal}{International Journal of Computer Applications} \bibinfo{volume}{130}, \bibinfo{pages}{7--13}.
\bibitem[{{Mohit Prabhushankar} et~al.(2022){Mohit Prabhushankar}, Kokilepersaud, {Yash-Yee Logan}, Corona, AlRegib and Wykoff}]{MohitPrabhushankar2022}
\bibinfo{author}{{Mohit Prabhushankar}}, \bibinfo{author}{Kokilepersaud, K.}, \bibinfo{author}{{Yash-Yee Logan}}, \bibinfo{author}{Corona, S.T.}, \bibinfo{author}{AlRegib, G.}, \bibinfo{author}{Wykoff, C.}, \bibinfo{year}{2022}.
\newblock \bibinfo{title}{Olives dataset: Ophthalmic labels for investigating visual eye semantics}.
\newblock \URLprefix \url{https://zenodo.org/record/7105232}, \DOIprefix\doi{10.5281/ZENODO.7105232}.
\bibitem[{Moon et~al.(2020)Moon, Lee, Ke, Lee, Huang and Chang}]{moon2020computer}
\bibinfo{author}{Moon, W.K.}, \bibinfo{author}{Lee, Y.W.}, \bibinfo{author}{Ke, H.H.}, \bibinfo{author}{Lee, S.H.}, \bibinfo{author}{Huang, C.S.}, \bibinfo{author}{Chang, R.F.}, \bibinfo{year}{2020}.
\newblock \bibinfo{title}{Computer-aided diagnosis of breast ultrasound images using ensemble learning from convolutional neural networks}.
\newblock \bibinfo{journal}{Computer methods and programs in biomedicine} \bibinfo{volume}{190}, \bibinfo{pages}{105361}.
\bibitem[{Muehllehner and Karp(2006)}]{muehllehner2006positron}
\bibinfo{author}{Muehllehner, G.}, \bibinfo{author}{Karp, J.S.}, \bibinfo{year}{2006}.
\newblock \bibinfo{title}{Positron emission tomography}.
\newblock \bibinfo{journal}{Physics in Medicine \& Biology} \bibinfo{volume}{51}, \bibinfo{pages}{R117}.
\bibitem[{Muhammad et~al.(2021)Muhammad, Alshehri, Karray, El~Saddik, Alsulaiman and Falk}]{muhammad2021comprehensive}
\bibinfo{author}{Muhammad, G.}, \bibinfo{author}{Alshehri, F.}, \bibinfo{author}{Karray, F.}, \bibinfo{author}{El~Saddik, A.}, \bibinfo{author}{Alsulaiman, M.}, \bibinfo{author}{Falk, T.H.}, \bibinfo{year}{2021}.
\newblock \bibinfo{title}{A comprehensive survey on multimodal medical signals fusion for smart healthcare systems}.
\newblock \bibinfo{journal}{Information Fusion} \bibinfo{volume}{76}, \bibinfo{pages}{355--375}.
\bibitem[{Mustafa and Luo(2023)}]{mustafa2023diagnosing}
\bibinfo{author}{Mustafa, Y.}, \bibinfo{author}{Luo, T.}, \bibinfo{year}{2023}.
\newblock \bibinfo{title}{Diagnosing alzheimer's disease using early-late multimodal data fusion with jacobian maps}.
\newblock \bibinfo{journal}{arXiv preprint arXiv:2310.16936} .
\bibitem[{Nagrani et~al.(2021)Nagrani, Yang, Arnab, Jansen, Schmid and Sun}]{nagrani2021attention}
\bibinfo{author}{Nagrani, A.}, \bibinfo{author}{Yang, S.}, \bibinfo{author}{Arnab, A.}, \bibinfo{author}{Jansen, A.}, \bibinfo{author}{Schmid, C.}, \bibinfo{author}{Sun, C.}, \bibinfo{year}{2021}.
\newblock \bibinfo{title}{Attention bottlenecks for multimodal fusion}.
\newblock \bibinfo{journal}{Advances in Neural Information Processing Systems} \bibinfo{volume}{34}, \bibinfo{pages}{14200--14213}.
\bibitem[{Narazani et~al.(2022)Narazani, Sarasua, P{\"o}lsterl, Lizarraga, Yakushev and Wachinger}]{narazani2022pet}
\bibinfo{author}{Narazani, M.}, \bibinfo{author}{Sarasua, I.}, \bibinfo{author}{P{\"o}lsterl, S.}, \bibinfo{author}{Lizarraga, A.}, \bibinfo{author}{Yakushev, I.}, \bibinfo{author}{Wachinger, C.}, \bibinfo{year}{2022}.
\newblock \bibinfo{title}{Is a pet all you need? a multi-modal study for alzheimer’s disease using 3d cnns}, in: \bibinfo{booktitle}{Medical Image Computing and Computer Assisted Intervention--MICCAI 2022: 25th International Conference, Singapore, September 18--22, 2022, Proceedings, Part I}, \bibinfo{organization}{Springer}. pp. \bibinfo{pages}{66--76}.
\bibitem[{Nevitt et~al.(2006)Nevitt, Felson and Lester}]{nevitt2006osteoarthritis}
\bibinfo{author}{Nevitt, M.}, \bibinfo{author}{Felson, D.}, \bibinfo{author}{Lester, G.}, \bibinfo{year}{2006}.
\newblock \bibinfo{title}{The osteoarthritis initiative}.
\newblock \bibinfo{journal}{Protocol for the cohort study} \bibinfo{volume}{1}.
\bibitem[{Odusami et~al.(2023)Odusami, Maskeli{\=u}nas, Dama{\v{s}}evi{\v{c}}ius and Misra}]{odusami2023explainable}
\bibinfo{author}{Odusami, M.}, \bibinfo{author}{Maskeli{\=u}nas, R.}, \bibinfo{author}{Dama{\v{s}}evi{\v{c}}ius, R.}, \bibinfo{author}{Misra, S.}, \bibinfo{year}{2023}.
\newblock \bibinfo{title}{Explainable deep-learning-based diagnosis of alzheimer’s disease using multimodal input fusion of pet and mri images}.
\newblock \bibinfo{journal}{Journal of Medical and Biological Engineering} , \bibinfo{pages}{1--12}.
\bibitem[{Omeroglu et~al.(2023)Omeroglu, Mohammed, Oral and Aydin}]{omeroglu2023novel}
\bibinfo{author}{Omeroglu, A.N.}, \bibinfo{author}{Mohammed, H.M.}, \bibinfo{author}{Oral, E.A.}, \bibinfo{author}{Aydin, S.}, \bibinfo{year}{2023}.
\newblock \bibinfo{title}{A novel soft attention-based multi-modal deep learning framework for multi-label skin lesion classification}.
\newblock \bibinfo{journal}{Engineering Applications of Artificial Intelligence} \bibinfo{volume}{120}, \bibinfo{pages}{105897}.
\bibitem[{Pai et~al.(2020)Pai, Mandal, Punjabi, Shukla, Goel, Joon, Roy, Sandal, Mishra and Lahoti}]{pai2020brahma}
\bibinfo{author}{Pai, P.P.}, \bibinfo{author}{Mandal, P.K.}, \bibinfo{author}{Punjabi, K.}, \bibinfo{author}{Shukla, D.}, \bibinfo{author}{Goel, A.}, \bibinfo{author}{Joon, S.}, \bibinfo{author}{Roy, S.}, \bibinfo{author}{Sandal, K.}, \bibinfo{author}{Mishra, R.}, \bibinfo{author}{Lahoti, R.}, \bibinfo{year}{2020}.
\newblock \bibinfo{title}{Brahma: Population specific t1, t2, and flair weighted brain templates and their impact in structural and functional imaging studies}.
\newblock \bibinfo{journal}{Magnetic resonance imaging} \bibinfo{volume}{70}, \bibinfo{pages}{5--21}.
\bibitem[{Pan et~al.(2020)Pan, Liu, Lian, Xia and Shen}]{pan2020spatially}
\bibinfo{author}{Pan, Y.}, \bibinfo{author}{Liu, M.}, \bibinfo{author}{Lian, C.}, \bibinfo{author}{Xia, Y.}, \bibinfo{author}{Shen, D.}, \bibinfo{year}{2020}.
\newblock \bibinfo{title}{Spatially-constrained fisher representation for brain disease identification with incomplete multi-modal neuroimages}.
\newblock \bibinfo{journal}{IEEE transactions on medical imaging} \bibinfo{volume}{39}, \bibinfo{pages}{2965--2975}.
\bibitem[{Pan et~al.(2018)Pan, Liu, Lian, Zhou, Xia and Shen}]{pan2018synthesizing}
\bibinfo{author}{Pan, Y.}, \bibinfo{author}{Liu, M.}, \bibinfo{author}{Lian, C.}, \bibinfo{author}{Zhou, T.}, \bibinfo{author}{Xia, Y.}, \bibinfo{author}{Shen, D.}, \bibinfo{year}{2018}.
\newblock \bibinfo{title}{Synthesizing missing pet from mri with cycle-consistent generative adversarial networks for alzheimer’s disease diagnosis}, in: \bibinfo{booktitle}{Medical Image Computing and Computer Assisted Intervention--MICCAI 2018: 21st International Conference, Granada, Spain, September 16-20, 2018, Proceedings, Part III 11}, \bibinfo{organization}{Springer}. pp. \bibinfo{pages}{455--463}.
\bibitem[{Pan et~al.(2021)Pan, Liu, Xia and Shen}]{pan2021disease}
\bibinfo{author}{Pan, Y.}, \bibinfo{author}{Liu, M.}, \bibinfo{author}{Xia, Y.}, \bibinfo{author}{Shen, D.}, \bibinfo{year}{2021}.
\newblock \bibinfo{title}{Disease-image-specific learning for diagnosis-oriented neuroimage synthesis with incomplete multi-modality data}.
\newblock \bibinfo{journal}{IEEE transactions on pattern analysis and machine intelligence} \bibinfo{volume}{44}, \bibinfo{pages}{6839--6853}.
\bibitem[{Parmar and Kher(2012)}]{parmar2012comparative}
\bibinfo{author}{Parmar, K.}, \bibinfo{author}{Kher, R.}, \bibinfo{year}{2012}.
\newblock \bibinfo{title}{A comparative analysis of multimodality medical image fusion methods}, in: \bibinfo{booktitle}{2012 Sixth Asia Modelling Symposium}, \bibinfo{organization}{IEEE}. pp. \bibinfo{pages}{93--97}.
\bibitem[{Pashevich et~al.(2021)Pashevich, Schmid and Sun}]{pashevich2021episodic}
\bibinfo{author}{Pashevich, A.}, \bibinfo{author}{Schmid, C.}, \bibinfo{author}{Sun, C.}, \bibinfo{year}{2021}.
\newblock \bibinfo{title}{Episodic transformer for vision-and-language navigation}, in: \bibinfo{booktitle}{Proceedings of the IEEE/CVF International Conference on Computer Vision}, pp. \bibinfo{pages}{15942--15952}.
\bibitem[{Pereira et~al.(2016)Pereira, Pinto, Alves and Silva}]{pereira2016brain}
\bibinfo{author}{Pereira, S.}, \bibinfo{author}{Pinto, A.}, \bibinfo{author}{Alves, V.}, \bibinfo{author}{Silva, C.A.}, \bibinfo{year}{2016}.
\newblock \bibinfo{title}{Brain tumor segmentation using convolutional neural networks in mri images}.
\newblock \bibinfo{journal}{IEEE transactions on medical imaging} \bibinfo{volume}{35}, \bibinfo{pages}{1240--1251}.
\bibitem[{P{\'e}rez-R{\'u}a et~al.(2019)P{\'e}rez-R{\'u}a, Vielzeuf, Pateux, Baccouche and Jurie}]{perez2019mfas}
\bibinfo{author}{P{\'e}rez-R{\'u}a, J.M.}, \bibinfo{author}{Vielzeuf, V.}, \bibinfo{author}{Pateux, S.}, \bibinfo{author}{Baccouche, M.}, \bibinfo{author}{Jurie, F.}, \bibinfo{year}{2019}.
\newblock \bibinfo{title}{Mfas: Multimodal fusion architecture search}, in: \bibinfo{booktitle}{Proceedings of the IEEE/CVF Conference on Computer Vision and Pattern Recognition}, pp. \bibinfo{pages}{6966--6975}.
\bibitem[{Petersen et~al.(2010)Petersen, Aisen, Beckett, Donohue, Gamst, Harvey, Jack, Jagust, Shaw, Toga et~al.}]{petersen2010alzheimer}
\bibinfo{author}{Petersen, R.C.}, \bibinfo{author}{Aisen, P.S.}, \bibinfo{author}{Beckett, L.A.}, \bibinfo{author}{Donohue, M.C.}, \bibinfo{author}{Gamst, A.C.}, \bibinfo{author}{Harvey, D.J.}, \bibinfo{author}{Jack, C.R.}, \bibinfo{author}{Jagust, W.J.}, \bibinfo{author}{Shaw, L.M.}, \bibinfo{author}{Toga, A.W.}, et~al., \bibinfo{year}{2010}.
\newblock \bibinfo{title}{Alzheimer's disease neuroimaging initiative (adni): clinical characterization}.
\newblock \bibinfo{journal}{Neurology} \bibinfo{volume}{74}, \bibinfo{pages}{201--209}.
\bibitem[{Plewes and Kucharczyk(2012)}]{plewes2012physics}
\bibinfo{author}{Plewes, D.B.}, \bibinfo{author}{Kucharczyk, W.}, \bibinfo{year}{2012}.
\newblock \bibinfo{title}{Physics of mri: a primer}.
\newblock \bibinfo{journal}{Journal of magnetic resonance imaging} \bibinfo{volume}{35}, \bibinfo{pages}{1038--1054}.
\bibitem[{Prabhu et~al.(2022)Prabhu, Berkebile, Rajagopalan, Yao, Shi, Giuste, Zhong, Sun and Wang}]{prabhu2022multi}
\bibinfo{author}{Prabhu, S.S.}, \bibinfo{author}{Berkebile, J.A.}, \bibinfo{author}{Rajagopalan, N.}, \bibinfo{author}{Yao, R.}, \bibinfo{author}{Shi, W.}, \bibinfo{author}{Giuste, F.}, \bibinfo{author}{Zhong, Y.}, \bibinfo{author}{Sun, J.}, \bibinfo{author}{Wang, M.D.}, \bibinfo{year}{2022}.
\newblock \bibinfo{title}{Multi-modal deep learning models for alzheimer's disease prediction using mri and ehr}, in: \bibinfo{booktitle}{2022 IEEE 22nd International Conference on Bioinformatics and Bioengineering (BIBE)}, \bibinfo{organization}{IEEE}. pp. \bibinfo{pages}{168--173}.
\bibitem[{Preston(2006)}]{preston2006magnetic}
\bibinfo{author}{Preston, D.C.}, \bibinfo{year}{2006}.
\newblock \bibinfo{title}{Magnetic resonance imaging (mri) of the brain and spine: Basics}.
\newblock \bibinfo{journal}{MRI Basics, Case Med} \bibinfo{volume}{30}.
\bibitem[{Princess et~al.(2014)Princess, Kumar and Begum}]{princess2014comprehensive}
\bibinfo{author}{Princess, M.R.}, \bibinfo{author}{Kumar, V.S.}, \bibinfo{author}{Begum, M.R.}, \bibinfo{year}{2014}.
\newblock \bibinfo{title}{Comprehensive and comparative study of different image fusion techniques}.
\newblock \bibinfo{journal}{Int. J. Adv. Res. Electr. Electron. Instrum. Eng} , \bibinfo{pages}{11800--11806}.
\bibitem[{Punjabi et~al.(2019)Punjabi, Martersteck, Wang, Parrish, Katsaggelos and ADNI}]{punjabi2019neuroimaging}
\bibinfo{author}{Punjabi, A.}, \bibinfo{author}{Martersteck, A.}, \bibinfo{author}{Wang, Y.}, \bibinfo{author}{Parrish, T.B.}, \bibinfo{author}{Katsaggelos, A.K.}, \bibinfo{author}{ADNI}, \bibinfo{year}{2019}.
\newblock \bibinfo{title}{Neuroimaging modality fusion in alzheimer’s classification using convolutional neural networks}.
\newblock \bibinfo{journal}{PloS one} \bibinfo{volume}{14}, \bibinfo{pages}{e0225759}.
\bibitem[{Puyol-Ant{\'o}n et~al.(2022)Puyol-Ant{\'o}n, Sidhu, Gould, Porter, Elliott, Mehta, Rinaldi and King}]{puyol2022multimodal}
\bibinfo{author}{Puyol-Ant{\'o}n, E.}, \bibinfo{author}{Sidhu, B.S.}, \bibinfo{author}{Gould, J.}, \bibinfo{author}{Porter, B.}, \bibinfo{author}{Elliott, M.K.}, \bibinfo{author}{Mehta, V.}, \bibinfo{author}{Rinaldi, C.A.}, \bibinfo{author}{King, A.P.}, \bibinfo{year}{2022}.
\newblock \bibinfo{title}{A multimodal deep learning model for cardiac resynchronisation therapy response prediction}.
\newblock \bibinfo{journal}{Medical Image Analysis} \bibinfo{volume}{79}, \bibinfo{pages}{102465}.
\bibitem[{Qian et~al.(2021)Qian, Pei, Zheng, Xie, Yan, Zhang, Han, Gao, Zhang, Zheng et~al.}]{qian2021prospective}
\bibinfo{author}{Qian, X.}, \bibinfo{author}{Pei, J.}, \bibinfo{author}{Zheng, H.}, \bibinfo{author}{Xie, X.}, \bibinfo{author}{Yan, L.}, \bibinfo{author}{Zhang, H.}, \bibinfo{author}{Han, C.}, \bibinfo{author}{Gao, X.}, \bibinfo{author}{Zhang, H.}, \bibinfo{author}{Zheng, W.}, et~al., \bibinfo{year}{2021}.
\newblock \bibinfo{title}{Prospective assessment of breast cancer risk from multimodal multiview ultrasound images via clinically applicable deep learning}.
\newblock \bibinfo{journal}{Nature biomedical engineering} \bibinfo{volume}{5}, \bibinfo{pages}{522--532}.
\bibitem[{Qian et~al.(2020)Qian, Zhang, Liu, Wang, Chen, Liu, Yang, Chen, Wei, Xiao et~al.}]{qian2020combined}
\bibinfo{author}{Qian, X.}, \bibinfo{author}{Zhang, B.}, \bibinfo{author}{Liu, S.}, \bibinfo{author}{Wang, Y.}, \bibinfo{author}{Chen, X.}, \bibinfo{author}{Liu, J.}, \bibinfo{author}{Yang, Y.}, \bibinfo{author}{Chen, X.}, \bibinfo{author}{Wei, Y.}, \bibinfo{author}{Xiao, Q.}, et~al., \bibinfo{year}{2020}.
\newblock \bibinfo{title}{A combined ultrasonic b-mode and color doppler system for the classification of breast masses using neural network}.
\newblock \bibinfo{journal}{European Radiology} \bibinfo{volume}{30}, \bibinfo{pages}{3023--3033}.
\bibitem[{Qin et~al.(2020)Qin, Wang, Jiang, Qiao, Hai, Chen, Xu, Shi and Yan}]{qin2020fine}
\bibinfo{author}{Qin, R.}, \bibinfo{author}{Wang, Z.}, \bibinfo{author}{Jiang, L.}, \bibinfo{author}{Qiao, K.}, \bibinfo{author}{Hai, J.}, \bibinfo{author}{Chen, J.}, \bibinfo{author}{Xu, J.}, \bibinfo{author}{Shi, D.}, \bibinfo{author}{Yan, B.}, \bibinfo{year}{2020}.
\newblock \bibinfo{title}{Fine-grained lung cancer classification from pet and ct images based on multidimensional attention mechanism}.
\newblock \bibinfo{journal}{Complexity} \bibinfo{volume}{2020}, \bibinfo{pages}{1--12}.
\bibitem[{Qiu et~al.(2023)Qiu, Zhao, Hou, Zhao, Zhang, Lin, Teng and Zhao}]{qiu2023hierarchical}
\bibinfo{author}{Qiu, L.}, \bibinfo{author}{Zhao, L.}, \bibinfo{author}{Hou, R.}, \bibinfo{author}{Zhao, W.}, \bibinfo{author}{Zhang, S.}, \bibinfo{author}{Lin, Z.}, \bibinfo{author}{Teng, H.}, \bibinfo{author}{Zhao, J.}, \bibinfo{year}{2023}.
\newblock \bibinfo{title}{Hierarchical multimodal fusion framework based on noisy label learning and attention mechanism for cancer classification with pathology and genomic features}.
\newblock \bibinfo{journal}{Computerized Medical Imaging and Graphics} , \bibinfo{pages}{102176}.
\bibitem[{Qiu et~al.(2022)Qiu, Miller, Joshi, Lee, Xue, Ni, Wang, De~Anda-Duran, Hwang, Cramer et~al.}]{qiu2022multimodal}
\bibinfo{author}{Qiu, S.}, \bibinfo{author}{Miller, M.I.}, \bibinfo{author}{Joshi, P.S.}, \bibinfo{author}{Lee, J.C.}, \bibinfo{author}{Xue, C.}, \bibinfo{author}{Ni, Y.}, \bibinfo{author}{Wang, Y.}, \bibinfo{author}{De~Anda-Duran, I.}, \bibinfo{author}{Hwang, P.H.}, \bibinfo{author}{Cramer, J.A.}, et~al., \bibinfo{year}{2022}.
\newblock \bibinfo{title}{Multimodal deep learning for alzheimer’s disease dementia assessment}.
\newblock \bibinfo{journal}{Nature communications} \bibinfo{volume}{13}, \bibinfo{pages}{3404}.
\bibitem[{Quellec et~al.(2010)Quellec, Lamard, Cazuguel, Roux and Cochener}]{quellec2010case}
\bibinfo{author}{Quellec, G.}, \bibinfo{author}{Lamard, M.}, \bibinfo{author}{Cazuguel, G.}, \bibinfo{author}{Roux, C.}, \bibinfo{author}{Cochener, B.}, \bibinfo{year}{2010}.
\newblock \bibinfo{title}{Case retrieval in medical databases by fusing heterogeneous information}.
\newblock \bibinfo{journal}{IEEE Transactions on Medical Imaging} \bibinfo{volume}{30}, \bibinfo{pages}{108--118}.
\bibitem[{Radford et~al.(2021)Radford, Kim, Hallacy, Ramesh, Goh, Agarwal, Sastry, Askell, Mishkin, Clark et~al.}]{radford2021learning}
\bibinfo{author}{Radford, A.}, \bibinfo{author}{Kim, J.W.}, \bibinfo{author}{Hallacy, C.}, \bibinfo{author}{Ramesh, A.}, \bibinfo{author}{Goh, G.}, \bibinfo{author}{Agarwal, S.}, \bibinfo{author}{Sastry, G.}, \bibinfo{author}{Askell, A.}, \bibinfo{author}{Mishkin, P.}, \bibinfo{author}{Clark, J.}, et~al., \bibinfo{year}{2021}.
\newblock \bibinfo{title}{Learning transferable visual models from natural language supervision}, in: \bibinfo{booktitle}{International conference on machine learning}, \bibinfo{organization}{PMLR}. pp. \bibinfo{pages}{8748--8763}.
\bibitem[{Rahaman et~al.(2021)Rahaman, Chen, Fu, Lewis, Iraji and Calhoun}]{rahaman2021multi}
\bibinfo{author}{Rahaman, M.A.}, \bibinfo{author}{Chen, J.}, \bibinfo{author}{Fu, Z.}, \bibinfo{author}{Lewis, N.}, \bibinfo{author}{Iraji, A.}, \bibinfo{author}{Calhoun, V.D.}, \bibinfo{year}{2021}.
\newblock \bibinfo{title}{Multi-modal deep learning of functional and structural neuroimaging and genomic data to predict mental illness}, in: \bibinfo{booktitle}{2021 43rd Annual International Conference of the IEEE Engineering in Medicine \& Biology Society (EMBC)}, \bibinfo{organization}{IEEE}. pp. \bibinfo{pages}{3267--3272}.
\bibitem[{Rahaman et~al.(2022)Rahaman, Garg, Iraj, Fu, Chen and Calhoun}]{rahaman2022two}
\bibinfo{author}{Rahaman, M.A.}, \bibinfo{author}{Garg, Y.}, \bibinfo{author}{Iraj, A.}, \bibinfo{author}{Fu, Z.}, \bibinfo{author}{Chen, J.}, \bibinfo{author}{Calhoun, V.}, \bibinfo{year}{2022}.
\newblock \bibinfo{title}{Two-dimensional attentive fusion for multi-modal learning of neuroimaging and genomics data}, in: \bibinfo{booktitle}{2022 IEEE 32nd International Workshop on Machine Learning for Signal Processing (MLSP)}, \bibinfo{organization}{IEEE}. pp. \bibinfo{pages}{1--6}.
\bibitem[{Rahman et~al.(2021)Rahman, Yang and Sigal}]{rahman2021tribert}
\bibinfo{author}{Rahman, T.}, \bibinfo{author}{Yang, M.}, \bibinfo{author}{Sigal, L.}, \bibinfo{year}{2021}.
\newblock \bibinfo{title}{Tribert: Full-body human-centric audio-visual representation learning for visual sound separation}.
\newblock \bibinfo{journal}{arXiv preprint arXiv:2110.13412} .
\bibitem[{Rallabandi and Seetharaman(2023)}]{rallabandi2023deep}
\bibinfo{author}{Rallabandi, V.S.}, \bibinfo{author}{Seetharaman, K.}, \bibinfo{year}{2023}.
\newblock \bibinfo{title}{Deep learning-based classification of healthy aging controls, mild cognitive impairment and alzheimer’s disease using fusion of mri-pet imaging}.
\newblock \bibinfo{journal}{Biomedical Signal Processing and Control} \bibinfo{volume}{80}, \bibinfo{pages}{104312}.
\bibitem[{Ramachandram and Taylor(2017)}]{ramachandram2017deep}
\bibinfo{author}{Ramachandram, D.}, \bibinfo{author}{Taylor, G.W.}, \bibinfo{year}{2017}.
\newblock \bibinfo{title}{Deep multimodal learning: A survey on recent advances and trends}.
\newblock \bibinfo{journal}{IEEE signal processing magazine} \bibinfo{volume}{34}, \bibinfo{pages}{96--108}.
\bibitem[{Ramesh et~al.(2021)Ramesh, Xing, Wang, Wang and Chen}]{ramesh2021vset}
\bibinfo{author}{Ramesh, K.}, \bibinfo{author}{Xing, C.}, \bibinfo{author}{Wang, W.}, \bibinfo{author}{Wang, D.}, \bibinfo{author}{Chen, X.}, \bibinfo{year}{2021}.
\newblock \bibinfo{title}{Vset: A multimodal transformer for visual speech enhancement}, in: \bibinfo{booktitle}{ICASSP 2021-2021 IEEE International Conference on Acoustics, Speech and Signal Processing (ICASSP)}, \bibinfo{organization}{IEEE}. pp. \bibinfo{pages}{6658--6662}.
\bibitem[{Sadjadi(2005)}]{sadjadi2005comparative}
\bibinfo{author}{Sadjadi, F.}, \bibinfo{year}{2005}.
\newblock \bibinfo{title}{Comparative image fusion analysais}, in: \bibinfo{booktitle}{2005 IEEE computer society conference on computer vision and pattern recognition (CVPR'05)-workshops}, \bibinfo{organization}{IEEE}. pp. \bibinfo{pages}{8--8}.
\bibitem[{Salahuddin et~al.(2022)Salahuddin, Woodruff, Chatterjee and Lambin}]{salahuddin2022transparency}
\bibinfo{author}{Salahuddin, Z.}, \bibinfo{author}{Woodruff, H.C.}, \bibinfo{author}{Chatterjee, A.}, \bibinfo{author}{Lambin, P.}, \bibinfo{year}{2022}.
\newblock \bibinfo{title}{Transparency of deep neural networks for medical image analysis: A review of interpretability methods}.
\newblock \bibinfo{journal}{Computers in biology and medicine} \bibinfo{volume}{140}, \bibinfo{pages}{105111}.
\bibitem[{Salvador et~al.(2019)Salvador, Canales-Rodr{\'\i}guez, Guerrero-Pedraza, Sarr{\'o}, Tordesillas-Guti{\'e}rrez, Maristany, Crespo-Facorro, McKenna and Pomarol-Clotet}]{salvador2019multimodal}
\bibinfo{author}{Salvador, R.}, \bibinfo{author}{Canales-Rodr{\'\i}guez, E.}, \bibinfo{author}{Guerrero-Pedraza, A.}, \bibinfo{author}{Sarr{\'o}, S.}, \bibinfo{author}{Tordesillas-Guti{\'e}rrez, D.}, \bibinfo{author}{Maristany, T.}, \bibinfo{author}{Crespo-Facorro, B.}, \bibinfo{author}{McKenna, P.}, \bibinfo{author}{Pomarol-Clotet, E.}, \bibinfo{year}{2019}.
\newblock \bibinfo{title}{Multimodal integration of brain images for mri-based diagnosis in schizophrenia}.
\newblock \bibinfo{journal}{Frontiers in neuroscience} \bibinfo{volume}{13}, \bibinfo{pages}{1203}.
\bibitem[{Sanford et~al.(2020)Sanford, Harmon, Turkbey, Kesani, Tuncer, Madariaga, Yang, Sackett, Mehralivand, Yan et~al.}]{sanford2020deep}
\bibinfo{author}{Sanford, T.}, \bibinfo{author}{Harmon, S.A.}, \bibinfo{author}{Turkbey, E.B.}, \bibinfo{author}{Kesani, D.}, \bibinfo{author}{Tuncer, S.}, \bibinfo{author}{Madariaga, M.}, \bibinfo{author}{Yang, C.}, \bibinfo{author}{Sackett, J.}, \bibinfo{author}{Mehralivand, S.}, \bibinfo{author}{Yan, P.}, et~al., \bibinfo{year}{2020}.
\newblock \bibinfo{title}{Deep-learning-based artificial intelligence for pi-rads classification to assist multiparametric prostate mri interpretation: A development study}.
\newblock \bibinfo{journal}{Journal of Magnetic Resonance Imaging} \bibinfo{volume}{52}, \bibinfo{pages}{1499--1507}.
\bibitem[{Saponaro et~al.(2024)Saponaro, Lizzi, Serra, Mainas, Oliva, Giuliano, Calderoni and Retico}]{saponaro2024deep}
\bibinfo{author}{Saponaro, S.}, \bibinfo{author}{Lizzi, F.}, \bibinfo{author}{Serra, G.}, \bibinfo{author}{Mainas, F.}, \bibinfo{author}{Oliva, P.}, \bibinfo{author}{Giuliano, A.}, \bibinfo{author}{Calderoni, S.}, \bibinfo{author}{Retico, A.}, \bibinfo{year}{2024}.
\newblock \bibinfo{title}{Deep learning based joint fusion approach to exploit anatomical and functional brain information in autism spectrum disorders}.
\newblock \bibinfo{journal}{Brain Informatics} \bibinfo{volume}{11}, \bibinfo{pages}{2}.
\bibitem[{Schelling et~al.(2000)Schelling, Braun, Kuhn, Bogner, Gruber, Gnirs, Schneider, Ulm, Rutke and Staudach}]{schelling2000combined}
\bibinfo{author}{Schelling, M.}, \bibinfo{author}{Braun, M.}, \bibinfo{author}{Kuhn, W.}, \bibinfo{author}{Bogner, G.}, \bibinfo{author}{Gruber, R.}, \bibinfo{author}{Gnirs, J.}, \bibinfo{author}{Schneider, K.T.}, \bibinfo{author}{Ulm, K.}, \bibinfo{author}{Rutke, S.}, \bibinfo{author}{Staudach, A.}, \bibinfo{year}{2000}.
\newblock \bibinfo{title}{Combined transvaginal b-mode and color doppler sonography for differential diagnosis of ovarian tumors: results of a multivariate logistic regression analysis}.
\newblock \bibinfo{journal}{Gynecologic oncology} \bibinfo{volume}{77}, \bibinfo{pages}{78--86}.
\bibitem[{Schelling et~al.(1997)Schelling, Gnirs, Braun, Busch, Maurer, Kuhn, Schneider and Graeff}]{schelling1997optimized}
\bibinfo{author}{Schelling, M.}, \bibinfo{author}{Gnirs, J.}, \bibinfo{author}{Braun, M.}, \bibinfo{author}{Busch, R.}, \bibinfo{author}{Maurer, S.}, \bibinfo{author}{Kuhn, W.}, \bibinfo{author}{Schneider, K.}, \bibinfo{author}{Graeff, H.}, \bibinfo{year}{1997}.
\newblock \bibinfo{title}{Optimized differential diagnosis of breast lesions by combined b-mode and color doppler sonography}.
\newblock \bibinfo{journal}{Ultrasound in Obstetrics and Gynecology: The Official Journal of the International Society of Ultrasound in Obstetrics and Gynecology} \bibinfo{volume}{10}, \bibinfo{pages}{48--53}.
\bibitem[{Sharma et~al.(2020)Sharma, Dogra, Goyal, Vig and Agrawal}]{sharma2020pyramids}
\bibinfo{author}{Sharma, A.M.}, \bibinfo{author}{Dogra, A.}, \bibinfo{author}{Goyal, B.}, \bibinfo{author}{Vig, R.}, \bibinfo{author}{Agrawal, S.}, \bibinfo{year}{2020}.
\newblock \bibinfo{title}{From pyramids to state-of-the-art: a study and comprehensive comparison of visible--infrared image fusion techniques}.
\newblock \bibinfo{journal}{IET Image Processing} \bibinfo{volume}{14}, \bibinfo{pages}{1671--1689}.
\bibitem[{Shen et~al.(2011)Shen, Xia, Kang, Yuan and Sheng}]{shen2011use}
\bibinfo{author}{Shen, J.M.}, \bibinfo{author}{Xia, X.W.}, \bibinfo{author}{Kang, W.G.}, \bibinfo{author}{Yuan, J.J.}, \bibinfo{author}{Sheng, L.}, \bibinfo{year}{2011}.
\newblock \bibinfo{title}{The use of mri apparent diffusion coefficient (adc) in monitoring the development of brain infarction}.
\newblock \bibinfo{journal}{BMC Medical Imaging} \bibinfo{volume}{11}, \bibinfo{pages}{1--4}.
\bibitem[{Shi et~al.(2022)Shi, Hsu, Lakhotia and Mohamed}]{shi2022learning}
\bibinfo{author}{Shi, B.}, \bibinfo{author}{Hsu, W.N.}, \bibinfo{author}{Lakhotia, K.}, \bibinfo{author}{Mohamed, A.}, \bibinfo{year}{2022}.
\newblock \bibinfo{title}{Learning audio-visual speech representation by masked multimodal cluster prediction}.
\newblock \bibinfo{journal}{arXiv preprint arXiv:2201.02184} .
\bibitem[{Shi et~al.(2017)Shi, Zheng, Li, Zhang and Ying}]{shi2017multimodal}
\bibinfo{author}{Shi, J.}, \bibinfo{author}{Zheng, X.}, \bibinfo{author}{Li, Y.}, \bibinfo{author}{Zhang, Q.}, \bibinfo{author}{Ying, S.}, \bibinfo{year}{2017}.
\newblock \bibinfo{title}{Multimodal neuroimaging feature learning with multimodal stacked deep polynomial networks for diagnosis of alzheimer's disease}.
\newblock \bibinfo{journal}{IEEE journal of biomedical and health informatics} \bibinfo{volume}{22}, \bibinfo{pages}{173--183}.
\bibitem[{Shoeibi et~al.(2022)Shoeibi, Khodatars, Jafari, Ghassemi, Moridian, Alizadesani, Ling, Khosravi, Alinejad-Rokny, Lam et~al.}]{shoeibi2022diagnosis}
\bibinfo{author}{Shoeibi, A.}, \bibinfo{author}{Khodatars, M.}, \bibinfo{author}{Jafari, M.}, \bibinfo{author}{Ghassemi, N.}, \bibinfo{author}{Moridian, P.}, \bibinfo{author}{Alizadesani, R.}, \bibinfo{author}{Ling, S.H.}, \bibinfo{author}{Khosravi, A.}, \bibinfo{author}{Alinejad-Rokny, H.}, \bibinfo{author}{Lam, H.}, et~al., \bibinfo{year}{2022}.
\newblock \bibinfo{title}{Diagnosis of brain diseases in fusion of neuroimaging modalities using deep learning: A review}.
\newblock \bibinfo{journal}{Information Fusion} .
\bibitem[{Simonyan and Zisserman(2014)}]{simonyan2014very}
\bibinfo{author}{Simonyan, K.}, \bibinfo{author}{Zisserman, A.}, \bibinfo{year}{2014}.
\newblock \bibinfo{title}{Very deep convolutional networks for large-scale image recognition}.
\newblock \bibinfo{journal}{arXiv preprint arXiv:1409.1556} .
\bibitem[{Singh and Nair(2022)}]{singh2022neural}
\bibinfo{author}{Singh, A.}, \bibinfo{author}{Nair, H.}, \bibinfo{year}{2022}.
\newblock \bibinfo{title}{A neural architecture search for automated multimodal learning}.
\newblock \bibinfo{journal}{Expert Systems with Applications} \bibinfo{volume}{207}, \bibinfo{pages}{118051}.
\bibitem[{Singh et~al.(2012)Singh, Gautam, Kumar and Umapathy}]{singh2012application}
\bibinfo{author}{Singh, B.}, \bibinfo{author}{Gautam, R.}, \bibinfo{author}{Kumar, S.}, \bibinfo{author}{Umapathy, S.}, \bibinfo{year}{2012}.
\newblock \bibinfo{title}{Application of vibrational microspectroscopy to biology and medicine} .
\bibitem[{Sleeman~IV et~al.(2022)Sleeman~IV, Kapoor and Ghosh}]{sleeman2022multimodal}
\bibinfo{author}{Sleeman~IV, W.C.}, \bibinfo{author}{Kapoor, R.}, \bibinfo{author}{Ghosh, P.}, \bibinfo{year}{2022}.
\newblock \bibinfo{title}{Multimodal classification: Current landscape, taxonomy and future directions}.
\newblock \bibinfo{journal}{ACM Computing Surveys} \bibinfo{volume}{55}, \bibinfo{pages}{1--31}.
\bibitem[{Song et~al.(2021)Song, Zheng, Li, Lu, Zhu and Shen}]{song2021effective}
\bibinfo{author}{Song, J.}, \bibinfo{author}{Zheng, J.}, \bibinfo{author}{Li, P.}, \bibinfo{author}{Lu, X.}, \bibinfo{author}{Zhu, G.}, \bibinfo{author}{Shen, P.}, \bibinfo{year}{2021}.
\newblock \bibinfo{title}{An effective multimodal image fusion method using mri and pet for alzheimer's disease diagnosis}.
\newblock \bibinfo{journal}{Frontiers in digital health} \bibinfo{volume}{3}, \bibinfo{pages}{637386}.
\bibitem[{Steitz et~al.(2022)Steitz, Pfeiffer, Gurevych and Roth}]{steitz2022txt}
\bibinfo{author}{Steitz, J.M.O.}, \bibinfo{author}{Pfeiffer, J.}, \bibinfo{author}{Gurevych, I.}, \bibinfo{author}{Roth, S.}, \bibinfo{year}{2022}.
\newblock \bibinfo{title}{Txt: Crossmodal end-to-end learning with transformers}, in: \bibinfo{booktitle}{Pattern Recognition: 43rd DAGM German Conference, DAGM GCPR 2021, Bonn, Germany, September 28--October 1, 2021, Proceedings}, \bibinfo{organization}{Springer}. pp. \bibinfo{pages}{405--420}.
\bibitem[{Sterne et~al.(2009)Sterne, White, Carlin, Spratt, Royston, Kenward, Wood and Carpenter}]{sterne2009multiple}
\bibinfo{author}{Sterne, J.A.}, \bibinfo{author}{White, I.R.}, \bibinfo{author}{Carlin, J.B.}, \bibinfo{author}{Spratt, M.}, \bibinfo{author}{Royston, P.}, \bibinfo{author}{Kenward, M.G.}, \bibinfo{author}{Wood, A.M.}, \bibinfo{author}{Carpenter, J.R.}, \bibinfo{year}{2009}.
\newblock \bibinfo{title}{Multiple imputation for missing data in epidemiological and clinical research: potential and pitfalls}.
\newblock \bibinfo{journal}{Bmj} \bibinfo{volume}{338}.
\bibitem[{Stokking et~al.(2001)Stokking, Zuiderveld and Viergever}]{stokking2001integrated}
\bibinfo{author}{Stokking, R.}, \bibinfo{author}{Zuiderveld, K.J.}, \bibinfo{author}{Viergever, M.A.}, \bibinfo{year}{2001}.
\newblock \bibinfo{title}{Integrated volume visualization of functional image data and anatomical surfaces using normal fusion}.
\newblock \bibinfo{journal}{Human Brain Mapping} \bibinfo{volume}{12}, \bibinfo{pages}{203--218}.
\bibitem[{Suk et~al.(2015)Suk, Lee, Shen and ADNI}]{suk2015latent}
\bibinfo{author}{Suk, H.I.}, \bibinfo{author}{Lee, S.W.}, \bibinfo{author}{Shen, D.}, \bibinfo{author}{ADNI}, \bibinfo{year}{2015}.
\newblock \bibinfo{title}{Latent feature representation with stacked auto-encoder for ad/mci diagnosis}.
\newblock \bibinfo{journal}{Brain Structure and Function} \bibinfo{volume}{220}, \bibinfo{pages}{841--859}.
\bibitem[{Suk et~al.(2014)Suk, Lee, Shen, ADNI et~al.}]{suk2014hierarchical}
\bibinfo{author}{Suk, H.I.}, \bibinfo{author}{Lee, S.W.}, \bibinfo{author}{Shen, D.}, \bibinfo{author}{ADNI}, et~al., \bibinfo{year}{2014}.
\newblock \bibinfo{title}{Hierarchical feature representation and multimodal fusion with deep learning for ad/mci diagnosis}.
\newblock \bibinfo{journal}{NeuroImage} \bibinfo{volume}{101}, \bibinfo{pages}{569--582}.
\bibitem[{Suk and Shen(2013)}]{suk2013deep}
\bibinfo{author}{Suk, H.I.}, \bibinfo{author}{Shen, D.}, \bibinfo{year}{2013}.
\newblock \bibinfo{title}{Deep learning-based feature representation for ad/mci classification}, in: \bibinfo{booktitle}{Medical Image Computing and Computer-Assisted Intervention--MICCAI 2013: 16th International Conference, Nagoya, Japan, September 22-26, 2013, Proceedings, Part II 16}, \bibinfo{organization}{Springer}. pp. \bibinfo{pages}{583--590}.
\bibitem[{Szegedy et~al.(2015)Szegedy, Liu, Jia, Sermanet, Reed, Anguelov, Erhan, Vanhoucke and Rabinovich}]{szegedy2015going}
\bibinfo{author}{Szegedy, C.}, \bibinfo{author}{Liu, W.}, \bibinfo{author}{Jia, Y.}, \bibinfo{author}{Sermanet, P.}, \bibinfo{author}{Reed, S.}, \bibinfo{author}{Anguelov, D.}, \bibinfo{author}{Erhan, D.}, \bibinfo{author}{Vanhoucke, V.}, \bibinfo{author}{Rabinovich, A.}, \bibinfo{year}{2015}.
\newblock \bibinfo{title}{Going deeper with convolutions}, in: \bibinfo{booktitle}{Proceedings of the IEEE conference on computer vision and pattern recognition}, pp. \bibinfo{pages}{1--9}.
\bibitem[{Taleb et~al.(2022)Taleb, Kirchler, Monti and Lippert}]{Taleb_2022_CVPR}
\bibinfo{author}{Taleb, A.}, \bibinfo{author}{Kirchler, M.}, \bibinfo{author}{Monti, R.}, \bibinfo{author}{Lippert, C.}, \bibinfo{year}{2022}.
\newblock \bibinfo{title}{Contig: Self-supervised multimodal contrastive learning for medical imaging with genetics}, in: \bibinfo{booktitle}{Proceedings of the IEEE/CVF Conference on Computer Vision and Pattern Recognition (CVPR)}, pp. \bibinfo{pages}{20908--20921}.
\bibitem[{Tan and Bansal(2019)}]{tan2019lxmert}
\bibinfo{author}{Tan, H.}, \bibinfo{author}{Bansal, M.}, \bibinfo{year}{2019}.
\newblock \bibinfo{title}{Lxmert: Learning cross-modality encoder representations from transformers}.
\newblock \bibinfo{journal}{arXiv preprint arXiv:1908.07490} .
\bibitem[{Tang et~al.(2022)Tang, Yan, Nan, Xiang, Krammer and Lasser}]{tang2022fusionm4net}
\bibinfo{author}{Tang, P.}, \bibinfo{author}{Yan, X.}, \bibinfo{author}{Nan, Y.}, \bibinfo{author}{Xiang, S.}, \bibinfo{author}{Krammer, S.}, \bibinfo{author}{Lasser, T.}, \bibinfo{year}{2022}.
\newblock \bibinfo{title}{Fusionm4net: A multi-stage multi-modal learning algorithm for multi-label skin lesion classification}.
\newblock \bibinfo{journal}{Medical Image Analysis} \bibinfo{volume}{76}, \bibinfo{pages}{102307}.
\bibitem[{Tang et~al.(2020)Tang, Xu, Han, Bai, Wang, Liu, Du, Liang, Zhang, Lu et~al.}]{tang2020elaboration}
\bibinfo{author}{Tang, X.}, \bibinfo{author}{Xu, X.}, \bibinfo{author}{Han, Z.}, \bibinfo{author}{Bai, G.}, \bibinfo{author}{Wang, H.}, \bibinfo{author}{Liu, Y.}, \bibinfo{author}{Du, P.}, \bibinfo{author}{Liang, Z.}, \bibinfo{author}{Zhang, J.}, \bibinfo{author}{Lu, H.}, et~al., \bibinfo{year}{2020}.
\newblock \bibinfo{title}{Elaboration of a multimodal mri-based radiomics signature for the preoperative prediction of the histological subtype in patients with non-small-cell lung cancer}.
\newblock \bibinfo{journal}{Biomedical engineering online} \bibinfo{volume}{19}, \bibinfo{pages}{1--17}.
\bibitem[{Tomczak et~al.(2015)Tomczak, Czerwi{\'n}ska and Wiznerowicz}]{tomczak2015review}
\bibinfo{author}{Tomczak, K.}, \bibinfo{author}{Czerwi{\'n}ska, P.}, \bibinfo{author}{Wiznerowicz, M.}, \bibinfo{year}{2015}.
\newblock \bibinfo{title}{Review the cancer genome atlas (tcga): an immeasurable source of knowledge}.
\newblock \bibinfo{journal}{Contemporary Oncology/Wsp{\'o}{\l}czesna Onkologia} \bibinfo{volume}{2015}, \bibinfo{pages}{68--77}.
\bibitem[{Tu et~al.(2024)Tu, Lin, Qiao, Zhuang, Wang and Wang}]{tu2024multimodal}
\bibinfo{author}{Tu, Y.}, \bibinfo{author}{Lin, S.}, \bibinfo{author}{Qiao, J.}, \bibinfo{author}{Zhuang, Y.}, \bibinfo{author}{Wang, Z.}, \bibinfo{author}{Wang, D.}, \bibinfo{year}{2024}.
\newblock \bibinfo{title}{Multimodal fusion diagnosis of alzheimer’s disease based on fdg-pet generation}.
\newblock \bibinfo{journal}{Biomedical Signal Processing and Control} \bibinfo{volume}{89}, \bibinfo{pages}{105709}.
\bibitem[{Tu et~al.(2022)Tu, Lin, Qiao, Zhuang and Zhang}]{tu2022alzheimer}
\bibinfo{author}{Tu, Y.}, \bibinfo{author}{Lin, S.}, \bibinfo{author}{Qiao, J.}, \bibinfo{author}{Zhuang, Y.}, \bibinfo{author}{Zhang, P.}, \bibinfo{year}{2022}.
\newblock \bibinfo{title}{Alzheimer’s disease diagnosis via multimodal feature fusion}.
\newblock \bibinfo{journal}{Computers in Biology and Medicine} \bibinfo{volume}{148}, \bibinfo{pages}{105901}.
\bibitem[{Vaswani et~al.(2017a)Vaswani, Shazeer, Parmar, Uszkoreit, Jones, Gomez, Kaiser and Polosukhin}]{https://doi.org/10.48550/arxiv.1706.03762}
\bibinfo{author}{Vaswani, A.}, \bibinfo{author}{Shazeer, N.}, \bibinfo{author}{Parmar, N.}, \bibinfo{author}{Uszkoreit, J.}, \bibinfo{author}{Jones, L.}, \bibinfo{author}{Gomez, A.N.}, \bibinfo{author}{Kaiser, L.}, \bibinfo{author}{Polosukhin, I.}, \bibinfo{year}{2017}a.
\newblock \bibinfo{title}{Attention is all you need}.
\newblock \URLprefix \url{https://arxiv.org/abs/1706.03762}, \DOIprefix\doi{10.48550/ARXIV.1706.03762}.
\bibitem[{Vaswani et~al.(2017b)Vaswani, Shazeer, Parmar, Uszkoreit, Jones, Gomez, Kaiser and Polosukhin}]{vaswani2017attention}
\bibinfo{author}{Vaswani, A.}, \bibinfo{author}{Shazeer, N.}, \bibinfo{author}{Parmar, N.}, \bibinfo{author}{Uszkoreit, J.}, \bibinfo{author}{Jones, L.}, \bibinfo{author}{Gomez, A.N.}, \bibinfo{author}{Kaiser, {\L}.}, \bibinfo{author}{Polosukhin, I.}, \bibinfo{year}{2017}b.
\newblock \bibinfo{title}{Attention is all you need}.
\newblock \bibinfo{journal}{Advances in neural information processing systems} \bibinfo{volume}{30}.
\bibitem[{Venugopalan et~al.(2021)Venugopalan, Tong, Hassanzadeh and Wang}]{venugopalan2021multimodal}
\bibinfo{author}{Venugopalan, J.}, \bibinfo{author}{Tong, L.}, \bibinfo{author}{Hassanzadeh, H.R.}, \bibinfo{author}{Wang, M.D.}, \bibinfo{year}{2021}.
\newblock \bibinfo{title}{Multimodal deep learning models for early detection of alzheimer’s disease stage}.
\newblock \bibinfo{journal}{Scientific reports} \bibinfo{volume}{11}, \bibinfo{pages}{3254}.
\bibitem[{Vu et~al.(2017)Vu, Yang, Nguyen, Oh and Kim}]{vu2017multimodal}
\bibinfo{author}{Vu, T.D.}, \bibinfo{author}{Yang, H.J.}, \bibinfo{author}{Nguyen, V.Q.}, \bibinfo{author}{Oh, A.R.}, \bibinfo{author}{Kim, M.S.}, \bibinfo{year}{2017}.
\newblock \bibinfo{title}{Multimodal learning using convolution neural network and sparse autoencoder}, in: \bibinfo{booktitle}{2017 IEEE international conference on big data and smart computing (BigComp)}, \bibinfo{organization}{IEEE}. pp. \bibinfo{pages}{309--312}.
\bibitem[{Wang(2020)}]{wang2020neural}
\bibinfo{author}{Wang, F.}, \bibinfo{year}{2020}.
\newblock \bibinfo{title}{Neural architecture search for gliomas segmentation on multimodal magnetic resonance imaging}.
\newblock \bibinfo{journal}{arXiv preprint arXiv:2005.06338} .
\bibitem[{Wang et~al.(2022)Wang, Wang, Yang, Zhang, Wang, Zhong, Zhang and Han}]{wang2022combining}
\bibinfo{author}{Wang, X.}, \bibinfo{author}{Wang, R.}, \bibinfo{author}{Yang, S.}, \bibinfo{author}{Zhang, J.}, \bibinfo{author}{Wang, M.}, \bibinfo{author}{Zhong, D.}, \bibinfo{author}{Zhang, J.}, \bibinfo{author}{Han, X.}, \bibinfo{year}{2022}.
\newblock \bibinfo{title}{Combining radiology and pathology for automatic glioma classification}.
\newblock \bibinfo{journal}{Frontiers in Bioengineering and Biotechnology} \bibinfo{volume}{10}.
\bibitem[{Wang et~al.(2018)Wang, Liu, Cheng, Wang, Yang and Cheng}]{wang2018automated}
\bibinfo{author}{Wang, Z.}, \bibinfo{author}{Liu, C.}, \bibinfo{author}{Cheng, D.}, \bibinfo{author}{Wang, L.}, \bibinfo{author}{Yang, X.}, \bibinfo{author}{Cheng, K.T.}, \bibinfo{year}{2018}.
\newblock \bibinfo{title}{Automated detection of clinically significant prostate cancer in mp-mri images based on an end-to-end deep neural network}.
\newblock \bibinfo{journal}{IEEE transactions on medical imaging} \bibinfo{volume}{37}, \bibinfo{pages}{1127--1139}.
\bibitem[{Wei et~al.(2022)Wei, Chen, Zhu, Zhang, Schönlieb, Price and Li}]{wei2022multimodal}
\bibinfo{author}{Wei, Y.}, \bibinfo{author}{Chen, X.}, \bibinfo{author}{Zhu, L.}, \bibinfo{author}{Zhang, L.}, \bibinfo{author}{Schönlieb, C.B.}, \bibinfo{author}{Price, S.J.}, \bibinfo{author}{Li, C.}, \bibinfo{year}{2022}.
\newblock \bibinfo{title}{Multi-modal learning for predicting the genotype of glioma}.
\newblock \href{http://arxiv.org/abs/2203.10852}{\tt arXiv:2203.10852}.
\bibitem[{Wei and Ji(2024)}]{wei2024multi}
\bibinfo{author}{Wei, Y.}, \bibinfo{author}{Ji, L.}, \bibinfo{year}{2024}.
\newblock \bibinfo{title}{Multi-modal bilinear fusion with hybrid attention mechanism for multi-label skin lesion classification}.
\newblock \bibinfo{journal}{Multimedia Tools and Applications} , \bibinfo{pages}{1--27}.
\bibitem[{Weiner et~al.(2017)Weiner, Veitch, Aisen, Beckett, Cairns, Green, Harvey, Jack~Jr, Jagust, Morris et~al.}]{weiner2017alzheimer}
\bibinfo{author}{Weiner, M.W.}, \bibinfo{author}{Veitch, D.P.}, \bibinfo{author}{Aisen, P.S.}, \bibinfo{author}{Beckett, L.A.}, \bibinfo{author}{Cairns, N.J.}, \bibinfo{author}{Green, R.C.}, \bibinfo{author}{Harvey, D.}, \bibinfo{author}{Jack~Jr, C.R.}, \bibinfo{author}{Jagust, W.}, \bibinfo{author}{Morris, J.C.}, et~al., \bibinfo{year}{2017}.
\newblock \bibinfo{title}{The alzheimer's disease neuroimaging initiative 3: Continued innovation for clinical trial improvement}.
\newblock \bibinfo{journal}{Alzheimer's \& Dementia} \bibinfo{volume}{13}, \bibinfo{pages}{561--571}.
\bibitem[{Weinstein et~al.(2013)Weinstein, Collisson, Mills, Shaw, Ozenberger, Ellrott, Shmulevich, Sander and Stuart}]{weinstein2013cancer}
\bibinfo{author}{Weinstein, J.N.}, \bibinfo{author}{Collisson, E.A.}, \bibinfo{author}{Mills, G.B.}, \bibinfo{author}{Shaw, K.R.}, \bibinfo{author}{Ozenberger, B.A.}, \bibinfo{author}{Ellrott, K.}, \bibinfo{author}{Shmulevich, I.}, \bibinfo{author}{Sander, C.}, \bibinfo{author}{Stuart, J.M.}, \bibinfo{year}{2013}.
\newblock \bibinfo{title}{The cancer genome atlas pan-cancer analysis project}.
\newblock \bibinfo{journal}{Nature genetics} \bibinfo{volume}{45}, \bibinfo{pages}{1113--1120}.
\bibitem[{Wu et~al.(2022)Wu, Fang, Li, Fu, Lin, Li, Huang, Yu, Song, Xu et~al.}]{wu2022gamma}
\bibinfo{author}{Wu, J.}, \bibinfo{author}{Fang, H.}, \bibinfo{author}{Li, F.}, \bibinfo{author}{Fu, H.}, \bibinfo{author}{Lin, F.}, \bibinfo{author}{Li, J.}, \bibinfo{author}{Huang, L.}, \bibinfo{author}{Yu, Q.}, \bibinfo{author}{Song, S.}, \bibinfo{author}{Xu, X.}, et~al., \bibinfo{year}{2022}.
\newblock \bibinfo{title}{Gamma challenge: glaucoma grading from multi-modality images}.
\newblock \bibinfo{journal}{arXiv preprint arXiv:2202.06511} .
\bibitem[{Wu et~al.(2023a)Wu, Fang, Li, Fu, Lin, Li, Huang, Yu, Song, Xu et~al.}]{wu2023gamma}
\bibinfo{author}{Wu, J.}, \bibinfo{author}{Fang, H.}, \bibinfo{author}{Li, F.}, \bibinfo{author}{Fu, H.}, \bibinfo{author}{Lin, F.}, \bibinfo{author}{Li, J.}, \bibinfo{author}{Huang, Y.}, \bibinfo{author}{Yu, Q.}, \bibinfo{author}{Song, S.}, \bibinfo{author}{Xu, X.}, et~al., \bibinfo{year}{2023}a.
\newblock \bibinfo{title}{Gamma challenge: glaucoma grading from multi-modality images}.
\newblock \bibinfo{journal}{Medical Image Analysis} \bibinfo{volume}{90}, \bibinfo{pages}{102938}.
\bibitem[{Wu et~al.(2023b)Wu, Wang, Zheng, Li, Alsaadi and Zeng}]{wu2023aggn}
\bibinfo{author}{Wu, P.}, \bibinfo{author}{Wang, Z.}, \bibinfo{author}{Zheng, B.}, \bibinfo{author}{Li, H.}, \bibinfo{author}{Alsaadi, F.E.}, \bibinfo{author}{Zeng, N.}, \bibinfo{year}{2023}b.
\newblock \bibinfo{title}{Aggn: Attention-based glioma grading network with multi-scale feature extraction and multi-modal information fusion}.
\newblock \bibinfo{journal}{Computers in biology and medicine} \bibinfo{volume}{152}, \bibinfo{pages}{106457}.
\bibitem[{Wu and Mebane~Jr(2022)}]{wu2022marmot}
\bibinfo{author}{Wu, P.Y.}, \bibinfo{author}{Mebane~Jr, W.R.}, \bibinfo{year}{2022}.
\newblock \bibinfo{title}{Marmot: A deep learning framework for constructing multimodal representations for vision-and-language tasks}.
\newblock \bibinfo{journal}{Computational Communication Research} \bibinfo{volume}{4}.
\bibitem[{Xi et~al.(2020)Xi, Zhao, Wang, Chang, Purkayastha, Chang, Huang, Silva, Valli{\`e}res, Habibollahi et~al.}]{xi2020deep}
\bibinfo{author}{Xi, I.L.}, \bibinfo{author}{Zhao, Y.}, \bibinfo{author}{Wang, R.}, \bibinfo{author}{Chang, M.}, \bibinfo{author}{Purkayastha, S.}, \bibinfo{author}{Chang, K.}, \bibinfo{author}{Huang, R.Y.}, \bibinfo{author}{Silva, A.C.}, \bibinfo{author}{Valli{\`e}res, M.}, \bibinfo{author}{Habibollahi, P.}, et~al., \bibinfo{year}{2020}.
\newblock \bibinfo{title}{Deep learning to distinguish benign from malignant renal lesions based on routine mr imagingdeep learning for characterization of renal lesions}.
\newblock \bibinfo{journal}{Clinical Cancer Research} \bibinfo{volume}{26}, \bibinfo{pages}{1944--1952}.
\bibitem[{Xi et~al.(2017)Xi, Luo, Zhang, You and Wu}]{xi2017multimodal}
\bibinfo{author}{Xi, X.X.}, \bibinfo{author}{Luo, X.Q.}, \bibinfo{author}{Zhang, Z.C.}, \bibinfo{author}{You, Q.J.}, \bibinfo{author}{Wu, X.}, \bibinfo{year}{2017}.
\newblock \bibinfo{title}{Multimodal medical volumetric image fusion based on multi-feature in 3-d shearlet transform}, in: \bibinfo{booktitle}{2017 International Smart Cities Conference (ISC2)}, \bibinfo{organization}{IEEE}. pp. \bibinfo{pages}{1--6}.
\bibitem[{Xie et~al.(2022)Xie, Wang, Huang, Lyu, Zheng, Zheng and Jin}]{https://doi.org/10.48550/arxiv.2202.06997}
\bibinfo{author}{Xie, G.}, \bibinfo{author}{Wang, J.}, \bibinfo{author}{Huang, Y.}, \bibinfo{author}{Lyu, J.}, \bibinfo{author}{Zheng, F.}, \bibinfo{author}{Zheng, Y.}, \bibinfo{author}{Jin, Y.}, \bibinfo{year}{2022}.
\newblock \bibinfo{title}{Cross-modality neuroimage synthesis: A survey}.
\newblock \URLprefix \url{https://arxiv.org/abs/2202.06997}, \DOIprefix\doi{10.48550/ARXIV.2202.06997}.
\bibitem[{Xing et~al.(2022a)Xing, Chen, Zhu, Hou, Gao and Yuan}]{Xing2022}
\bibinfo{author}{Xing, X.}, \bibinfo{author}{Chen, Z.}, \bibinfo{author}{Zhu, M.}, \bibinfo{author}{Hou, Y.}, \bibinfo{author}{Gao, Z.}, \bibinfo{author}{Yuan, Y.}, \bibinfo{year}{2022}a.
\newblock \bibinfo{title}{Discrepancy and gradient-guided multi-modal knowledge distillation for pathological glioma grading}, in: \bibinfo{editor}{Wang, L.}, \bibinfo{editor}{Dou, Q.}, \bibinfo{editor}{Fletcher, P.T.}, \bibinfo{editor}{Speidel, S.}, \bibinfo{editor}{Li, S.} (Eds.), \bibinfo{booktitle}{Medical Image Computing and Computer Assisted Intervention -- MICCAI 2022}, \bibinfo{publisher}{Springer Nature Switzerland}, \bibinfo{address}{Cham}. pp. \bibinfo{pages}{636--646}.
\bibitem[{Xing et~al.(2022b)Xing, Liang, Zhang, Khanal, Lin and Jacobs}]{xing2022advit}
\bibinfo{author}{Xing, X.}, \bibinfo{author}{Liang, G.}, \bibinfo{author}{Zhang, Y.}, \bibinfo{author}{Khanal, S.}, \bibinfo{author}{Lin, A.L.}, \bibinfo{author}{Jacobs, N.}, \bibinfo{year}{2022}b.
\newblock \bibinfo{title}{Advit: Vision transformer on multi-modality pet images for alzheimer disease diagnosis}, in: \bibinfo{booktitle}{2022 IEEE 19th International Symposium on Biomedical Imaging (ISBI)}, \bibinfo{organization}{IEEE}. pp. \bibinfo{pages}{1--4}.
\bibitem[{Xiong et~al.(2022)Xiong, Li, Song, Tang, He, Gao, Zhang, Cheng, Song, Lin et~al.}]{xiong2022multimodal}
\bibinfo{author}{Xiong, J.}, \bibinfo{author}{Li, F.}, \bibinfo{author}{Song, D.}, \bibinfo{author}{Tang, G.}, \bibinfo{author}{He, J.}, \bibinfo{author}{Gao, K.}, \bibinfo{author}{Zhang, H.}, \bibinfo{author}{Cheng, W.}, \bibinfo{author}{Song, Y.}, \bibinfo{author}{Lin, F.}, et~al., \bibinfo{year}{2022}.
\newblock \bibinfo{title}{Multimodal machine learning using visual fields and peripapillary circular oct scans in detection of glaucomatous optic neuropathy}.
\newblock \bibinfo{journal}{Ophthalmology} \bibinfo{volume}{129}, \bibinfo{pages}{171--180}.
\bibitem[{Xu et~al.(2024)Xu, Wang, Cai and Heng}]{xu2024cross}
\bibinfo{author}{Xu, D.}, \bibinfo{author}{Wang, X.}, \bibinfo{author}{Cai, J.}, \bibinfo{author}{Heng, P.A.}, \bibinfo{year}{2024}.
\newblock \bibinfo{title}{Cross-modality guidance-aided multi-modal learning with dual attention for mri brain tumor grading}.
\newblock \bibinfo{journal}{arXiv preprint arXiv:2401.09029} .
\bibitem[{Xu et~al.(2022)Xu, Li, Zhao, Quellec, Lu and Hatt}]{xu2022joint}
\bibinfo{author}{Xu, H.}, \bibinfo{author}{Li, Y.}, \bibinfo{author}{Zhao, W.}, \bibinfo{author}{Quellec, G.}, \bibinfo{author}{Lu, L.}, \bibinfo{author}{Hatt, M.}, \bibinfo{year}{2022}.
\newblock \bibinfo{title}{Joint nnu-net and radiomics approaches for segmentation and prognosis of head and neck cancers with pet/ct images}.
\newblock \bibinfo{journal}{arXiv preprint arXiv:2211.10138} .
\bibitem[{Xu et~al.(2023)Xu, Zhong and Zhang}]{xu2023multi}
\bibinfo{author}{Xu, H.}, \bibinfo{author}{Zhong, S.}, \bibinfo{author}{Zhang, Y.}, \bibinfo{year}{2023}.
\newblock \bibinfo{title}{Multi-level fusion network for mild cognitive impairment identification using multi-modal neuroimages}.
\newblock \bibinfo{journal}{Physics in Medicine \& Biology} \bibinfo{volume}{68}, \bibinfo{pages}{095018}.
\bibitem[{Xu et~al.(2016)Xu, Zhang, Huang, Zhang and Metaxas}]{xu2016multimodal}
\bibinfo{author}{Xu, T.}, \bibinfo{author}{Zhang, H.}, \bibinfo{author}{Huang, X.}, \bibinfo{author}{Zhang, S.}, \bibinfo{author}{Metaxas, D.N.}, \bibinfo{year}{2016}.
\newblock \bibinfo{title}{Multimodal deep learning for cervical dysplasia diagnosis}, in: \bibinfo{booktitle}{Medical Image Computing and Computer-Assisted Intervention--MICCAI 2016: 19th International Conference, Athens, Greece, October 17-21, 2016, Proceedings, Part II 19}, \bibinfo{organization}{Springer}. pp. \bibinfo{pages}{115--123}.
\bibitem[{Yan et~al.(2021)Yan, Zhang, Rao, Lv, Li, Zhang, Liang, Li, Ren, Zheng et~al.}]{yan2021richer}
\bibinfo{author}{Yan, R.}, \bibinfo{author}{Zhang, F.}, \bibinfo{author}{Rao, X.}, \bibinfo{author}{Lv, Z.}, \bibinfo{author}{Li, J.}, \bibinfo{author}{Zhang, L.}, \bibinfo{author}{Liang, S.}, \bibinfo{author}{Li, Y.}, \bibinfo{author}{Ren, F.}, \bibinfo{author}{Zheng, C.}, et~al., \bibinfo{year}{2021}.
\newblock \bibinfo{title}{Richer fusion network for breast cancer classification based on multimodal data}.
\newblock \bibinfo{journal}{BMC Medical Informatics and Decision Making} \bibinfo{volume}{21}, \bibinfo{pages}{1--15}.
\bibitem[{Yang et~al.(2017)Yang, Liu, Wang, Yang, Le~Min, Wang and Cheng}]{yang2017co}
\bibinfo{author}{Yang, X.}, \bibinfo{author}{Liu, C.}, \bibinfo{author}{Wang, Z.}, \bibinfo{author}{Yang, J.}, \bibinfo{author}{Le~Min, H.}, \bibinfo{author}{Wang, L.}, \bibinfo{author}{Cheng, K.T.T.}, \bibinfo{year}{2017}.
\newblock \bibinfo{title}{Co-trained convolutional neural networks for automated detection of prostate cancer in multi-parametric mri}.
\newblock \bibinfo{journal}{Medical image analysis} \bibinfo{volume}{42}, \bibinfo{pages}{212--227}.
\bibitem[{Yap et~al.(2018)Yap, Yolland and Tschandl}]{yap2018multimodal}
\bibinfo{author}{Yap, J.}, \bibinfo{author}{Yolland, W.}, \bibinfo{author}{Tschandl, P.}, \bibinfo{year}{2018}.
\newblock \bibinfo{title}{Multimodal skin lesion classification using deep learning}.
\newblock \bibinfo{journal}{Experimental dermatology} \bibinfo{volume}{27}, \bibinfo{pages}{1261--1267}.
\bibitem[{Ye et~al.(2017)Ye, Pu, Wang, Li and Zha}]{ye2017glioma}
\bibinfo{author}{Ye, F.}, \bibinfo{author}{Pu, J.}, \bibinfo{author}{Wang, J.}, \bibinfo{author}{Li, Y.}, \bibinfo{author}{Zha, H.}, \bibinfo{year}{2017}.
\newblock \bibinfo{title}{Glioma grading based on 3d multimodal convolutional neural network and privileged learning}, in: \bibinfo{booktitle}{2017 IEEE International Conference on Bioinformatics and Biomedicine (BIBM)}, \bibinfo{organization}{IEEE}. pp. \bibinfo{pages}{759--763}.
\bibitem[{Yin et~al.(2022)Yin, Huang and Zhang}]{yin2022bm}
\bibinfo{author}{Yin, Y.}, \bibinfo{author}{Huang, S.}, \bibinfo{author}{Zhang, X.}, \bibinfo{year}{2022}.
\newblock \bibinfo{title}{Bm-nas: Bilevel multimodal neural architecture search}, in: \bibinfo{booktitle}{Proceedings of the AAAI Conference on Artificial Intelligence}, pp. \bibinfo{pages}{8901--8909}.
\bibitem[{Ying et~al.(2021)Ying, Xing, Liu, Lin, Jacobs and Liang}]{ying2021multi}
\bibinfo{author}{Ying, Q.}, \bibinfo{author}{Xing, X.}, \bibinfo{author}{Liu, L.}, \bibinfo{author}{Lin, A.L.}, \bibinfo{author}{Jacobs, N.}, \bibinfo{author}{Liang, G.}, \bibinfo{year}{2021}.
\newblock \bibinfo{title}{Multi-modal data analysis for alzheimer’s disease diagnosis: An ensemble model using imagery and genetic features}, in: \bibinfo{booktitle}{2021 43rd Annual International Conference of the IEEE Engineering in Medicine \& Biology Society (EMBC)}, \bibinfo{organization}{IEEE}. pp. \bibinfo{pages}{3586--3591}.
\bibitem[{Yoo et~al.(2022)Yoo, Kim, Kim, Lee, Byeon, Kim, Yeo and Choi}]{yoo2022deeppdt}
\bibinfo{author}{Yoo, T.K.}, \bibinfo{author}{Kim, S.H.}, \bibinfo{author}{Kim, M.}, \bibinfo{author}{Lee, C.S.}, \bibinfo{author}{Byeon, S.H.}, \bibinfo{author}{Kim, S.S.}, \bibinfo{author}{Yeo, J.}, \bibinfo{author}{Choi, E.Y.}, \bibinfo{year}{2022}.
\newblock \bibinfo{title}{Deeppdt-net: predicting the outcome of photodynamic therapy for chronic central serous chorioretinopathy using two-stage multimodal transfer learning}.
\newblock \bibinfo{journal}{Scientific Reports} \bibinfo{volume}{12}, \bibinfo{pages}{18689}.
\bibitem[{Yu et~al.(2020)Yu, Cui, Yu, Wang, Tao and Tian}]{yu2020deep}
\bibinfo{author}{Yu, Z.}, \bibinfo{author}{Cui, Y.}, \bibinfo{author}{Yu, J.}, \bibinfo{author}{Wang, M.}, \bibinfo{author}{Tao, D.}, \bibinfo{author}{Tian, Q.}, \bibinfo{year}{2020}.
\newblock \bibinfo{title}{Deep multimodal neural architecture search}, in: \bibinfo{booktitle}{Proceedings of the 28th ACM International Conference on Multimedia}, pp. \bibinfo{pages}{3743--3752}.
\bibitem[{Zhang et~al.(2019)Zhang, Li, Zhang, Du, Wang and Zhang}]{zhang2019multi}
\bibinfo{author}{Zhang, F.}, \bibinfo{author}{Li, Z.}, \bibinfo{author}{Zhang, B.}, \bibinfo{author}{Du, H.}, \bibinfo{author}{Wang, B.}, \bibinfo{author}{Zhang, X.}, \bibinfo{year}{2019}.
\newblock \bibinfo{title}{Multi-modal deep learning model for auxiliary diagnosis of alzheimer’s disease}.
\newblock \bibinfo{journal}{Neurocomputing} \bibinfo{volume}{361}, \bibinfo{pages}{185--195}.
\bibitem[{Zhang et~al.(2022a)Zhang, He, Qing, Gao and Wang}]{zhang2022bpgan}
\bibinfo{author}{Zhang, J.}, \bibinfo{author}{He, X.}, \bibinfo{author}{Qing, L.}, \bibinfo{author}{Gao, F.}, \bibinfo{author}{Wang, B.}, \bibinfo{year}{2022}a.
\newblock \bibinfo{title}{Bpgan: brain pet synthesis from mri using generative adversarial network for multi-modal alzheimer’s disease diagnosis}.
\newblock \bibinfo{journal}{Computer Methods and Programs in Biomedicine} \bibinfo{volume}{217}, \bibinfo{pages}{106676}.
\bibitem[{Zhang et~al.(2018)Zhang, Liu, Blum, Han and Tao}]{zhang2018sparse}
\bibinfo{author}{Zhang, Q.}, \bibinfo{author}{Liu, Y.}, \bibinfo{author}{Blum, R.S.}, \bibinfo{author}{Han, J.}, \bibinfo{author}{Tao, D.}, \bibinfo{year}{2018}.
\newblock \bibinfo{title}{Sparse representation based multi-sensor image fusion for multi-focus and multi-modality images: A review}.
\newblock \bibinfo{journal}{Information Fusion} \bibinfo{volume}{40}, \bibinfo{pages}{57--75}.
\bibitem[{Zhang and Shi(2020)}]{zhang2020multi}
\bibinfo{author}{Zhang, T.}, \bibinfo{author}{Shi, M.}, \bibinfo{year}{2020}.
\newblock \bibinfo{title}{Multi-modal neuroimaging feature fusion for diagnosis of alzheimer’s disease}.
\newblock \bibinfo{journal}{Journal of Neuroscience Methods} \bibinfo{volume}{341}, \bibinfo{pages}{108795}.
\bibitem[{Zhang et~al.(2021)Zhang, Lin, Xiao and Ji}]{zhang2021multimodal}
\bibinfo{author}{Zhang, X.}, \bibinfo{author}{Lin, W.}, \bibinfo{author}{Xiao, M.}, \bibinfo{author}{Ji, H.}, \bibinfo{year}{2021}.
\newblock \bibinfo{title}{Multimodal 2.5 d convolutional neural network for diagnosis of alzheimer's disease with magnetic resonance imaging and positron emission tomography.}
\newblock \bibinfo{journal}{Progress In Electromagnetics Research} \bibinfo{volume}{171}.
\bibitem[{Zhang et~al.(2022b)Zhang, Deng, Zhou, Zhang, Jiao and Zhao}]{zhang2022multimodal}
\bibinfo{author}{Zhang, Y.}, \bibinfo{author}{Deng, Y.}, \bibinfo{author}{Zhou, Z.}, \bibinfo{author}{Zhang, X.}, \bibinfo{author}{Jiao, P.}, \bibinfo{author}{Zhao, Z.}, \bibinfo{year}{2022}b.
\newblock \bibinfo{title}{Multimodal learning for fetal distress diagnosis using a multimodal medical information fusion framework}.
\newblock \bibinfo{journal}{Frontiers in Physiology} , \bibinfo{pages}{2362}.
\bibitem[{Zhang et~al.(2022c)Zhang, Jiang, Miura, Manning and Langlotz}]{zhang2022contrastive}
\bibinfo{author}{Zhang, Y.}, \bibinfo{author}{Jiang, H.}, \bibinfo{author}{Miura, Y.}, \bibinfo{author}{Manning, C.D.}, \bibinfo{author}{Langlotz, C.P.}, \bibinfo{year}{2022}c.
\newblock \bibinfo{title}{Contrastive learning of medical visual representations from paired images and text}, in: \bibinfo{booktitle}{Machine Learning for Healthcare Conference}, \bibinfo{organization}{PMLR}. pp. \bibinfo{pages}{2--25}.
\bibitem[{Zhou et~al.(2021a)Zhou, Zhang, Zhu, Lan, Fu, Wang and Wen}]{zhou2021cohesive}
\bibinfo{author}{Zhou, J.}, \bibinfo{author}{Zhang, X.}, \bibinfo{author}{Zhu, Z.}, \bibinfo{author}{Lan, X.}, \bibinfo{author}{Fu, L.}, \bibinfo{author}{Wang, H.}, \bibinfo{author}{Wen, H.}, \bibinfo{year}{2021}a.
\newblock \bibinfo{title}{Cohesive multi-modality feature learning and fusion for covid-19 patient severity prediction}.
\newblock \bibinfo{journal}{IEEE Transactions on Circuits and Systems for Video Technology} \bibinfo{volume}{32}, \bibinfo{pages}{2535--2549}.
\bibitem[{Zhou et~al.(2021b)Zhou, Jiang, Yu, Feng, Chen, Li, Liu and Huang}]{zhou2021use}
\bibinfo{author}{Zhou, P.}, \bibinfo{author}{Jiang, S.}, \bibinfo{author}{Yu, L.}, \bibinfo{author}{Feng, Y.}, \bibinfo{author}{Chen, C.}, \bibinfo{author}{Li, F.}, \bibinfo{author}{Liu, Y.}, \bibinfo{author}{Huang, Z.}, \bibinfo{year}{2021}b.
\newblock \bibinfo{title}{Use of a sparse-response deep belief network and extreme learning machine to discriminate alzheimer's disease, mild cognitive impairment, and normal controls based on amyloid pet/mri images}.
\newblock \bibinfo{journal}{Frontiers in Medicine} \bibinfo{volume}{7}, \bibinfo{pages}{621204}.
\bibitem[{Zhou et~al.(2019a)Zhou, Ruan and Canu}]{zhou2019review}
\bibinfo{author}{Zhou, T.}, \bibinfo{author}{Ruan, S.}, \bibinfo{author}{Canu, S.}, \bibinfo{year}{2019}a.
\newblock \bibinfo{title}{A review: Deep learning for medical image segmentation using multi-modality fusion}.
\newblock \bibinfo{journal}{Array} \bibinfo{volume}{3}, \bibinfo{pages}{100004}.
\bibitem[{Zhou et~al.(2017)Zhou, Thung, Zhu and Shen}]{zhou2017feature}
\bibinfo{author}{Zhou, T.}, \bibinfo{author}{Thung, K.H.}, \bibinfo{author}{Zhu, X.}, \bibinfo{author}{Shen, D.}, \bibinfo{year}{2017}.
\newblock \bibinfo{title}{Feature learning and fusion of multimodality neuroimaging and genetic data for multi-status dementia diagnosis}, in: \bibinfo{booktitle}{Machine Learning in Medical Imaging: 8th International Workshop, MLMI 2017, Held in Conjunction with MICCAI 2017, Quebec City, QC, Canada, September 10, 2017, Proceedings 8}, \bibinfo{organization}{Springer}. pp. \bibinfo{pages}{132--140}.
\bibitem[{Zhou et~al.(2019b)Zhou, Thung, Zhu and Shen}]{zhou2019effective}
\bibinfo{author}{Zhou, T.}, \bibinfo{author}{Thung, K.H.}, \bibinfo{author}{Zhu, X.}, \bibinfo{author}{Shen, D.}, \bibinfo{year}{2019}b.
\newblock \bibinfo{title}{Effective feature learning and fusion of multimodality data using stage-wise deep neural network for dementia diagnosis}.
\newblock \bibinfo{journal}{Human brain mapping} \bibinfo{volume}{40}, \bibinfo{pages}{1001--1016}.
\bibitem[{Zhou et~al.(2023)Zhou, Adrada, Candelaria, Elshafeey, Boge, Mohamed, Pashapoor, Sun, Xu, Panthi et~al.}]{zhou2023prediction}
\bibinfo{author}{Zhou, Z.}, \bibinfo{author}{Adrada, B.E.}, \bibinfo{author}{Candelaria, R.P.}, \bibinfo{author}{Elshafeey, N.A.}, \bibinfo{author}{Boge, M.}, \bibinfo{author}{Mohamed, R.M.}, \bibinfo{author}{Pashapoor, S.}, \bibinfo{author}{Sun, J.}, \bibinfo{author}{Xu, Z.}, \bibinfo{author}{Panthi, B.}, et~al., \bibinfo{year}{2023}.
\newblock \bibinfo{title}{Prediction of pathologic complete response to neoadjuvant systemic therapy in triple negative breast cancer using deep learning on multiparametric mri}.
\newblock \bibinfo{journal}{Scientific Reports} \bibinfo{volume}{13}, \bibinfo{pages}{1171}.
\bibitem[{Zhu and Yang(2020)}]{zhu2020actbert}
\bibinfo{author}{Zhu, L.}, \bibinfo{author}{Yang, Y.}, \bibinfo{year}{2020}.
\newblock \bibinfo{title}{Actbert: Learning global-local video-text representations}, in: \bibinfo{booktitle}{Proceedings of the IEEE/CVF conference on computer vision and pattern recognition}, pp. \bibinfo{pages}{8746--8755}.
\bibitem[{Zhu et~al.(2019)Zhu, Zheng, Qi, Wang and Xiang}]{zhu2019phase}
\bibinfo{author}{Zhu, Z.}, \bibinfo{author}{Zheng, M.}, \bibinfo{author}{Qi, G.}, \bibinfo{author}{Wang, D.}, \bibinfo{author}{Xiang, Y.}, \bibinfo{year}{2019}.
\newblock \bibinfo{title}{A phase congruency and local laplacian energy based multi-modality medical image fusion method in nsct domain}.
\newblock \bibinfo{journal}{IEEE Access} \bibinfo{volume}{7}, \bibinfo{pages}{20811--20824}.
\bibitem[{Zong et~al.(2020)Zong, Lee, Liu, Carver, Feldman, Janic, Elshaikh, Pantelic, Hearshen, Chetty et~al.}]{zong2020deep}
\bibinfo{author}{Zong, W.}, \bibinfo{author}{Lee, J.K.}, \bibinfo{author}{Liu, C.}, \bibinfo{author}{Carver, E.N.}, \bibinfo{author}{Feldman, A.M.}, \bibinfo{author}{Janic, B.}, \bibinfo{author}{Elshaikh, M.A.}, \bibinfo{author}{Pantelic, M.V.}, \bibinfo{author}{Hearshen, D.}, \bibinfo{author}{Chetty, I.J.}, et~al., \bibinfo{year}{2020}.
\newblock \bibinfo{title}{A deep dive into understanding tumor foci classification using multiparametric mri based on convolutional neural network}.
\newblock \bibinfo{journal}{Medical physics} \bibinfo{volume}{47}, \bibinfo{pages}{4077--4086}.
\bibitem[{Zou et~al.(2017)Zou, Zheng, Miao, Mckeown and Wang}]{zou20173d}
\bibinfo{author}{Zou, L.}, \bibinfo{author}{Zheng, J.}, \bibinfo{author}{Miao, C.}, \bibinfo{author}{Mckeown, M.J.}, \bibinfo{author}{Wang, Z.J.}, \bibinfo{year}{2017}.
\newblock \bibinfo{title}{3d cnn based automatic diagnosis of attention deficit hyperactivity disorder using functional and structural mri}.
\newblock \bibinfo{journal}{Ieee Access} \bibinfo{volume}{5}, \bibinfo{pages}{23626--23636}.
\bibitem[{Zuo et~al.(2023)Zuo, Shen, Zhong, Chen, Lei and Wang}]{zuo2023alzheimer}
\bibinfo{author}{Zuo, Q.}, \bibinfo{author}{Shen, Y.}, \bibinfo{author}{Zhong, N.}, \bibinfo{author}{Chen, C.P.}, \bibinfo{author}{Lei, B.}, \bibinfo{author}{Wang, S.}, \bibinfo{year}{2023}.
\newblock \bibinfo{title}{Alzheimer’s disease prediction via brain structural-functional deep fusing network}.
\newblock \bibinfo{journal}{IEEE Transactions on Neural Systems and Rehabilitation Engineering} \bibinfo{volume}{31}, \bibinfo{pages}{4601--4612}.
\bibitem[{Zwiebel and Pellerito(2005)}]{zwiebel2005introduction}
\bibinfo{author}{Zwiebel, W.J.}, \bibinfo{author}{Pellerito, J.S.}, \bibinfo{year}{2005}.
\newblock \bibinfo{title}{Introduction to vascular ultrasonography}.
\newblock \bibinfo{publisher}{Elsevier Saunders Philadelphia}.

\end{thebibliography}

\end{document}